\documentclass[12pt]{amsart}
\usepackage[utf8]{inputenc}
\usepackage{amsmath, amssymb, amsthm}
\usepackage{geometry}
\usepackage{enumerate}
\usepackage{hyperref}
\usepackage{natbib}

\usepackage{fullpage}
\usepackage{algorithm}
\usepackage{algpseudocode}

\theoremstyle{definition}

\usepackage{amsbsy, amstext, amssymb, amsthm, amsmath, bm,bbm,wasysym}
\usepackage{mathrsfs}
\usepackage{graphicx,booktabs,rotating,enumitem}
\usepackage{float}
\usepackage{anyfontsize} 
\usepackage{listings}
\usepackage{algorithm}
\usepackage{algpseudocode}
\usepackage{array,stmaryrd}
\usepackage{tabularx}
\usepackage{pdflscape}
\usepackage{adjustbox}
\usepackage{comment}
\usepackage{subfig}
\usepackage{longtable}
\usepackage{multicol}
\usepackage[table]{xcolor}

\usepackage{etoolbox}
\patchcmd{\APACjournalVolNumPages}{\unskip({#3})}{}{}{} 
\patchcmd{\APACjournalVolNumPages}{\Bem{#2}}{#2}{}{}

\usepackage{tikz}
\usetikzlibrary{shapes.geometric, arrows}
\usetikzlibrary{arrows.meta}
\usetikzlibrary{shapes,backgrounds,calc,positioning}
\DeclareMathOperator*{\argmin}{argmin}

\usepackage{blkarray}

\newcommand{\T}{\mathrm{\scriptscriptstyle T}}
\newcommand{\floor}[1]{\lfloor #1 \rfloor}

\title{Amortized Bayesian inference for actigraph time sheet data from mobile devices}
\author{Daniel Zhou and Sudipto Banerjee\\ UCLA Department of Biostatistics, 650 Charles E. Young Drive, Los Angeles, 90095-1772.}

\date{\today}

\begin{document}

\begin{abstract}
Mobile data technologies use ``actigraphs'' to furnish information on health variables as a function of a subject's movement. The advent of wearable devices and related technologies has propelled the creation of health databases consisting of human movement data to conduct research on mobility patterns and health outcomes. Statistical methods for analyzing high-resolution actigraph data depend on the specific inferential context, but the advent of Artificial Intelligence (AI) frameworks require that the methods be congruent to transfer learning and amortization. This article devises amortized Bayesian inference for actigraph time sheets. We pursue a Bayesian approach to ensure full propagation of uncertainty and its quantification using a hierarchical dynamic linear model. We build our analysis around actigraph data from the Physical Activity through Sustainable Transport Approaches in Los Angeles (PASTA-LA) study conducted by the Fielding School of Public Health in the University of California, Los Angeles. Apart from achieving probabilistic imputation of actigraph time sheets, we are also able to statistically learn about the time-varying impact of explanatory variables on the magnitude of acceleration (MAG) for a cohort of subjects.  
\end{abstract}

\maketitle

\section{Introduction}
\label{sec:intro}

Recent developments in wearable device technologies have led to the collection of mobile health data \citep{james2016, drewnowski2020} at very high resolutions (e.g., step rates, blood pressure, heartbeat, activity counts, etc.). \emph{Actigraphs} are small portable devices with motion sensors, or \emph{accelerometers}, that detect physical movement by measuring acceleration along different axes and are able to collect large amounts of data \citep{plasqui2007physical, sikka2019analytics}. They are increasingly conspicuous because of their affordability, accuracy, and availability in smart-phones, smart-watches, and other wearable devices. The collected data can be quickly downloaded and easily accessible for statistical analysis to obtain insight into their pattern and structure. 

There is a rapidly emerging literature on statistical analysis of data streaming from wearable devices \citep{chang2022empirical, luo2023streaned, banker2023accelerometer}.  
Actigraph data analysis, which can be regarded as a subset of streaming data analysis, is witnessing rapid growth in interest among biomedical and health scientists seeking to understand how environmental factors interact with the personal attributes of a subject to define activity-related health outcomes. Examples include, but are not limited to, the use of data from wearable devices as biomarkers and risk factors in the study of adverse health outcomes for respiratory health \citep[][]{Kim2024}. 

In this article, we devise a framework to carry out temporal analysis of an original actigraph data set from the \textbf{P}hysical \textbf{A}ctivity through \textbf{S}ustainable \textbf{T}ransport \textbf{A}pproaches in \textbf{L}os \textbf{A}ngeles (PASTA-LA) study. The scientific objectives of the PASTA-LA study are broad and diverse. This article is concerned with devising a transfer learning framework for the PASTA-LA study using amortized Bayesian inference \citep{radev2022, zammitmangion2024, ZammitMangion2025NeuralMethodsAmortizedInference}, which constitutes a key component of producing fast and reliable inference using an AI-based interface for health providers. 

The outcome we model is the magnitude of acceleration (MAG). Let $x$, $y$ and $z$ be the dynamic acceleration of the body. MAG at time point $t$ is defined as:
\begin{equation}
\label{eq: MAG}
   \text{MAG}_{t}=\sqrt{x_{t}^2+y_{t}^2+z_{t}^2}, \qquad t=1,\dots,T .
\end{equation} 
However, the instantaneous MAG evaluated at the original frequency $30$Hz is erratic and does not adequately represent the subject's intensity of physical activity at that time. To amend this problem, we average the MAG values over 20 second-time steps, using an approach similar to \citep{migueles2017accelerometer, doherty2017large}.

To motivate people to exercise, one of our interests is in deploying a human-interpretable system that takes into account environmental factors in the data to preemptively construct a path or paths to exercise. Doing so would provide a numerical benchmark for people to achieve as part of an exercise goal, as well as reduce or eliminate the overhead of having to decide on a route for exercise beforehand. Thus, the domain demands a temporal regression model, which can be employed for analysis and feedback; as well as a generative model to sample new trajectories should they be deemed relevant.

Our specific contribution is to train neural networks using hierarchical dynamic models \citep{west_harrison} within the \texttt{BayesFlow} learning framework \citep{radev2022}. In this process, we demonstrate the utility and structural flexibility of amortized inference towards analyzing MAGs from wearable devices. Bayesian analysis using exact posterior sampling from the hierarchical dynamic linear model serves as our benchmark for evaluating the performance of amortized inference. While much of the statistical distribution theory used here is fairly familiar, we believe that the transfer learning framework devised here for wearable devices data is a novel contribution.

The remainder of the article proceeds as follows. Section~\ref{sec:actigraph_data} presents some details on the processing and structure of our actigraph data. Section~\ref{sec:actigraph_timesheet} introduces the actigraph time sheet and explains how its format accommodates analysis using dynamic linear models. Section~\ref{sec:ABI} provides an overview of amortized Bayesian inference as executed by \texttt{BayesFlow}. Section~\ref{sec:DLM} extracts and organizes the essential modeling ingredients for training. Section~\ref{sec:illustrations} illustrates the training of the network using simulated data and applies the trained network to our actigraph time sheet. Finally, Section~\ref{sec:discussion} presents a brief discussion to conclude the paper. 

\section{Actigraph Data}
\label{sec:actigraph_data}

Actigraphy is a noninvasive method of monitoring human rest/activity cycles. The resulting data are collected from the relevant actigraph unit to assess the cycles of activity and rest over several days to several weeks. The data used in this study come from the \textbf{P}hysical \textbf{A}ctivity through \textbf{S}ustainable \textbf{T}ransport \textbf{A}pproaches in \textbf{L}os \textbf{A}ngeles (PASTA-LA) study conducted on a cohort of 460 individuals monitored between May 2017 and June 2018 to assess their physical activity in the Westwood borough of Los Angeles for two separate periods of one week each. Data were collected from various sources, including online questionnaires, a GPS device (Global-Sat DG-500), and a portable actigraph unit (Actigraph GT3X+). The resulting actigraph data are joined from biological and livelihood factors such as age, height, weight, ethnicity, sex, and BMI; geospatial coordinates, time measures, and geographical measures such as latitude, longitude, altitude, distance from home, distance from work, and distance from some other point of interest; and the MAG actigraph measurement. The primary outcome MAG is directly measured from the motion of the actigraph sensor and has statistical relationships with other physical activity measures such as the metabolic equivalent of task (MET) and other energy expenditure measures. \citep{hildebrand2014age, staudenmayer2015methods,sasaki2011validation,aguilar2019comparing,mortazavi2013METcalcs} 
The study protocols to protect participant information received the necessary approval from the institutional review board (IRB). The data were stored on a secure computer and a redacted version was created for data sharing purposes.

The GPS and actigraph devices were deployed on a nested sample due to cost considerations, and then retrieved from 94 of these individuals. We take a slightly roundabout approach and focus instead on the trajectories taken by the subjects: we trim the dataset further to specific days and trajectories that are both sufficiently active and fall within certain time intervals to acquire a subset of trajectories that contain enough activity to be further extended as desired. \citep{actigraph} 

\subsection{Preprocessing}
\label{sec:data_preprocessing}

The primary motivation to pursue inference on actigraph data is to build a recommender system to generate trajectories and intensities for prospective patients. This is especially relevant for people who do not exercise for a significant period of time in their day-to-day lives. We also infer that the particular time the individual exercises has more of an impact than the particular day individual exercises, even more so as the days the subjects were recorded fell into the fall and spring seasons in the Westwood borough of Los Angeles, where temperatures were less likely to be extreme.

Each trajectory taken by a subject has its own start time and end time and rarely, if ever, intersects temporally with other subjects' trajectories. This is because subjects are typically inclined to move and exercise at their own chosen times. Approaches must be taken to circumvent the sparsity and granularity of the raw data; the sparsity that exists by the design of the data collection procedure and the granularity that results from the precision of the time measurements recorded by the actigraph devices.

\begin{figure}[t]
\begin{centering}
    \includegraphics[height=8cm]{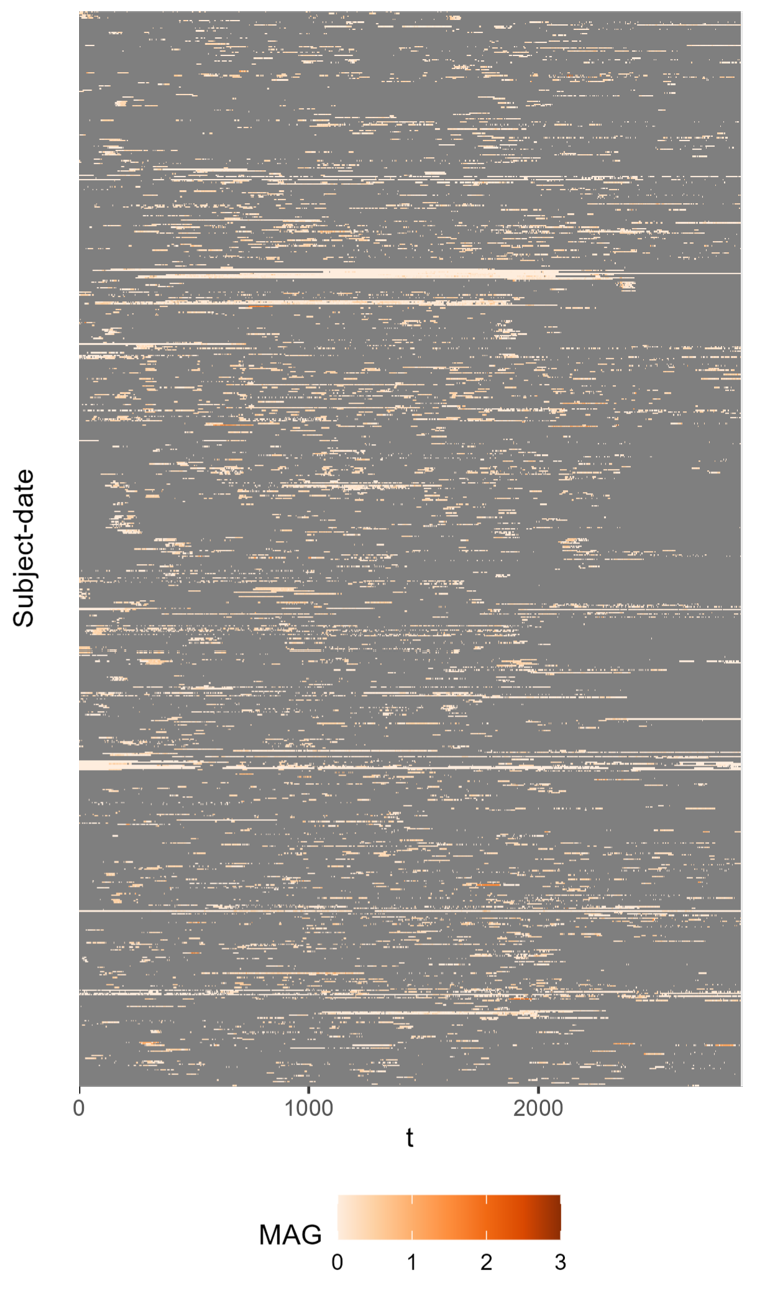}
    \includegraphics[height=8.022cm]{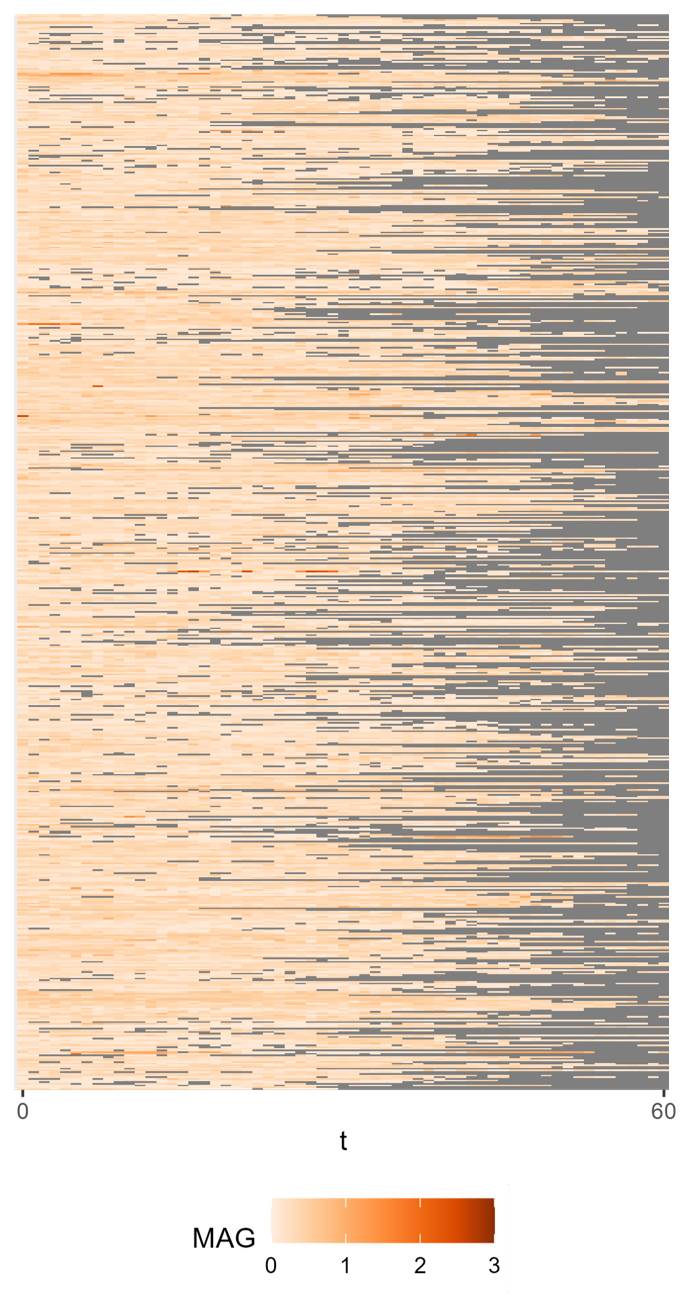}
    \caption{Left: The data covered by each subject for each date covered in the dataset under the absolute time step setup, where time ranges from 7 am - 11 pm. Right: The same under the relative time step setup from start of trajectory to up to 20 minutes later, with trajectories going beyond 20 minutes filtered out or cut off at the 22 minute mark. The color corresponds to the MAG value for the subject at a particular date for that time. Gray cells correspond to missing data for the subject-date pair at that particular epoch.}
    \label{fig:existMAG}
\end{centering}
\end{figure}

We define trajectories as distinct where the end time of one trajectory and the start time of the next on the same day is greater than 3 minutes. To focus our analysis on the trajectories most relevant to recommend to prospective patients, we also filter out trajectories that are shorter than 5 minutes to minimize the influence of short bursts of activity, which generally amount to a short walk, as well as trajectories longer than 22 minutes. We also filter out time points for which the MAG is less than $0.05$ as a precaution against small amounts of activity that are not considered exercise, such as driving or riding an electric scooter around the neighborhood, which will involve small amounts of motion in the wrists that can be registered by the actigraph device. Finally, we cut off the remaining trajectories between 20 and 22 minutes at the 20 minute marks. Hence, the exercise trajectories we consider comprise 5 - 20 minutes of activity over the course of 20 second-time steps at a relatively consistent rate of at least minimal activity. The sparsity of the actigraph dataset is then substantially reduced, so that the data set can be more easily managed by the desired algorithms. Figure~\ref{fig:existMAG} illustrates the change in sparsity in the data before and after trajectory transformation: Under the absolute time step treatment, the data would span approximately 9.0\% of all possible time steps over each day covered by a subject, not all of which are directly relevant to their exercise periods or regiments.\footnote{Most people would not need a recommender system to assist in building an exercise routine that spans 7 am - 11 pm.} Under the relative time step setup, the data would span 45.2\% of the total time points under consideration. This setup reduces the number of subjects to 92.

Finally, we recall from Equation \eqref{eq: MAG} that the MAG is nonnegative by nature of its formulation and equals zero only when all $x_t$, $y_t$, and $z_t$ are zero. Since the dynamic linear model which we will apply to the data is designed for positive and negative outcomes, our main focus would be the log(MAG), where $\log$ is the natural logarithm, to give us outcomes that may be negative. Removing MAG's less than $0.05$ in order to build our recommender system also removes numerical issues with taking the $\log$ of the minimum MAG (i.e. $\log(0)$), while in principle allowing for negative values below $\log(0.05)$ to translate into small MAG's, and therefore, small activity levels. Our choice of transformation, which can extend to other transformations of the MAG, will be less of an issue for amortized Bayesian inference and is merely chosen to give us a starting point for the dynamic linear model.

\section{The Actigraph Timesheet}
\label{sec:actigraph_timesheet}

Preprocessing the actigraph data set removes most of the data gaps (Figure \ref{fig:existMAG}) and reworks the data to fit a recommender system. However, not all trajectories reach the 20 minute-mark and some trajectories contain gaps so that it does not span every time step up to the last time step in the trajectory. Hence, we pair subjects with the dates they were active, so that we are only recording subject activities for the dates for which we would expect them to be active, and index the trajectory further by starting time point. This (subject, date, first time step) tuple suffices to index each trajectory in the dataset. The existing data are then used to train the model. We name this data rearrangement the \textit{actigraph timesheet}.

\begin{table}[]
    \centering
    \begin{tabular}{|c|c|c|c|c|}
    \hline
        Subject-Date & Time 1 & Time 2 & $\cdots$ & Time $T$ \\\hline\hline
        S1\_D1 & A111 & A112 & $\cdots$ & A11$T$ \\\hline
        S1\_D2 & \cellcolor{lightgray}{\textit{n/a}} & A122 & $\cdots$ & A12$T$\\\hline
        $\vdots$ & $\vdots$ & $\vdots$ & $\vdots$ & $\vdots$\\\hline
        S2\_D1 & \cellcolor{lightgray}{\textit{n/a}} & \cellcolor{lightgray}{\textit{n/a}} & $\cdots$ & A21$T$\\\hline
        S2\_D2 & \cellcolor{lightgray}{\textit{n/a}} & A212 & $\cdots$ & A22$T$\\\hline
        $\vdots$ & $\vdots$ & $\vdots$ & $\vdots$ & $\vdots$\\\hline
        Ss\_D$n_{s}$ & As$n_{s}$1 & As$n_{s}$2 & $\cdots$ & \cellcolor{lightgray}{\textit{n/a}}\\\hline
    \end{tabular}
    \caption{A sample actigraph timesheet with $s$ subjects and $n_{s}$ dates recorded for $T$ discrete time steps. Subjects and dates are paired (e.g. S1\_D1 denotes subject 1 paired with date 1), and data is recorded at each time (e.g. A111 denotes the data recorded for subject 1 at date 1 and time 1). {\textit{n/a}} (highlighted in grey) denotes cells where the data is missing.}
    \label{tab:actigraph_timesheet}
\end{table}

\subsection{Imputation Schema}
\label{sec:imputation_schema}

The actigraph timesheet is amenable to discrete temporal analysis by design and groups the specific dates an individual exercises with the subject who exercises on that day. Either dataset visualized in Figure \ref{fig:existMAG} can be expressed in the structure of Table \ref{tab:actigraph_timesheet}, a visual representation of a hypothetical dataset with missing values across time and sets up the dataset for the timesheet to impute. Let $\bm{X}_{t,o}$ and $\bm{y}_{t,o}$ be the $n_{t,o}\times p$ observed covariate design matrix and $n_{t,o}\times 1$ outcome respectively at time $t$. $\bm{X}_{t,u}$ denotes the $n_{t,u}\times p$ design matrix of covariates that we treat as known, but may not be observed. $\bm{y}_{t,u}$ is the $n_{t,u}\times 1$ unobserved outcome. For full generality, suppose that we are working with the known model $\bm y_t = g_t(\bm\theta_t, \bm X_t, \bm V_t)$, which is governed by a vector or matrix of other known parameters $\bm V_t$ and may be involve random components such as noise. Then the full imputation procedure for the missing time steps of the actigraph timesheet can be listed as follows:
\begin{enumerate}
    \label{AT_procedure}
    \item  Compute the posterior $\bm{\theta}_{1:T}\mid\bm y_{1:T,o}$ by regressing on the observed data $\bm y_{1:T,o}$ using observed covariates $\bm X_{1:T,o}$ and parameters $\bm V_{t,o}$.
    \item Analytically impute the unobserved covariates $\bm{X}_{t,u}$, generating $\bm{X}_{t,u}^*$.\footnote{This is ideally done with a procedure that accounts for time placement and geospatial locations, as we do in this paper.}
    \item Sample $\bm\theta^{(l)}_{1:T} \sim p(\bm{\theta}_{1:T}\mid\bm y_{1:T,o})$ from the posterior and synthesize outcomes to impute at the time steps where the outcomes are not observed: $\bm{y}_{t,u}^{*,(l)} = g_t(\bm\theta^{(l)}_{t}, \bm X_{t,u}^*, \bm V_{t,u})$ for $t=1,\ldots,T$ and $l = 1,\ldots,L$ desired samples. 
\end{enumerate}
Step 3 shows that multiple predictive samples of imputed outcomes can be easily acquired from this framework by taking multiple posterior samples of $\bm\theta_{1:T}\mid \bm y_{1:T,o}$ and then passing each $\bm\theta_t\mid \bm y_{1:T,o}$ to $g_{t}(\cdot, \bm X_{t,u}^*, \bm V_{t,u})$ for all $t$.

\subsection{A Normal-Normal Example}
\label{sec:AT_NN_example}

To give a concrete example for the actigraph timesheet procedure and to prepare the procedure for application to our use case in Section \ref{sec:predict_AT_impute}, we will assume that our observations at each time point are normal at each time point $t$; and that they share a common scale variance $\sigma^2$, though the variances $\bm y_{t,o}$ and $\bm y_{t,u}$ also depend on matrices which we will assume to be known. Here, let $\bm\theta_t := \bm\beta_t$, the fixed effects vector at time $t$; we will assume that $\sigma^2$ is known for simplicity. We specify $g_t$ as follows:
\begin{equation}
    \label{eq:AT_lin_setup}
    g_t(\bm\beta_t, \bm X_{t}, \sigma^2\bm V_t) := \bm X_t\bm\beta_t + \bm\nu_t,\quad \bm\nu_t \sim \mathcal{N}_{n}(\bm 0,\sigma^2\bm V_t)
\end{equation}
where $\mathcal{N}_{n}(\bm 0,\sigma^2\bm{V}_t)$ is the $n$-dimensional multivariate normal distribution with mean $\bm{0}$ and variance $\sigma^2\bm{V}_t$. We assume in Equation \eqref{eq:AT_lin_setup} that $\bm V_{t}$ are known for all $t$.

A formulation of $g_t$ in terms of both observed and unobserved terms takes the form of Equation \eqref{eq:AT_norm_at_t}:

\begin{equation}
    \label{eq:AT_norm_at_t}
    \bm y_t^\dagger = g_t(\bm\beta_t, \bm X_t^\dagger, \sigma^2\bm V_t^\dagger)= \bm X_t^\dagger\bm{\beta}_{t} + \bm n_t^\dagger
\end{equation}
where
\begin{equation*}
    \bm y_t^\dagger = \begin{bmatrix}
        \bm{y}_{t,o} \\ \bm{y}_{t,u}
    \end{bmatrix}, \quad \bm X_t^\dagger = \begin{bmatrix}
        \bm{X}_{t,o} \\ \bm{X}_{t,u}
    \end{bmatrix}, \quad
    \bm \nu_t^\dagger = \begin{bmatrix}
        \bm{\nu}_{t,o}\\ \bm{\nu}_{t,u}
    \end{bmatrix}\sim \mathcal{N}_{n_{t,o} + n_{t,u}}(\bm 0,\sigma^2 \bm V_{t}^\dagger), \quad \bm{V}_{t}^\dagger = \begin{bmatrix}
        \bm{V}_{t,oo} & \bm{V}_{t,ou}\\
        \bm{V}_{t,uo} & \bm{V}_{t,uu}
    \end{bmatrix}.
\end{equation*}

Conforming to the problem setting and structure of actigraph data, $n_{t,o}$ and $n_{t,u}$ are allowed to vary over time $t$. As with $\bm V_{t}$ in Equation \eqref{eq:AT_lin_setup}, we also assume that $\bm{V}_{t}^\dagger$ and its submatrices $\bm V_{t,oo}$, $\bm V_{t,ou}$, $\bm V_{t,uo}$, and $\bm V_{t,uu}$ are all known for all $t$.

Now let's further suppose that $\bm\beta_t$ follows a normal density, so that $\bm\beta_t\sim \mathcal{N}(\bm m_t,\sigma^2\bm M_t)$, and that $\bm\beta_t$ are independent across time $t$. Conjugacy gives us the posterior $\bm\beta_t\mid\bm y_t \sim \mathcal{N}_{p}(\bm m_{t,\text{post}}, \sigma^2\bm M_{t,\text{post}})$, where
\begin{equation*}
\bm M_{t,\text{post}} = (\bm X_{t,o}^\T\bm V_{t,oo}^{-1}\bm X_{t,o} + \bm M_{t}^{-1})^{-1} 
\quad\mbox{ and }\quad \bm m_{t,\text{post}} = \bm M_{t,\text{post}}(\bm X_{t,o}^\T\bm V_{t,oo}^{-1}\bm y_{t,o} + \bm M_{t}^{-1}\bm m_{t}).
\end{equation*}
Since we have specified that $\bm\beta_t$ are independent across different $t$ for this example, we don't need to worry about accounting for outcomes from times other than $t$.

To construct $\bm{X}_{t,u}^*$, covariates which depend on the subject-specific paths of traversal (e.g. GPS coordinates) can be linearly-interpolated, since the time between the previous and next points are assumed to be linearly distributed. Covariates which depend on the geographical coordinates can be taken directly from the associated latitude-longitude pair if the coordinate pair exists in the dataset and then averaged among locations in a set radius around the coordinates using a Gaussian-weighted average of the points by distance if the pair does not exist in the dataset. Subject-specific information, such as age, which was not assumed to change significantly during the course of the study, was taken directly from the subject's recorded information in the data set.

Finally, we sample $L$ vectors from the posterior: $\bm\beta_t^{(l)} \sim \mathcal{N}(\bm m_{t,\text{post}}, \sigma^2\bm M_{t,\text{post}})$, $l = 1,\ldots,L$, and generate synthetic outcomes $\bm y_{t,u}^{*,(1:L)} \sim \mathcal{N}_{n_t}(\bm X_{t,u}^*\bm\beta_t,\sigma^2\bm V_{t,uu})$ for $t = 1,\ldots,T$. Credible intervals for each $\bm y_{t,u}$ can be computed by taking quantiles for $\bm y_{t,u}^{*,(1:L)}$ across the $L$ samples.

\section{Amortized Bayesian Inference}
\label{sec:ABI}

In one interpretation of the transfer learning framework, the goalis to build a system that takes a data set $\bm y$ as input and builds posterior samples $\bm\theta\mid \bm y$ of parameter estimates as output. This system generalizes across different datasets within its application domain without overfitting to any one of them. Hence, we want to train a function $f_{\bm\phi}$ with parameters $\bm\phi$ so that, given the separate data outcomes $\bm y^{(d)}$, we can obtain posterior parameter estimates $\bm\theta\mid \bm y^{(d)} = f_{\bm\phi}(\bm y^{(d)})$ for all $d=1,\ldots,D$. (We momentarily dispense with the time suffix $t$ from notation in this section to address the framework in full generality.)

The particular approach to learning $f_{\bm\phi}$ is called \textit{amortized bayesian inference} (ABI). The defining approach of ABI is to train $f_{\bm\phi}$ on a sequence of artificially-generated synthetic datasets, and then produce posterior samples for each individual dataset that is fed into the trained network $f_{\widehat{\bm\phi}}$. "Amortization" refers to the high time cost of training the neural network system, which is then averaged over passing multiple datasets through the trained system to yield a low average runtime. Thus, evaluating $f_{\bm\phi}(\bm y^{(d)})$ would no longer involve the high time cost of the training step in a standard machine learning framework where we train each data set $\bm y^{(d)}$ separately and from scratch; the average time cost of training and running the system reduces ("amortizes") over the number of times or different settings $d=1,\ldots,D$ the trained system is run on.

To support the interpretability of our samples from our trained ABI system and to give ourselves a first step to model our data, we begin with a baseline model defined by a prior distribution $p(\bm\theta)$ and data generating distribution $p(\bm y\mid \bm\theta)$ (which we may combine into the joint distribution $p(\bm y,\bm\theta)$) to remove the dependence of $f_{\widehat{\bm\phi}}$ on any particular dataset, and to simultaneously provide a model that is structurally suited to the common application domain of the datasets (such as a time-dependent model, in the context of actigraph data). $p(\bm y, \bm\theta)$ then generates our prior and synthetic data samples, both of which will be used to train $f_{\bm\phi}$ to learn the posterior sampling function, which mandates $f_{\bm\phi}$ to take in $\bm\theta$ and $\bm y$. The particular method we utilize $f_{\bm\phi}$ is to transform $\bm\theta\mid \bm y$ to a random variable following a simpler, more recognizable distribution for training purposes and back when sampling is required. The latter approach is already widely applied to sampling from arbitrary distributions: Sampling from the distribution of an arbitrary random variable $X$ typically computes $F_{X}^{-1}(U)$, where $F_X$ is the cumulative distribution function of $X$ and $U \sim U(0,1)$ is a sampled uniform random variable on the interval $(0,1)$; theoretical developments utilizing this framework have been advanced by \citep{chen_gopinath_2000, tabak2010density, tabakturner2013}.

$f_{\bm\phi}$ is called a \textit{normalizing flow} on the basis of its application to bidirectional variable transformation. Our use of normalizing flows is inspired by and derives from the success of existing methods \citep{rezende15varinfnormflow, dinh2015nicenonlinearindependentcomponents, radev2022}.\footnote{For a thorough treatment of normalizing flows, \citep[see][and references therein]{DeepmindNormalizingFlows2021}.} The method pursues variational approximation of the true posterior $p(\bm\theta\mid \bm y)$ by the reverse flow $p(f_{\bm\phi}^{-1}(\bm z;\bm y))$, where $\bm z$ follows the simple distribution (in our applications, a multivariate standard normal $\mathcal{N}(\bm 0,\bm I)$), by finding the parameters $\bm\phi$ that minimize the Kullback-Leibler (KL) Divergence between the two quantities \citep{ren2011variational, blei2017variational}. The expectation across all datasets $y$ is taken for the KL Divergence to account for the underlying application of the ABI model to diverse datasets, and further makes the expression more tractable:
\begin{equation}
    \label{eq:kl_2_objective}
    \begin{split}
&\argmin_{\bm\phi} \mathbb{E}_{p(\bm y)}\mathbb{KL}(p(\bm\theta\mid \bm y)\mid\mid p(f_{\bm\phi}^{-1}(\bm z; \bm y)))  \\
        &\qquad\qquad=\argmin_{\bm\phi} \mathbb{E}_{p(\bm y)}\mathbb{E}_{p(\bm\theta\mid\bm y)}[\log p(\bm\theta\mid \bm y) - \log p(f_{\bm\phi}^{-1}(\bm z; \bm y)))]\\
        &\qquad\qquad= -\argmin_{\bm\phi}\mathbb{E}_{p(\bm y)}\mathbb{E}_{p(\bm\theta\mid\bm y)}\log p(f_{\bm\phi}^{-1}(\bm z; \bm y))\\
        &\qquad\qquad= -\argmin_{\bm\phi}\iint p(\bm y, \bm\theta)\log p(f_{\bm\phi}^{-1}(\bm z; \bm y))d\bm y d\bm \theta.
    \end{split}
\end{equation}
We finish with a change of variables $\bm z = f_{\bm\phi}(\bm\theta; \bm y)$ in the final expression to remove the dependence of our equation on $\bm z$, which is not sampled in the training stage:
\begin{equation}
    \label{eq:kl_objective_noz}
    \begin{split}
        &-\argmin_{\bm\phi}\iint p(\bm y, \bm\theta)\log p(f_{\bm\phi}^{-1}(\bm z; \bm y))d\bm y d\bm \theta\\
        &\qquad\qquad= -\argmin_{\bm\phi}\iint p(\bm y, \bm\theta)[\log p(f_{\bm\phi}(\bm \theta; \bm y)) + \log \lvert\det \bm J_{f_{\bm\phi}}\rvert]d\bm y d\bm \theta
    \end{split}
\end{equation}
where $\bm{J}_{f_{\bm{\phi}}} = \partial f_{\bm{\phi}}(\bm\theta; \bm{y})/\partial\bm{\theta}$, the Jacobian of the invertible network in the forward direction. The final integral is then approximated with the Monte Carlo expression over $M$ priors $\bm\theta^{(m)}$, $m=1,\ldots,M$, with their corresponding datasets $\bm y^{(m)}$ generated from each prior:
\begin{equation*}
    \label{eq:bf_objfunc_presimp}
    \begin{split}
        \widehat{\bm\phi} = -\argmin_{\bm\phi}\frac{1}{M}\sum_{m=1}^{M}\left(\log p(f_{\bm\phi}(\bm\theta^{(m)}; \bm y^{(m)})) + \log\lvert\det \bm J_{f_{\bm\phi}}^{(m)}\rvert\right).
    \end{split}
\end{equation*}
Since $f_{\bm\phi}$ transforms $\bm\theta$ into a standard normal with additional input from $\bm y$, we can treat $p(f_{\bm\phi}(\bm\theta^{(m)}; \bm y^{(m)}))$ as a standard normal random variable and simplify $\log p(f_{\bm\phi}(\bm\theta^{(m)}; \bm y^{(m)}))$ to $-\frac{D}{2}\log(2\pi) - \frac{1}{2}\lvert\lvert f_{\bm\phi}(\bm\theta^{(m)}; \bm y^{(m)})\rvert\rvert^{2}$, where $D$ is the dimension of $\bm\theta$. Factoring in the $\argmin$ allows us to remove the constant term from our sum, giving us our objective function:
\begin{equation}
    \label{eq:bf_objfunc}
    \begin{split}
        \widehat{\bm\phi} = \argmin_{\bm\phi}\frac{1}{M}\sum_{m=1}^{M}\left(\frac{1}{2}\lvert\lvert f_{\bm\phi}(\bm\theta^{(m)}; \bm y^{(m)})\rvert\rvert^{2} - \log\lvert\det \bm J_{f_{\bm\phi}}^{(m)}\rvert\right).
    \end{split}
\end{equation}

Algorithm~\ref{alg:ABI} explicates the \texttt{BayesFlow} procedure in terms of the two functions \\\texttt{TRAIN\_BAYESFLOW}() and \texttt{SAMPLE\_BAYESFLOW}(): At each training iteration, $M$ priors and synthetic data are generated and passed to the inference network $f_{\bm\phi}$ to output Monte Carlo samples for Equation \eqref{eq:bf_objfunc}. The parameters of the invertible neural network $\bm\phi$ are updated at each training cycle using backpropagation (discussed further in Algorithm \ref{alg:backprop_CF} and Algorithm \ref{alg:ADAM}).

\begin{algorithm}
    \caption{Amortized Bayesian Inference with the \texttt{BayesFlow} Method}\label{alg:ABI}
    \begin{algorithmic}[1]
    \vspace{2mm}
    \State \textbf{Input:} Prior generating process $p(\bm\theta)$, synthetic outcome generating process $p(\bm y\mid \bm\theta)$
    \vspace{2mm}
    \State \qquad\quad INN $f_{\bm\phi}(\bm\theta;\bm y)$ to be trained
    \vspace{2mm}
    \State \qquad\quad Batch size $M$, number of iterations to use for training $N_{ITER}$
    \vspace{2mm}
    \Function{\texttt{Train\_BayesFlow}}{$p(\bm\theta), p(\bm y\mid \bm\theta), f_{\bm\phi}(\bm\theta;\bm y),M,N_{ITER}$}
    \vspace{2mm}
    \For{$j=1,\ldots,N_{ITER}$}
    \vspace{2mm}
    \For{$m=1,\ldots,M$}
        \vspace{2mm}
        \State Sample model parameters from prior $\bm{\theta}^{(m,j)} \sim p(\bm{\theta})$.
        \vspace{2mm}
        \State Generate synthetic outcomes $\bm{y}^{(m,j)} \sim p(\bm{y}\mid \bm{\theta}^{(m,j)})$.
        \vspace{2mm}
        \State Pass $(\bm{\theta}^{(m,j)}, \bm{y}^{(m,j)})$ through the inference network in the forward direction: 
        \vspace{2mm}
        \State \qquad\quad $\bm{z}^{(m,j)} := f_{\bm{\phi}}(\bm{\theta}^{(m,j)}; \bm{y}^{(m,j)})$.
        \vspace{2mm}
    \EndFor
        \vspace{2mm}
        \State Compute loss according to Equation \eqref{eq:bf_objfunc}.
        \vspace{2mm}
        \State Update neural network parameters $\bm{\phi}$ via backpropagation. 
        \vspace{2mm}
        \EndFor \Comment{Convergence to $\bm{\phi}$ ideally achieved}
    \vspace{2mm}
    \EndFunction
    \vspace{2mm}
    \State
    \vspace{2mm}
    \State \textbf{Input:} Observed data $\bm y^{(obs)}$
    \vspace{2mm}, trained INN $f_{\widehat{\bm\phi}}(\bm\theta;\bm y)$ from \texttt{TRAIN\_BAYESFLOW}
    \vspace{2mm}
    \State \qquad\quad Number of samples $L$
    \vspace{2mm}
    \State \textbf{Output:} $L$ posterior samples from $p(\bm\theta\mid \bm{y}^{(obs)})$
    \vspace{2mm}
    \Function{\texttt{Sample\_BayesFlow}}{$\bm y^{(obs)}, f_{\widehat{\bm\phi}}(\bm\theta;\bm y), L$}    
    \vspace{2mm}
    \For{$l = 1,\ldots,L$}
        \vspace{2mm}
        \State Sample a latent variable instance $\bm{z}^{(l)}\sim \mathcal{N}_{D}(\bm{0},\bm{I})$.
        \vspace{2mm}
        \State Pass $(\bm{y}^{(obs)}, \bm{z}^{(l)})$ through the inference network in the inverse direction:
        \vspace{2mm}
        \State \qquad\quad $\bm{\theta}^{(l)} = f_{\widehat{\bm\phi}}^{-1}(\bm{z}^{(l)}; \bm{y}^{(obs)})$.
        \vspace{2mm}
    \EndFor
    \vspace{2mm}
    \State \Return $\bm{\theta}^{(1:L)}$ .
    \vspace{2mm}
    \EndFunction
    \vspace{2mm}
    \end{algorithmic}
\end{algorithm}

\subsection{Approximating Posteriors with Invertible Neural Networks}
\label{sec:inn_theory}

The transformed variable $f_{\widehat{\bm\phi}}(\bm\theta;\bm y)$ is probabilistically independent of $\bm y$ despite taking it as an argument. This is vital for the sampling scheme to function by back-transforming standard normal random variables. It suffices to note that $f_{\widehat{\bm\phi}}(\bm\theta; \bm y) = f_{\widehat{\bm\phi}}(\bm\theta; \bm y)\mid \bm y$ and hence $f_{\widehat{\bm\phi}}^{-1}(\bm z; \bm y) \equiv_{d} \bm\theta \mid \bm y$ (i.e. $f_{\widehat{\bm\phi}}^{-1}(\bm z;\bm y)$ has the same distribution as $\bm\theta\mid\bm y$) by the setup in Equation \eqref{eq:kl_2_objective} under perfect convergence. Since $\bm z\perp \bm y$ by design, so does $f_{\widehat{\bm\phi}}(\bm\theta; \bm y) \perp \bm y$.

While perfect convergence is unrealistic as there are always errors in practice, algorithms can converge close enough to such that the errors become negligible.\footnote{\citep{radev2022} specifically names Monte Carlo error and error from the invertible network that does not "accurately transform the true posterior into the prescribed Gaussian latent space", relevant to our use case.} While various universal approximation theorems exist for neural networks, which ensure that the family of neural networks is dense in the function space and suffices to approximate any function, the specifications needed are not known offhand and must be experimented with.  

\subsection{A Normal-Gamma Example}
\label{sec:NG_ABI_theory}

The results of our proof in the previous section may seem counterintuitive since $f_{\bm \phi}(\bm\theta; \bm y) \mid \bm y$ manifestly depends on $\bm y$, and yet we have found that it does not. Define $\bm\theta := \{\bm\beta, \sigma^{2}\}$. Then $\bm y \mid \bm\theta = \bm y \mid \bm\beta, \sigma^{2} \sim \mathcal{N}_{n}(\bm X\bm\beta, \sigma^{2}\bm V)$, and $\bm \beta, \sigma^{-2} \sim \mathcal{NG}_{p}(\bm m, \bm M, a_0, b_0)$, where $\bm X$ is a known covariate design matrix, $\bm m$ and $\bm M$ are defined as in Algorithm \ref{alg:KF}, and $a_0$ and $b_0$ are the shape and rate parameters. Then $\bm\beta, \sigma^{-2}\mid \bm y \sim \mathcal{NG}_{p}(\bm m_{\text{post}}, \bm M_{\text{post}}, a_{\text{post}}, b_{\text{post}})$, where
\begin{equation}
    \begin{split}
    &\bm M_{\text{post}} = (\bm X^{\T}\bm V^{-1}\bm X + \bm M^{-1})^{-1},\quad \bm \mu_{\text{post}} = \bm M_{\text{post}}(\bm X^{\T}\bm V^{-1}\bm y + \bm M^{-1}\bm m)\\
    &a_{\text{post}} = a_0 + p,\quad b_{\text{post}} = b_0 + \frac{1}{2}\left\{\bm y^{\T}\bm V^{-1}\bm y + \bm m^{\T}\bm M^{-1}\bm m - \bm m_{\text{post}}^{\T}\bm M_{\text{post}}^{-1}\bm m_{\text{post}}\right\}.
    \end{split}
\end{equation}

Rescaling $\bm\beta$ to a standard normal can be readily calculated outside the ABI framework: Let $\tilde{f}(\bm\beta,\sigma^{2}; \bm y)$ be the true transformation of $\{\bm\beta,\sigma^{2}\}$ to $\bm z \sim \mathcal{N}_{p+1}(\bm 0,\bm I)$. Then the elements of $\tilde{f}$ corresponding to $\bm\beta$ are straightforward to compute: $\tilde{f}(\bm\beta,\cdot;\bm y) = \bm L_{\text{post}}^{-1}(\bm\beta - \bm m_{\text{post}})$, where $\bm L_{\text{post}}$ is the Cholesky decomposition of the posterior covariance matrix $\bm M_{\text{post}}$. For the last element of $f$, corresponding to $\sigma^{2}$, we have $\Phi^{-1}(\Gamma(\sigma^{-2},a_{\text{post}}, b_{\text{post}}))$, where $\Gamma(\sigma^{-2}, a, b)$ is the cumulative distribution function (CDF) of a gamma random variable with shape $a$ and rate $b$ and $\Phi$ is the CDF of a standard normal random variable.

Notice that a nonlinear function is required to transform $\sigma^2$ into a standard normal. Here, a neural network with sufficient nodes, layers, and an activation function between layers (for example, a $\lceil\log_2(p+1)\rceil$-layer coupling flow where each entry of $\bm\beta$ is provided to facilitate computation of the posterior parameters) is able to approximate even nonlinear functions. Empirically testing these neural networks would then facilitate the specific choice of invertible network to obtain posterior samples from the normal-gamma.

Training via ABI aims to solve for the parameters $\bm\phi$ so that the inference network $f_{\bm\phi}$ best approximates $\tilde{f}$. Begin by generating synthetic data samples from the model, i.e. $\bm y^{(m)} \mid \bm\beta^{(m)}, \sigma^{2,(m)} \sim \mathcal{N}_{n}(\bm X\bm\beta^{(m)}, \sigma^{2,(m)}\bm V)$ and $\bm\beta^{(m)}, \sigma^{-2,(m)} \sim \mathcal{NG}_{p}(\bm m, \bm M, a_0, b_0)$, and that we correctly specify $\bm V$ and $\bm C$. Then at training, for each Monte Carlo iteration $m=1,\ldots,M$, $\bm\beta^{(m)}, \sigma^{-2,(m)}$ are sampled from the normal-gamma prior. Next, $\bm\beta^{(m)}, \sigma^{-2,(m)}$, and $\bm y^{(m)}$ are passed to the individual summand term in Equation \eqref{eq:bf_objfunc} to supply the integral expression with individual Monte Carlo terms. At the end of the training cycle, $\bm\phi$ is updated via backpropagation. The next training cycle then repeats with the sampling of new priors and synthetic data.

By design, $\mathbb{E}[f_{\bm\phi}(\bm\beta, \sigma^{2}; \bm y)] = \bm 0$ and $\mathrm{Var}[f_{\bm\phi}(\bm\beta, \sigma^{2};\bm y)] = \bm I$. Since the distribution of the scaled variable $f_{\bm\phi}(\bm\beta, \sigma^{2}; \bm y)$ is also normal, the mean and variance are sufficient to fully parametrize the distribution, and despite the scaled quantity $f_{\bm\phi}(\bm\beta, \sigma^{2};\bm y)$ taking in the data $\bm y$ as an argument, the resulting variable transformed by the normalizing flow does not ultimately depend on the data. Consequently, to produce a posterior from the randomly sampled standard normal $\bm z \sim \mathcal{N}_{p}(\bm 0, \bm I)$, $f_{\bm\phi}^{-1}(\bm z; \bm y)$ takes in $\bm y$ as an argument, samples $\bm z$, and the transforms $\bm z$ into a posterior sample $\bm\beta, \sigma^2\mid \bm y$. Extensions across time, which will be relevant from the next section forward, can be achieved by appending the $\bm\beta_t$'s and $\bm y_t$'s vertically and applying the procedure in this section the same way.

\section{Dynamic Linear Model}
\label{sec:DLM}

Having covered the theory behind amortized inference, we return to discussing the time-varying model needed for the actigraph data in Section \ref{sec:actigraph_data}. Specifically, we utilize a time-varying geospatial model:
\begin{equation}
\label{eq:dlm}
    \begin{split}
        \bm{y}_{t} &= \bm{X}_{t}\bm{\beta}_{t} + \bm{\nu}_{t}\text{;}\quad\quad\  \bm{\nu}_{t} \overset{ind}{\sim} \mathcal{N}_{n_{t}}(0, \sigma^{2}\bm{V}_{t})\\
        \bm{\beta}_{t}&= \bm{G}_{t}\bm{\beta}_{t-1} + \bm{\omega}_{t}\text{;}\quad\ \bm{\omega}_{t} \overset{ind}{\sim} \mathcal{N}_{p}(0, \sigma^{2}\bm{W}_{t})\\
        \bm{\beta}_{0} &\sim \mathcal{N}_{p}(\bm{m}_{0}, \sigma^{2}\bm{M}_{0})\text{;}\quad\ \sigma^{-2} \sim \mathcal{G}(a_{0},b_{0})
    \end{split}
\end{equation}
where $\mathcal{G}(a,b)$ is the gamma distribution with parameters $a$ and $b$. $a_{0}$, $b_{0}$, $\bm{m}_{0}$, and $\bm{M}_{0}$ are hyperparameters for their respective distributions. $\bm\beta_t$ and $\sigma^2$ retain their definitions from the Actigraph Timesheet setup in Equation \eqref{eq:AT_norm_at_t}, though not the structure in the normal-normal example.

Equation \eqref{eq:dlm} is called the \textit{dynamic linear model} (DLM), due to its linear terms and properties and temporal dependence being a key application for the model's hierarchical structure \citep{west_harrison}. The linear property makes the DLM relatively easy to understand and implement and the Markovian property enables inference to resume from the latest time point instead of the entire history of the fit. The DLM has the Forward Filter Backwards-Sampling algorithm (FFBS) as a well-established solution to acquire the posterior distribution of $\bm{\beta}_{1:T},\sigma^2\mid \bm y_{1:T}$ (Algorithm \ref{alg:FFBS}). The distributions in Equation \eqref{eq:dlm} also set up the distributions of the individual terms at each level of the model with the desired property of conjugacy: each term of the sequence of distributions remains in the same family of distributions across time, differing only by their particular parameters \citep{carter_kohn}. Our ensuing development makes use of accessible distribution theory for implementing FFBS for gaussian models \citep[see for example][for technical details]{west_harrison, petris_etal, 2025BayesianModelingMechSystems}.

\subsection{Forward Filter}
\label{sec:ff}

\begin{algorithm}
    \caption{Forward Filter Backwards Sampling algorithm}\label{alg:FFBS}
    \begin{algorithmic}[1]
    \State \textbf{Input:} Data $\bm y_{1:T}$ and Kalman filter starting values $a_0$, $b_0$, $\bm{m}_0$, $\bm{M}_0$
    \vspace{2mm}
    \State {\color{white}\textbf{Input:}} Observation and state transition matrices $\bm{X}_{1:T}$ and $\bm{G}_{1:T}$
    \vspace{2mm}
    \State {\color{white}\textbf{Input:}} Correlation matrices $\bm{V}_{1:T}$ and $\bm{W}_{1:T}$
    \vspace{2mm}
    \State \textbf{Output:} Sample from posterior $p(\bm\beta_{1:T},\sigma^{2}|\bm{y}_{1:T})$
    \vspace{2mm}
    \Function{\texttt{FFBS}}{$a_0, b_0, \bm y_{1:T}, \bm X_{1:T}, \bm G_{1:T}, \bm V_{1:T}, \bm W_{1:T}$}
    \vspace{2mm}
    \State $\{a_t, b_t,\bm{c}_t,\bm{C}_t,\bm{m}_t,\bm{M}_t\}_{t=1}^T\gets \texttt{Filter}(\bm y_{1:T}, a_0, b_0,\bm m_0, \bm M_0, \bm G_{1:T}, \bm X_{1:T}, \bm V_{1:T},\bm W_{1:T}$)
    \vspace{2mm}
    \State $\{\bm\beta_{1:T},\sigma^{2}\}\gets \texttt{BackwardSample}(a_T, b_T, \{\bm{c}_t,\bm{C}_t,\bm{m}_t,\bm{M}_t, \bm G_t\}_{t=1}^T$)
    \vspace{2mm}
    \State \Return $\{\bm\beta_{1:T},\sigma^{2}\}$ \Comment{Return sample from $p(\bm\beta_{1:T},\sigma^{2}|\bm y_{1:T})$}
    \vspace{2mm}
    \EndFunction
\end{algorithmic}
\end{algorithm}

The FFBS algorithm (Algorithm \ref{alg:FFBS}) consists of two steps: the forward filter (FF) and backwards sampling (BS) algorithms. The forward filter derives its name from the Kalman filter, a Markovian model that takes in the data sequentially and updates the parameters at the next time step according to the data and existing parameter estimate. The density we wish to sample from is:
\begin{equation}
    \label{eq:ff}
    \begin{split}
        p(\bm{\beta}_{1:T}, \sigma^{2}\vert \bm y_{1:T}) \propto p(\sigma^{2}\vert a_{0}, b_{0}, \bm y_{1:T})p(\bm{\beta}_{T}\vert \bm y_{1:T}, \sigma^{2})\prod_{t=0}^{T-1}p(\bm{\beta}_{t}\vert \bm y_{1:t},\bm{\beta}_{t+1},\sigma^{2}).
    \end{split}
\end{equation}

\begin{algorithm}[t]
    \caption{Kalman (forward) filter}\label{alg:KF}
    \begin{algorithmic}[1]
    \State \textbf{Input:} Data $\bm y_{1:T}$, hyperparameters $a_0$, $b_0$, $\bm{m}_0$, $\bm{M}_0$
    \vspace{2mm}
    \State {\color{white}\textbf{Input:}} Observation and state transition matrices $\bm{X}_{1:T}$ and $\bm{G}_{1:T}$
    \vspace{2mm}
    \State {\color{white}\textbf{Input:}} Correlation matrices $\bm{V}_{1:T}$, $\bm{W}_{1:T}$
    \vspace{2mm}
    \State \textbf{Output:} Filtering distribution parameters at time $t=1,\ldots, T$
    \vspace{2mm}
    \Function{\texttt{Filter}}{$\bm y_{1:T}, a_0, b_0,\bm m_0, \bm M_0,\bm G_{1:T},\bm X_{1:T}, \bm V_{1:T}, \bm W_{1:T}$}
    \vspace{2mm}
    \For{$t=1$ to $T$}
        \vspace{2mm}
        \State \textit{\# Compute prior distribution} $p(\bm\beta_t, \sigma^{-2} \mid \bm{y}_{1:t-1})\sim \mathcal{NG}(\bm{c}_t, \bm{C}_t, a_t^*, b_t^*)$:
        \vspace{2mm}
        \State  $\bm c_t\gets \bm G_t\bm m_{t-1}, \ \bm C_t\gets \bm G_t\bm M_{t-1}\bm G_t^\T+\bm W_t$
        \vspace{2mm}
        \State $a_t^*\gets a_{t-1}, b_t^*\gets b_{t-1}$
        \vspace{2mm}
        \State \textit{\# Compute one-step-ahead forecast}   $p(\bm{y}_t|\bm{y}_{1:t-1})\sim\mathcal{T}_{2a_t^*}(\bm{q}_t, \frac{b_t^*}{a_t^*}\bm{Q}_t)$
        \vspace{2mm}
        \State  $\bm q_t\gets \bm{X}_t\bm{c}_t, \ \boldsymbol Q_t\gets\boldsymbol{X}_t\boldsymbol{C}_t\boldsymbol{X}_t^\T+\boldsymbol V_t$
        \vspace{2mm}
        \State \textit{\# Compute filtering distribution} $p(\bm{\beta}_{t}, \sigma^{-2} \mid \bm{y}_{1:t})\sim\mathcal{NG}(\bm{m}_t,\bm M_t, a_t, b_t):$
        \vspace{2mm}
        \State  $\bm m_t\gets\bm{c}_t+\bm{C}_t\bm{X}_t^\T\bm{Q}_t^{-1}(\bm{y}_{t}-\bm{q}_t), \ \bm M_t\gets\bm{C}_t - \bm{C}_t\bm{X}_t^\T\bm{Q}_t^{-1}\bm{X}_t\bm{C}_t^\T$
        \vspace{2mm}
        \State $a_t\gets a_t^* + \frac{n_t}{2}, \ b_t \gets b_t^* + \frac{1}{2}(\bm{y}_t-\bm{q}_t)^\T\bm{Q}_t^{-1}(\bm{y}_t-\bm{q}_t)$
        \vspace{2mm}
    \EndFor
    \vspace{2mm}
    \State \Return $\left\{a_t, b_t,\bm{c}_t,\bm{C}_t,\bm{m}_t,\bm{M}_t\right\}_{t=1}^T$
    \vspace{2mm}
    \EndFunction
    \end{algorithmic}
\end{algorithm}

In practice, we opt to sample from $p(\bm \beta_t\mid \bm y_{1:T},\sigma^2)$ to acquire full posteriors from the product series by marginalizing out $\bm\beta_{t+1}$ from each term in the product of Equation \eqref{eq:ff} and bringing the $\bm y_{(t+1):T}$ into parameter calculations for computational convenience and to lower storage requirements for the samples.

We proceed to compute the parameters at each time step using the procedure outlined in Algorithm \ref{alg:KF}. Due to the sequential nature of the updates of $a_t$ and $b_t$ in line 14, we can rewrite $a_T$ and $b_T$ may be written in compact form:
\begin{equation}
    \label{eq:ffsigsol}
    \begin{split}
        a_{T} = a_{0} + \frac{1}{2}\sum_{t=1}^{T}n_{t}\quad\mbox{, and }\quad
        b_{T} = b_{0} + \frac{1}{2}\sum_{t=1}^{T}(\bm{y}_{t} - \bm{q}_{t})^{\T}\bm{Q}_{t}^{-1}(\bm{y}_{t} - \bm{q}_{t}).
    \end{split}
\end{equation}

Ordinarily, the parameters from the FF are not used for sampling; the FF's main purpose is to compute the parameters for the BS to enable full posterior sampling. However, it is sometimes desirable to sample from the FF as a benchmark for comparison with other algorithms.

\subsection{Backwards Sampling}
\label{sec:bs}

Backwards sampling obtains samples of each $\bm{\beta}_{t}\vert \bm y_{1:T}$ across time $t$. The posterior mean and variance for $\bm{\beta}_{t}$ are further refined with the data across all time; the coefficients from the FF are used to compute the parameters for smoothing. As with the FF, BS also takes advantage of the conjugacy of the underlying distributions to efficiently compute the underlying parameters for the full posterior samples. The procedure is outlined in Algorithm \ref{alg:BS}.\footnote{$\bm{s}_{T} = \bm{m}_{T}$ and $\bm{S}_{T} = \bm{M}_{T}$, since the parameters for $p(\bm{\beta}_{T}\vert \sigma^{2},\bm y_{1:T})$ are given to us by the FF.}

\begin{algorithm}[t]
    \caption{Backward Sampler}\label{alg:BS}
    \begin{algorithmic}[1]
    \State \textbf{Input:} Filtering parameters and inputs from Algorithm \ref{alg:KF}
    \vspace{2mm}
    \State \textbf{Output:} Posterior sample from $p(\bm\beta_{1:T},\sigma^{2}\mid \bm y_{1:T})$
    \vspace{2mm}
    \Function{\texttt{BackwardSample}}{$a_T, b_T, \left\{\bm{c}_t,\bm{C}_t,\bm{m}_t,\bm{M}_t, \bm G_t\right\}_{t=1}^T$}
    \vspace{2mm} 
    \State Draw $\sigma^{-2}\sim\mathcal{G}(a_T, b_T)$
    \vspace{2mm}
    \State Draw $\bm\beta_T\sim\mathcal{N}(\bm{m}_T, \sigma^{2}\bm M_T)$
    \vspace{2mm}
    \For{$t=T-1$ to $1$}
    \vspace{2mm}
    \State $\bm s_t \gets \bm{m}_t+\bm{M}_t\bm G_{t+1}^\T\bm{C}_{t+1}^{-1}\left(\bm{s}_{t+1}-\bm{c}_{t+1}\right)$
    \vspace{2mm}
    \State $\bm{S}_t \gets \bm{M}_t-\bm{M}_t\bm G_{t+1}^\T\bm{C}_{t+1}^{-1}\left(\bm{C}_{t+1}-\bm{S}_{t+1}\right)\bm{C}_{t+1}^{-1}\bm G_{t+1}\bm{M}_t$
    \vspace{2mm}
    \State Draw $\bm\beta_t\sim\mathcal{N}(\bm{s}_t, \sigma^{2}\bm S_t)$
    \vspace{2mm}
    \EndFor
    \vspace{2mm}
    \State \Return $\{\bm\beta_{1:T}, \sigma^{2}\}$
    \vspace{2mm}
    \EndFunction
\end{algorithmic}
\end{algorithm}

The normal density of $\bm{\beta}_{t}\vert \sigma^{2}, \bm y_{1:T}$ and gamma density of $\sigma^{2}\vert \bm y_{1:T}$ can be combined into a single normal-gamma density:

\begin{equation}
    \label{eq:smooth_nig}
    \bm{\beta}_t,\sigma^{-2}\vert \bm y_{1:T} \sim \mathcal{NG}(\bm{s}_{t}, \bm{S}_{t}, a_{T}, b_{T}).
\end{equation}
Each sample from the normal gamma involves sampling $\sigma^{-2} \sim \mathcal{G}(a_{T},b_{T})$ and then $\bm{\beta}_t \sim \mathcal{N}_{p}(\bm{s}_{t}, \sigma^{2}\bm{S}_{t})$. Furthermore, if we integrate out $\sigma^{2}\vert \bm y_{1:T}$ from Equation \eqref{eq:smooth_nig}, we obtain:

\begin{equation}
    \label{eq:smooth_t}
    \bm{\beta}_{t} \vert \bm y_{1:T} \sim \mathcal{T}_{2a_{T}}\left(\bm{s}_{t},\frac{b_{T}}{a_{T}}\bm{S}_{t}\right)
\end{equation}
where $\mathcal{T}_{2a_{T}}(\bm{s}_{t},\frac{b_{T}}{a_{T}}\bm{S}_{t})$ is a multivariate Student's t-distribution with degrees of freedom $2a_T$, mean $\bm{s}_t$ and scale matrix $\frac{b_{T}}{a_{T}}\bm{S}_{t}$. We utilize Equation \eqref{eq:smooth_t} to analytically compute credible intervals at each time point without sampling $\sigma^2\mid \bm y_{1:T}$.

\begin{algorithm}
    \caption{Generate Synthetic Outcome from DLM}\label{alg:y_from_DLM}
    \begin{algorithmic}[1]
        \State \textbf{Input:} Sample size $L$
        \vspace{2mm}
        \State \qquad \quad All parameters listed in Algorithm \ref{alg:KF} except for data $\bm y_{1:T}$
        \vspace{2mm}
        \State \textbf{Output:} A single set of parameters from the DLM.
        \vspace{2mm}
        \State \qquad\quad\ \  A sample of $L$ synthetic outcomes $\bm y_{1:T}^{(1:L)}$ following the DLM given a single set of parameters.
        \vspace{2mm}
        \Function{\texttt{DLM\_Prior}}{$a_0, b_0, \bm m_0, \bm M_0, \bm G_{1:T}, \bm W_{1:T}$}
            \vspace{2mm}
            \State Draw $\sigma^{-2} \sim \mathcal{G}(a_0, b_0)$
            \vspace{2mm}
            \State Draw $\bm\beta_0 \sim \mathcal{N}(\bm m_0,\sigma^2 \bm M_0)$
            \vspace{2mm}
            \For{$t = 1$ to $T$}
                \vspace{2mm}
                \State Draw $\bm\beta_t \sim \mathcal{N}(\bm G_t \bm \beta_{t-1},\sigma^2 \bm W_t)$
                \vspace{2mm}
            \EndFor
            \vspace{2mm}
            \State\Return $\left\{\bm\beta_{1:T}, \sigma^2\right\}$
            \vspace{2mm}
        \EndFunction
        \vspace{2mm}
        \State 
        \vspace{2mm}
        \Function{\texttt{Y\_From\_DLM}}{$\bm\beta_{1:T}, \sigma^2, \bm X_{1:T}, \bm V_{1:T}, L$}
        \vspace{2mm}
        \For{$l=1$ to $L$ and $t=1$ to $T$}
            \vspace{2mm}
            \State Draw $\bm{y}_t^{(l)} \sim \mathcal{N}(\bm X_t\bm\beta_t, \sigma^{2}\bm V_t)$
            \vspace{2mm}
        \EndFor
        \vspace{2mm}
        \State \Return $\bm y_{1:T}^{(1:L)}$
        \vspace{2mm}
        \EndFunction
    \end{algorithmic}
\end{algorithm}

\section{Illustrations for ABI}
\label{sec:illustrations}

\subsection{Stationary Normal-Gamma}
\label{sec:normflow_normal_gamma}

\begin{figure}[b]
    \centering
    \includegraphics[width=0.8\linewidth]{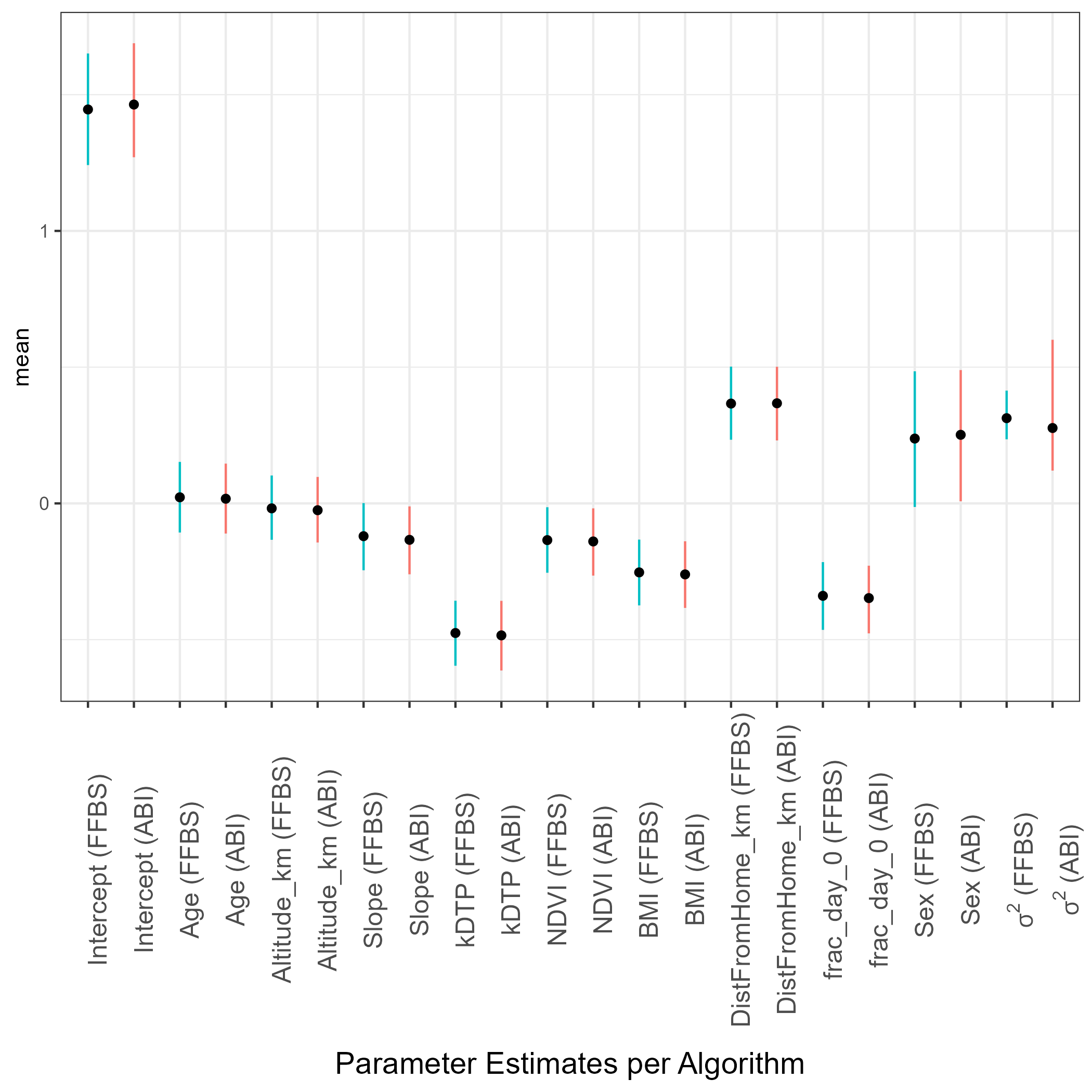}
    \caption{95\% credible intervals of the estimates of $\bm\beta_1, \sigma^2\mid \bm y_1$ from \texttt{BayesFlow} (red) and the FFBS (cyan). The means and intervals differ slightly for all parameter estimates except $\sigma^2$, which differs considerably.}
    \label{fig:betat_CI}
\end{figure}

To test the theoretical underpinnings of our results, we use \texttt{DLM\_PRIOR} and \texttt{Y\_FROM\_DLM} from Algorithm \ref{alg:y_from_DLM} to generate our ground truth parameters $\bm\beta_{\text{true}},\sigma^2_{\text{true}}$, and data $\bm y_{\text{true}}$, specifying $T = 1$ and $L = 1$. We normalize the covariates in the actigraph data at $t=1$ to have zero mean and unit variance to form $\bm X$; scaling the covariates is essential to facilitate convergence of machine learning algorithms. We utilize the \texttt{BayesFlow} (version 2.0.6) \citep{radev2022} software package to run our simulations because it contains existing code needed to run the ABI siulation, and set $f_{\bm\phi}$ to be a coupling flow neural network with 4 invertible layers with 128-length single-layer perceptrons acting as sub-networks within each layer.

Figure \ref{fig:betat_CI} shows ABI's ability to learn the posteriors of the normal-gamma from running Algorithm \ref{alg:ABI}. Almost all the parameters (sans $\sigma^2\mid \bm y$), the differences in the means of the distributions are small relative to the overall scale spanned by the individual entries of $\bm\beta$. The model manages to condense information from $\bm y$ and demonstrates significant ability to capture the normal-gamma model even within $N_{ITER}=5,000$ online training cycles and a decently small batch size of $M=32$.

\subsection{Actigraph Data}
\label{sec:actigraph_illustrations}

We next extend our approach in section \ref{sec:normflow_normal_gamma} to $T=61$. To facilitate this extension, we retain Equation \eqref{eq:dlm} as a prior model to generate the synthetic data to be learned by the amortizer, implemented as Algorithm \ref{alg:y_from_DLM}. As with the previous subsection, we set $N_{ITER}=5,000$ and a small batch size of $M = 32$. 

To test our approach and run a feasible subset of our data, we subset the actigraph data to the first trajectory for all 92 subjects that contains a MAG measurement greater than 0.1, the threshold for moderate exercise defined by \citep[Supplement]{actigraph}. We test the run on a synthetic activity measure $\bm y_{1:T}$ sampled using Algorithm \ref{alg:y_from_DLM} with $L = 1$, $a_0=3$, $b_0=1$, $\bm m_0=\bm 0$, $\bm M_0 = \bm I_p$, $\bm G_{t} = \bm I_p$, $\bm W_{t} = \bm I_p$, and $\bm V_{t} = \bm I_{n_t}$ for $t=1,\ldots,T$: $\bm{M}_0$ is set to identity to correspond to the prior claim that the effects of the covariates on fitness level are not correlated with one another; $\bm V_{t} = \bm I_{n_t}$ for all $t$ is justified by the expected lack of correlation between different subjects walking or running their individual trajectories at different times, assuming that none of them went running with any of the other subjects. As with the stationary normal-gamma model, we facilitate fast convergence of the training algorithm by separately scaling each column of the covariates $\bm X_t$ for each $t$ so that each column is centered around zero with unit variances for each time point $t$ before running Algorithm \ref{alg:hier_ABI_DLM_timebatch}.\footnote{While separate scaling for each $t$ would lead to different interpretations of $\bm\beta_t$ across time $t$, capturing the original common interpretation of each $\bm\beta_t$ can be obtained by rescaling $\bm X_t$ to its original scale.}

We also divide up the dataset into $B=45$ sets of time intervals which span the set of time steps expressed by our dataset to further facilitate computational efficiency and stability: 41 temporal singletons for $t=1,\ldots,41$, and then 5-time step intervals up to $t = 61$ to group together the sparsest trajectories for the end of analysis. The idea is that for each separate $b=1,\ldots,B$, we separately fit neural networks to subsets of the data - both corresponding to earlier time intervals and using priors encompassing the simulations in the earlier subset to supply the priors for the later subset. We set $f_{\bm\phi}$ to be a coupling flow neural network with 4 invertible layers for the temporal singletons and 6 layers for the last four blocks of 5 time steps, each layer comprised of 128-length single-layer perceptrons acting as sub-networks within the invertible layers to sufficiently capture the minimum required number of parameters. For $b > 1$, we sample from a prior informed by the theoretical FF parameters where the outcome $\bm y_{1:T}$ is generated from the DLM with the parameters specified in this subsection. The theoretical derivation of the parameters for the appropriate priors is detailed in Section \ref{sec:bridge_priors}. 7.52 hours were spent training the ABI system under these settings.

\begin{figure}[p]
    \centering
    \subfloat[Age]{
        \label{fig:synth_Age}
        \includegraphics[width=4cm]{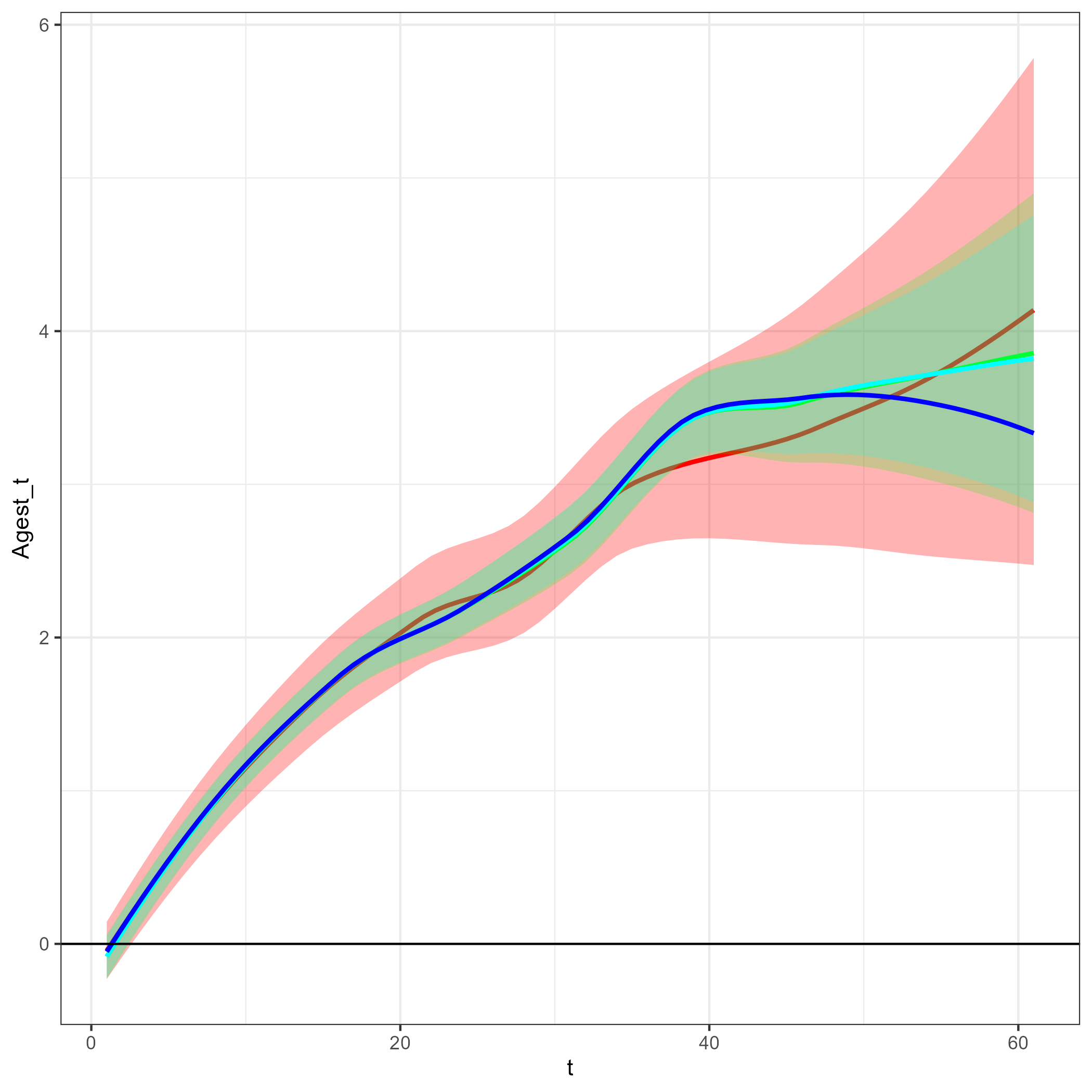}
    }
    \subfloat[Alt.]{
        \label{fig:synth_Altitude}
        \includegraphics[width=4cm]{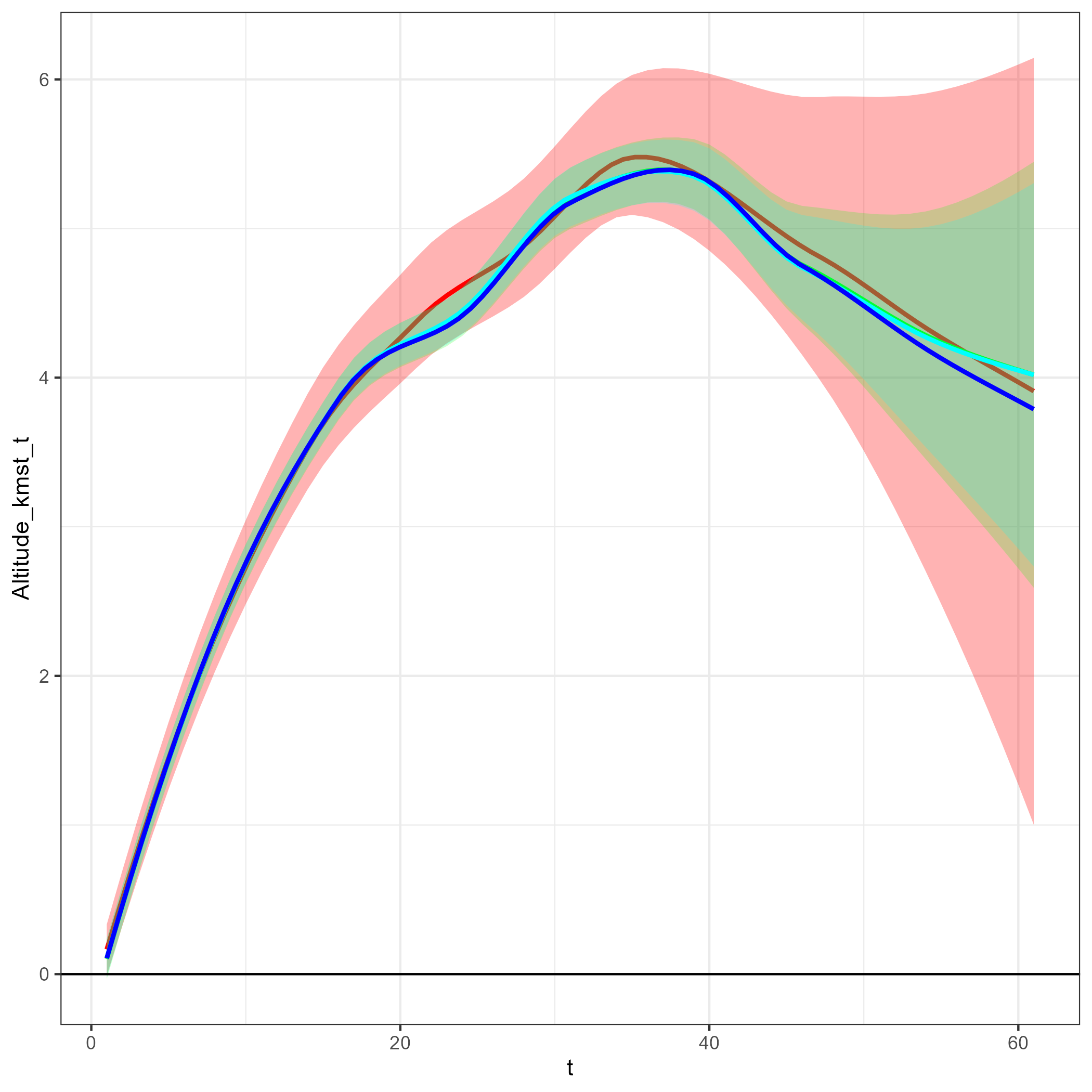}
        
    }
    \subfloat[BMI]{
        \label{fig:synth_BMI}
        \includegraphics[width=4cm]{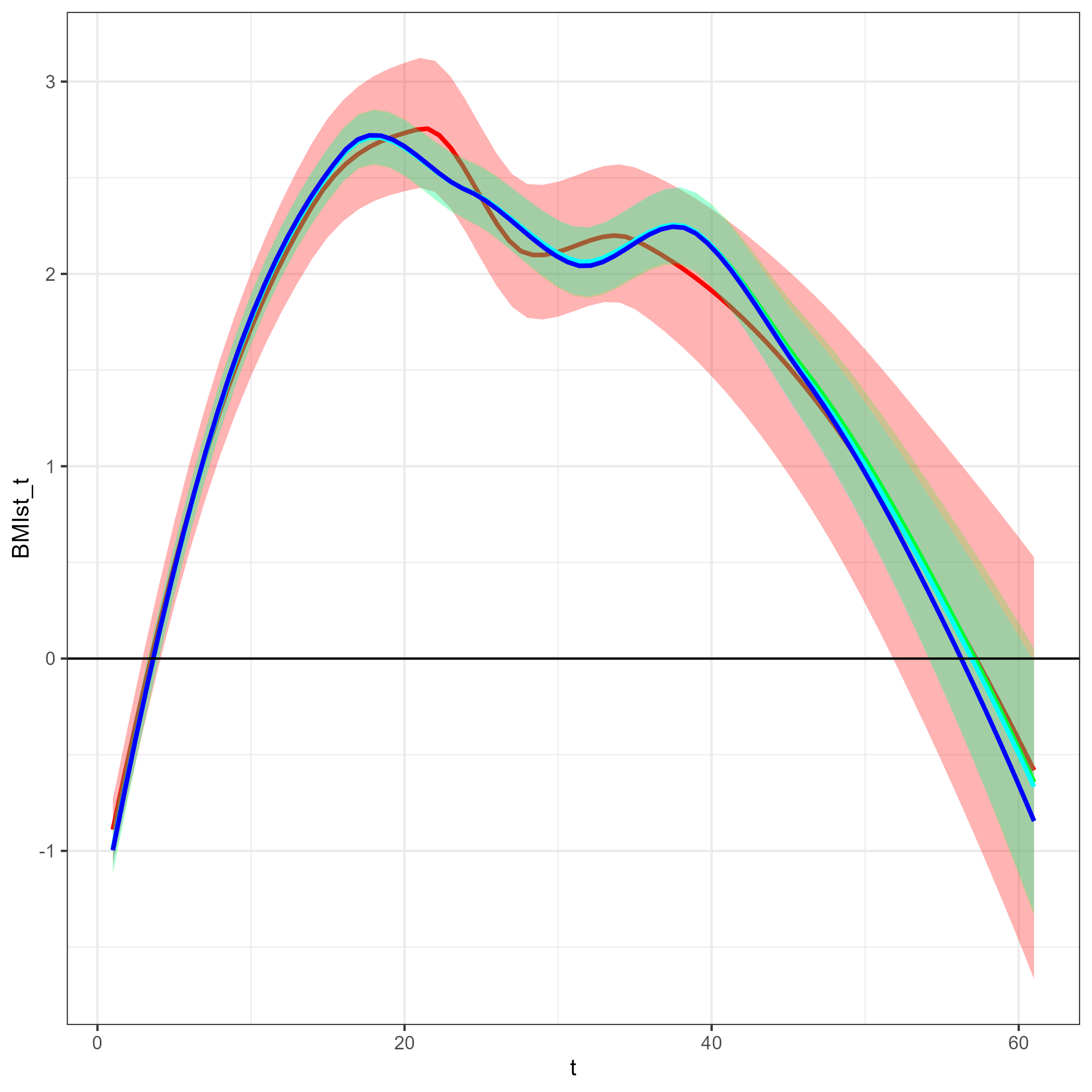}
    }\\
    \subfloat[Dist. from Home (km)]{
        \label{fig:synth_DistFromHome_km}
        \includegraphics[width=4cm]{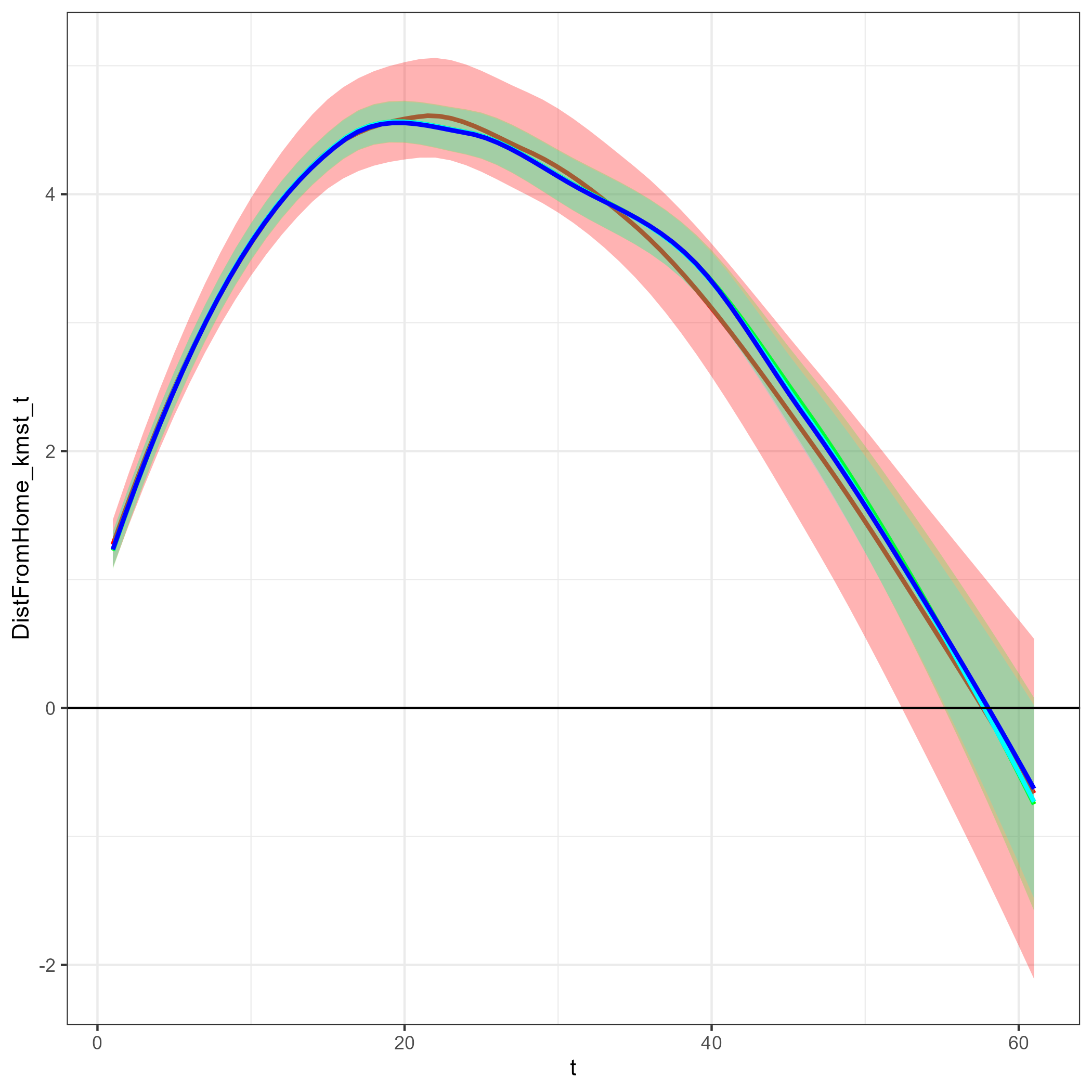}
    }
    \subfloat[Dist. to Parks (km)]{
        \label{fig:synth_kDTP}
        \includegraphics[width=4cm]{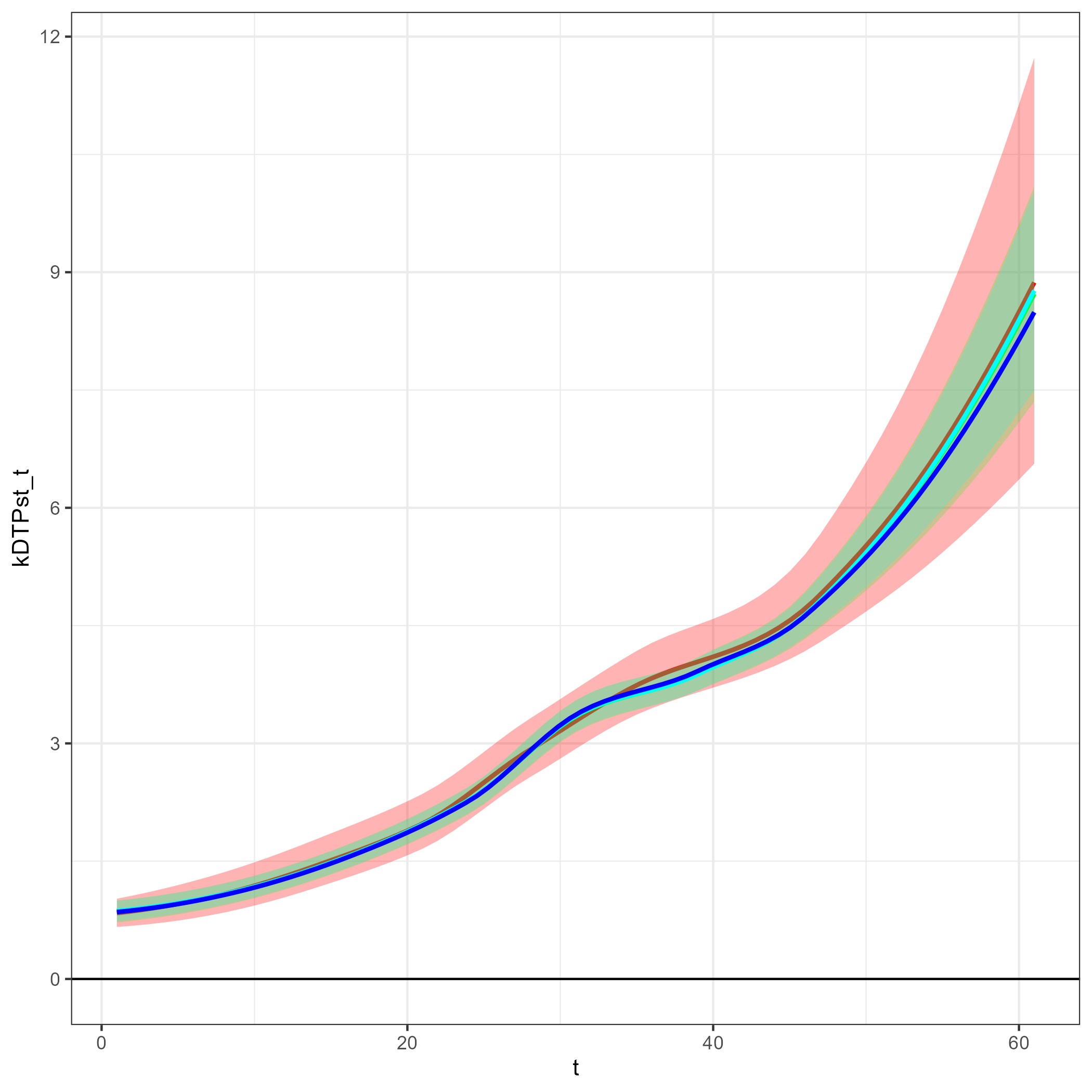}
    }
    \subfloat[Start Time in Day]{
        \label{fig:synth_frac_day_0}
        \includegraphics[width=4cm]{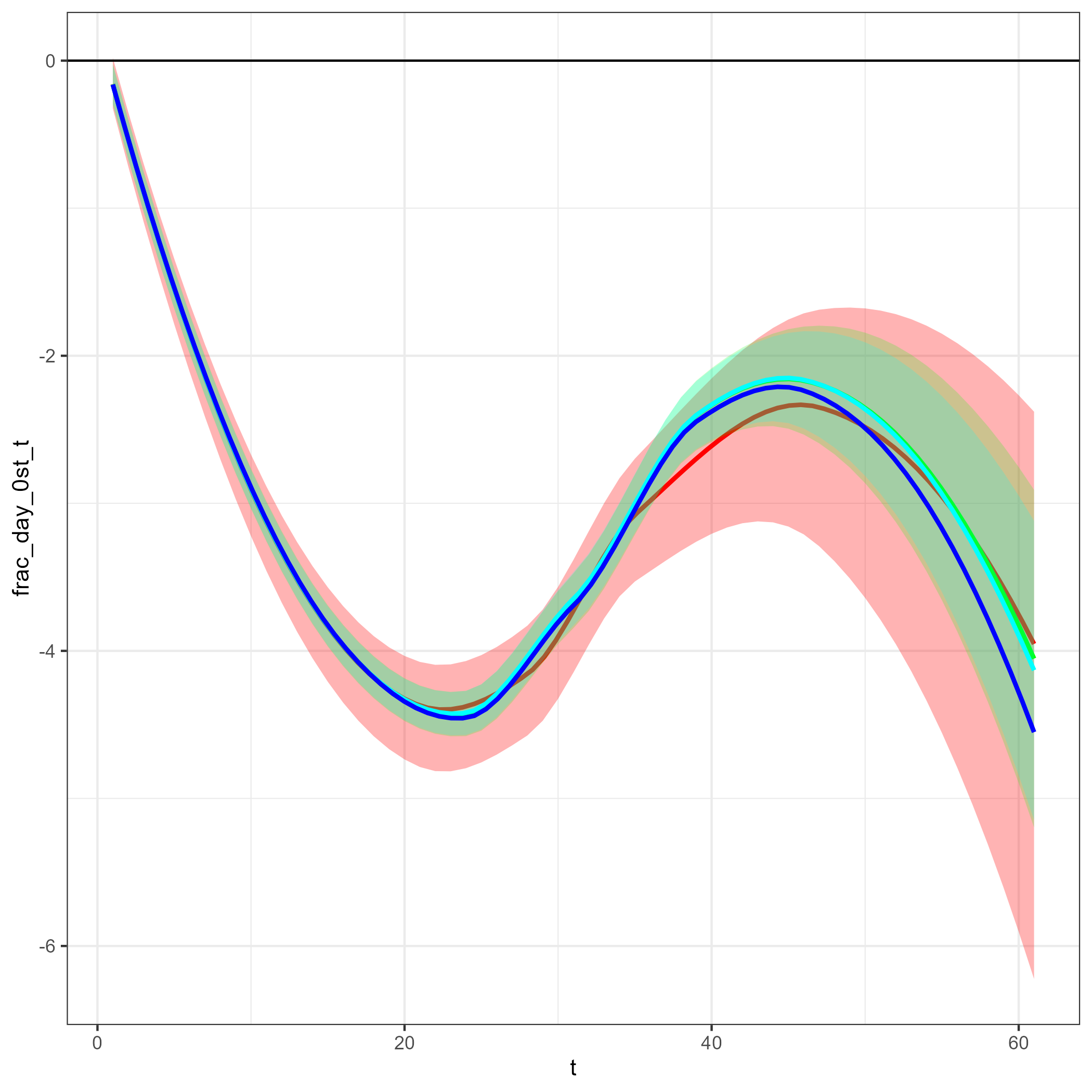}
    }\\
    \subfloat[NDVI]{
        \label{fig:synth_NDVI}
        \includegraphics[width=4cm]{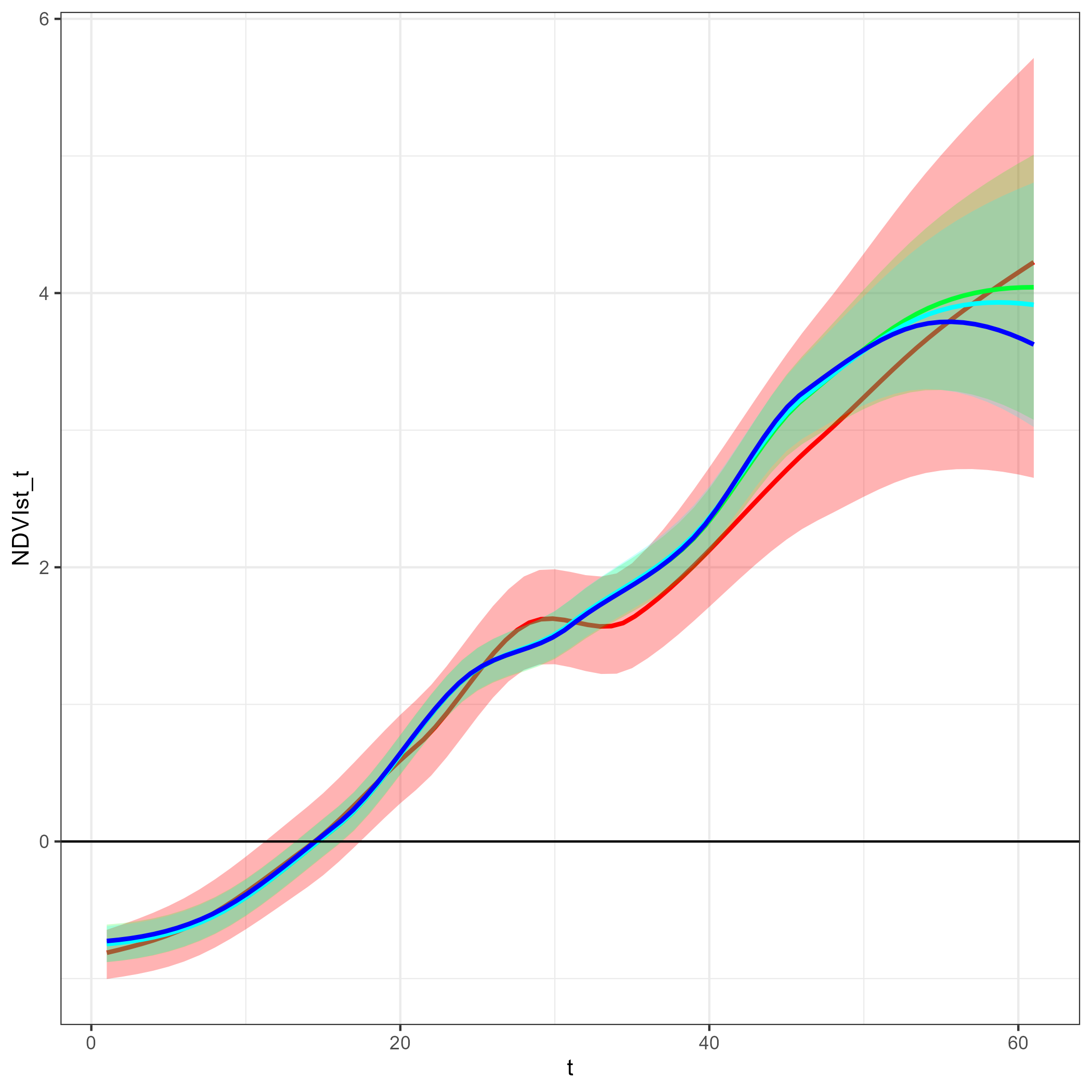}
    }
    \subfloat[Sex]{
        \label{fig:synth_Sex}
        \includegraphics[width=4cm]{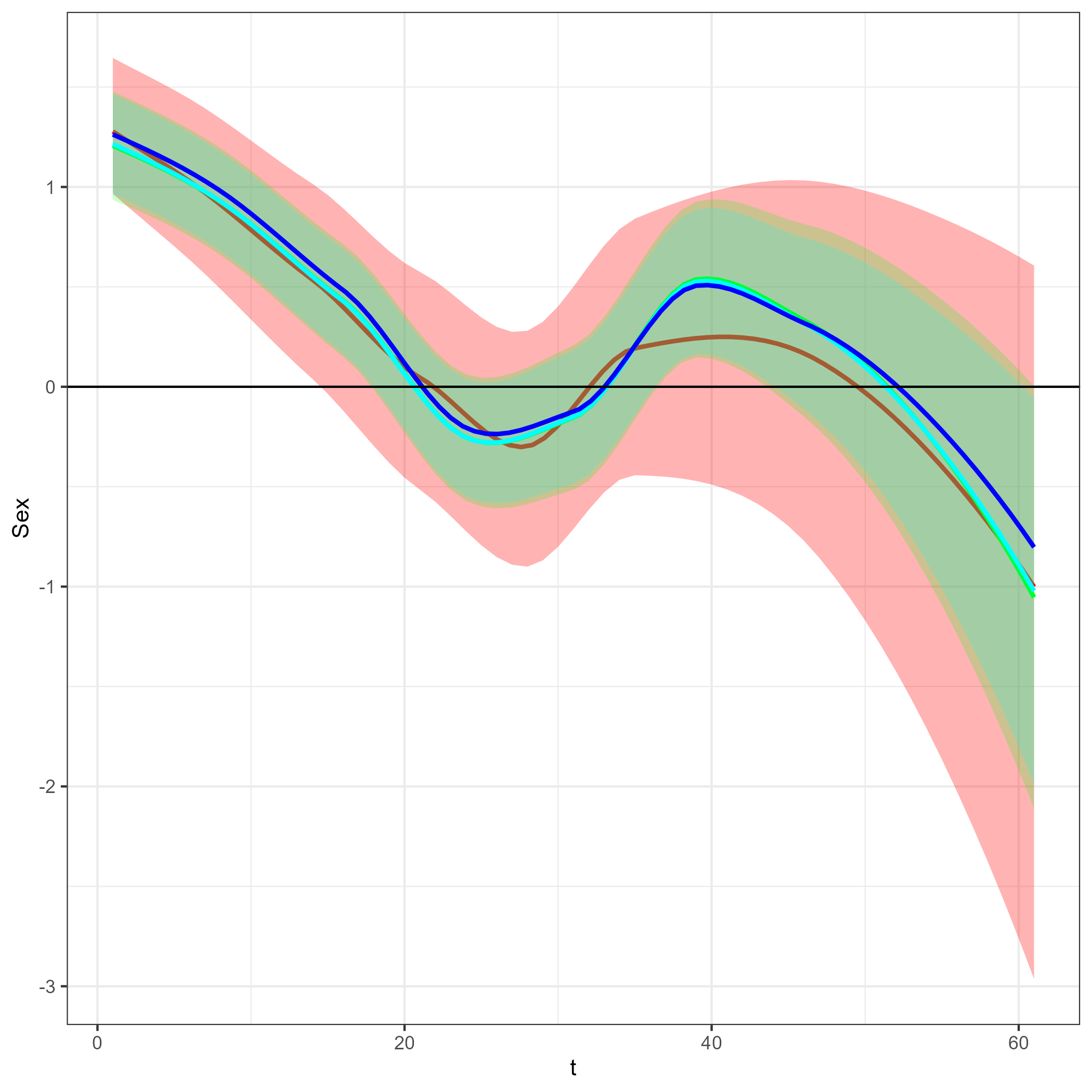}
    }
    \subfloat[Slope]{
        \label{fig:synth_Slope}
        \includegraphics[width=4cm]{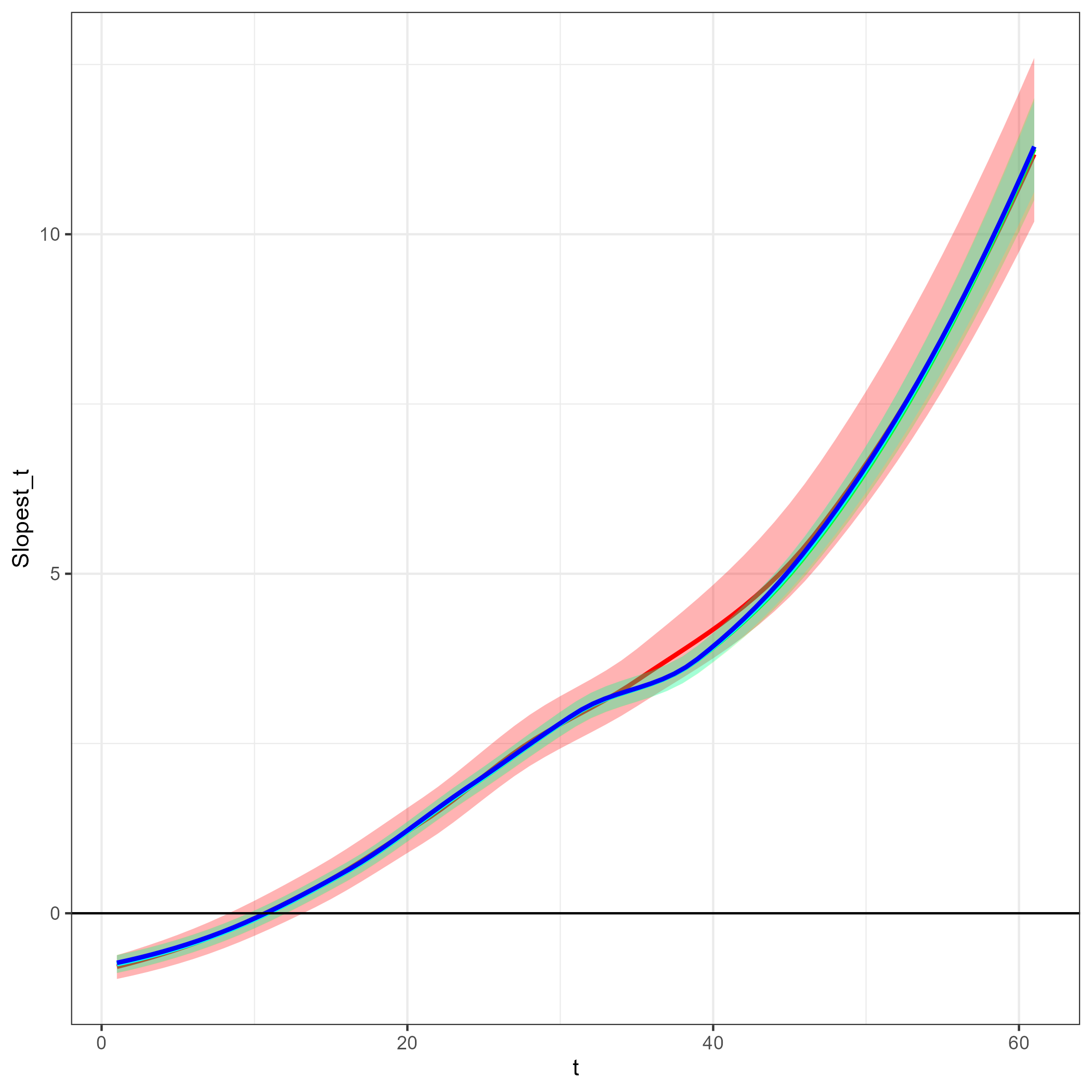}
    }
    
    \caption{The estimated parameter trajectories (red) due to Algorithm \ref{alg:hier_ABI_DLM_timebatch}, estimated trajectories due to the FFBS (cyan), the FF only (green), and true trajectories (blue) of the coefficients used to generate the synthetic outcomes at each time step. The true coverage rate of the synthetic outcomes after the FFBS processes the outcome is about 97.6\%. The credible intervals for the ABI output were taken from the 2.5\% and 97.5\% quantiles of 10,000 samples generated from the trained networks. A horizontal line is added at $\bm\beta_{t,j} = 0$ for non-intercept elements $j$ to visualize the statistical significance of each parameter over time.} 
    \label{fig:synth_plots}
\end{figure}

The credible intervals of nine of the covariates across time are displayed in Figure \ref{fig:synth_plots}: while generally wider, the ABI intervals manage to capture the trend of the FFBS and true parameters. Additional precision may be obtained by rerunning the network with a greater batch size $M$, which adds more Monte Carlo samples to approximate the loss but with a proportionately greater time cost.

Note that the model fitted using Algorithm \ref{alg:hier_ABI_DLM_timebatch} learns the posteriors of a slightly different model than that of Equation \eqref{eq:dlm}: the $\sigma^2$ generated at each time step is allowed to differ from the others for each batch rather than being strictly shared across time (though the shape and rate used to generate it remain the same throughout), and information from the output of future time steps does not make it back to the parameters of past time steps. However, the model's results are close enough to that of the DLM and its fitted parameters are close enough to those generated by the FFBS that it is apparent that the Algorithm \ref{alg:hier_ABI_DLM_timebatch} manages to learn the DLM and the parameters of the FFBS with only slightly wider credible interval bands. This is apparent through Figure \ref{fig:synth_plots}, as well as through the figures in further sections.

\begin{algorithm}
    \caption{Hierarchical ABI on the Dynamic Linear Model with Time Intervals}\label{alg:hier_ABI_DLM_timebatch}
    \begin{algorithmic}[1]
    \vspace{2mm}
    \State \textbf{Input:} A set of $B$ predefined time intervals $\{[T_{b-1},T_{b}]\}_{b=1}^{B}$
    \vspace{2mm}
    \State \qquad \quad All parameters listed in Algorithm \ref{alg:KF} except for data $\bm y_{1:T}$
    \vspace{2mm}
    \State \qquad\quad Batch size $M$.
    \vspace{2mm}
    \For{$b=1,\ldots,B$}
    \vspace{2mm}
    \State Train neural networks for block $b$ (corresponding to $[T_{b-1}, T_{b}]$): 
    \vspace{2mm}
    \State \texttt{Train\_BayesFlow}(\texttt{DLM\_Prior}$(a_0, b_0, \bm m_0, \bm M_0 + (T_{b-1}-1)\bm I_p, \bm G_{T_{b-1}:T_{b}}, \bm W_{T_{b-1}:T_{b}})$, 
    \vspace{2mm}
    \State \qquad\qquad\qquad\qquad\quad\  \texttt{Y\_From\_DLM}$(\bm\beta_{T_{b-1}:T_{b}},\sigma^{-2}_{b}, \bm X_{T_{b-1}:T_{b}}, \bm V_{T_{b-1}:T_{b}}, 1)$,
    \vspace{2mm}
    \State \qquad\qquad\qquad\qquad\quad\  $f_{\bm\phi_b}(\bm\beta_{T_{b-1}:T_{b}},\sigma^{-2}_{b}; \bm y_{T_{b-1}:T_{b}})$, $M$)
    \vspace{2mm}
    \EndFor
    \vspace{2mm}
    \end{algorithmic}
\end{algorithm}

\subsection{Prediction of New Actigraph Trajectories}
\label{sec:actigraph_prediction}


It is also relevant to determine whether the FFBS and ABI can accurately reproduce actigraph trajectories that were not part of its training set to determine its generalizability. We randomly select a small subset of eight trajectories that were not part of the training dataset, totaling 233 observations, and attempt to predict their synthetic values with the fitted parameters from both algorithms. To check the extent of the generalizability of our algorithms, we also extend this assessment to the entire set of 2,102 trajectories held out from the training dataset, totaling 57,786 observations.

\begin{figure}[h!]
    \centering
    \includegraphics[width=6.0cm]{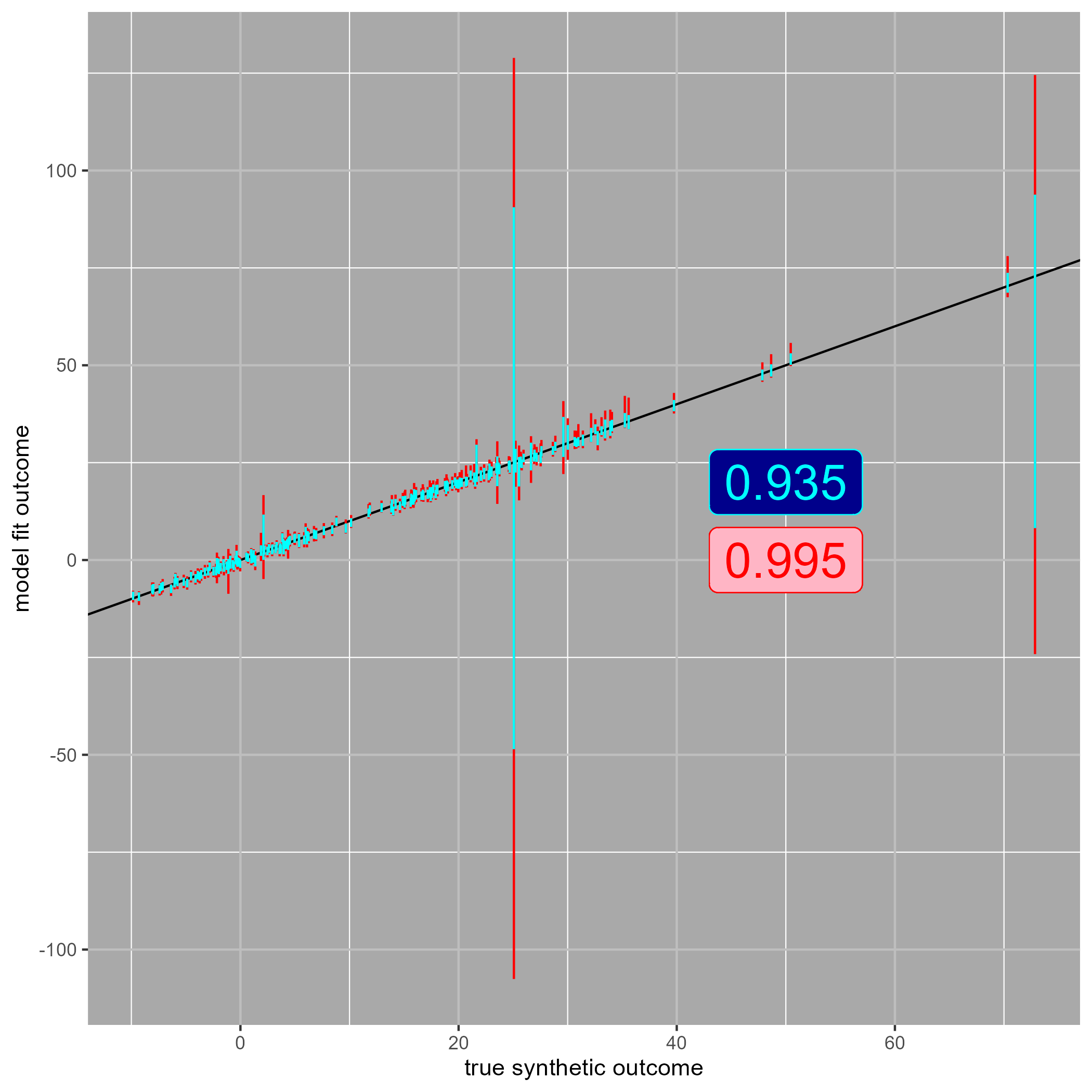}
    \includegraphics[width=6.0cm]{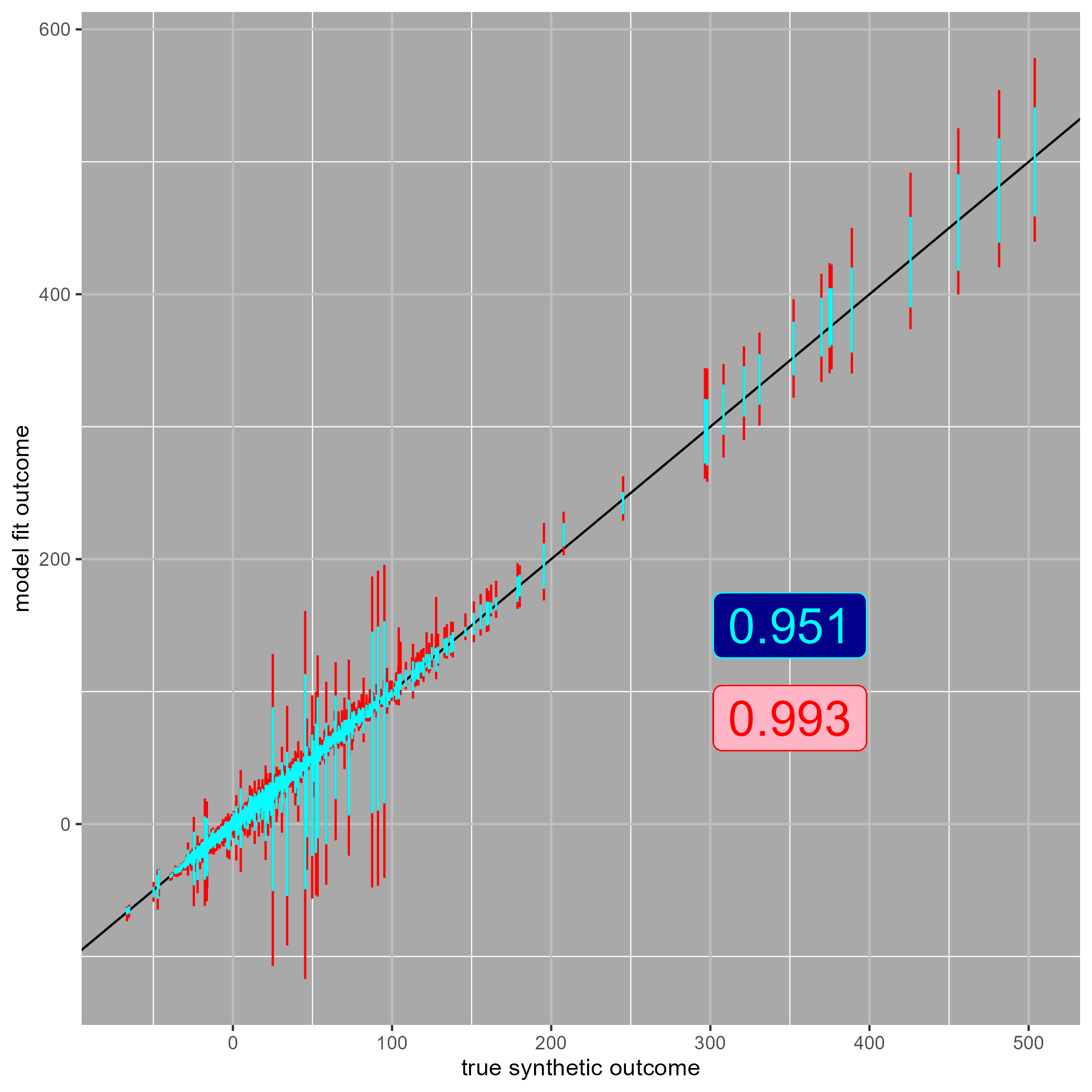} 
    \caption{The credible intervals compared with their synthetic values generated from a single run of \texttt{Y\_FROM\_DLM} from Algorithm \ref{alg:y_from_DLM} for a subset of the actigraph data not present in the training data (left) and the whole actigraph data not present (right). The coverage rates (ranging from 0 to 1) for each setting are also depicted in each plot, with FFBS in blue with cyan text and ABI in pink with red text. The FFBS intervals serve as a reference for prediction, with ABI's visibly wider and proportionately longer than the FFBS intervals depending on the latter's length.}
    \label{fig:predict_yvy}
\end{figure}

The results are shown in Figure \ref{fig:predict_yvy}, where we do so successfully with both FFBS and ABI. Since the subset of trajectories we have trained on yield output that was generated through identical procedures and is thus representative of the data, the posteriors $\bm\beta_{1:T},\sigma^2\mid \bm y_{1:T}$ that are computed are able to generalize well to the entire set of trajectories held out from training.

\subsection{Actigraph Timesheet Imputation}
\label{sec:predict_AT_impute}

We assess the suitability of the Actigraph \\Timesheet imputation strategy: We begin by imputing the covariate matrices to fill in the gaps in time up to the last recorded time step of each trajectory; going further risks biasing the outcome of the coverage.\footnote{The number of time steps to be imputed under this setting comprises about 17.4\% of the combined real and imputed data, i.e. $\sum_{t}n_{t,u}/\sum_{t}(n_{t,u} + n_{t,o})\approx 0.174$.} Letting $\bm V_{t,oo} = \bm I_{n_{t,o}}$ and $\bm V_{t,u} = \bm I_{n_{t,u}}$ for $t = 1,\ldots,T$\footnote{$n_{t,u} = 0$ yields a trivial covariance matrix and is not computationally relevant, since every subject's trajectory at that time step is measured (e.g. $t = 1$).}, we then run \texttt{DLM\_PRIOR} with its specified parameters once to obtain the ground truth parameters and compute one run each of \texttt{Y\_FROM\_DLM}($\bm\beta_{1:T}, \sigma^2, \bm X_{1:T,o}, \bm V_{1:T,oo}, 1$) and \texttt{Y\_FROM\_DLM}($\bm\beta_{1:T}, \sigma^2, \bm X_{1:T,u}^{*}, \bm V_{1:T,uu}, 1$) to generate synthetic outcomes $\bm y_{1:T,o}$ and $\bm y_{1:T,u}$ that align with the DLM as a method of test. Both to leverage the transferability of ABI and to demonstrate its flexibility, we allow for $\bm y_{1:T,o}$ generated in this setting to differ from that in Section \ref{sec:actigraph_prediction}. The analytical details for the imputation strategy for $\bm X_{1:T,u}^*$ are contained in Section \ref{sec:imputation_details}.

\begin{figure}[h!]
    \centering
    \includegraphics[width=6.0cm]{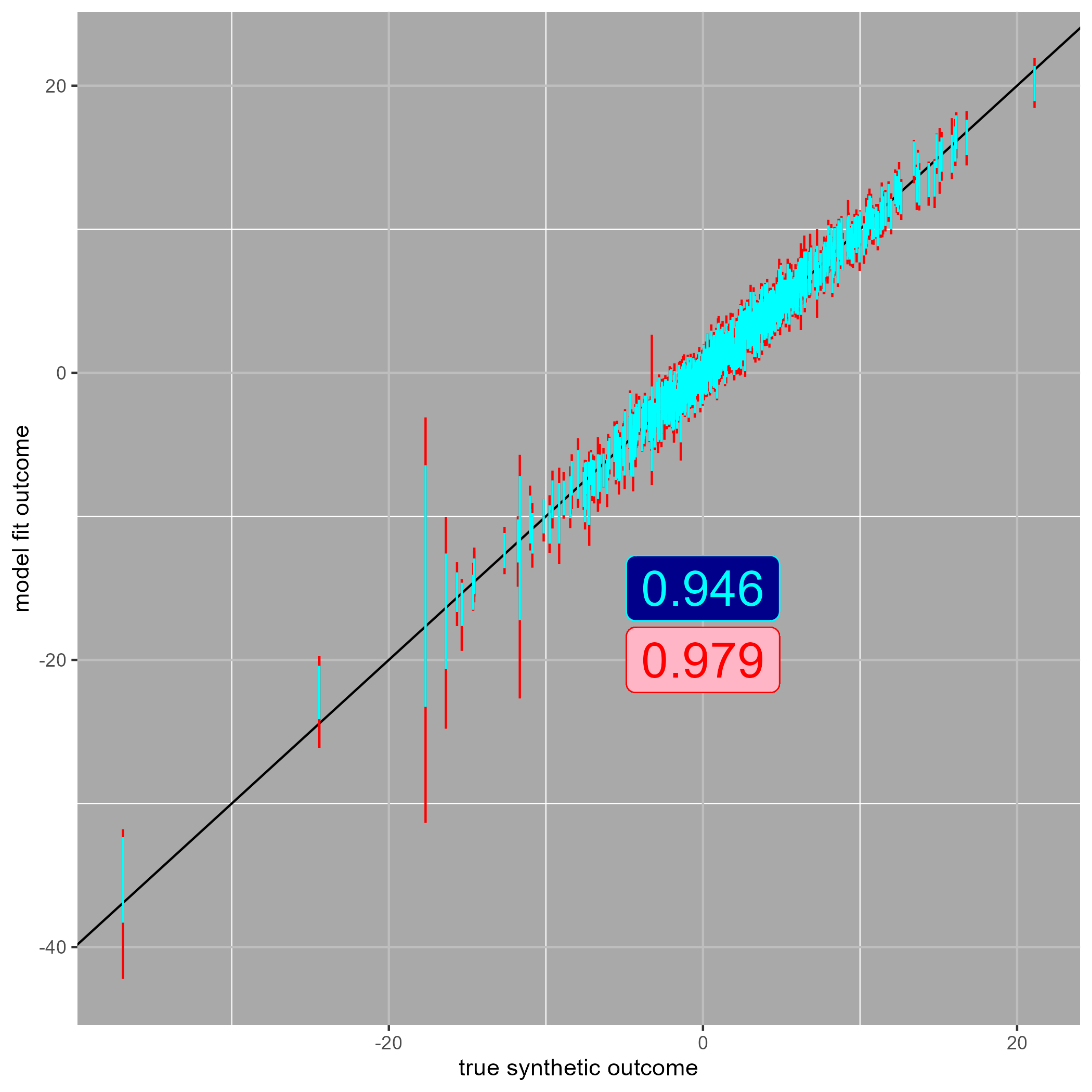}
    \caption{The credible intervals compared with their synthetic values generated from Equation \eqref{eq:dlm} for Actigraph Timesheet imputed data.}
    \label{fig:AT_yvy}
\end{figure}

We then acquire posterior samples from both ABI and the DLM using only the observed outcomes and covariates $\bm X_{1:T,o}$, generate predictions for the unobserved synthetic outcomes using the posterior samples from the existing data $\bm y_{1:T,o}$, and compare the prediction samples $\bm y_{1:T,u}^{*(1:L)}$ to the synthetic outcomes $\bm y_{1:T,u}$. The results are shown in Figure \ref{fig:AT_yvy}, where both intervals cover a substantial portion of the imputed data.

\subsection{Case Study}
\label{sec:case_study}

We next explore the performance of ABI on the real data against the $\log(\text{MAG})$: we pass it on a full run through the FFBS, as well as to the trained ABI. We retain the covariates from training in Section \ref{sec:actigraph_illustrations}. Note that training was completed assuming underlying knowledge of the same set of covariates $\bm X_{1:T,o}$; the main difference here will be that the outcome is recorded and postprocessed from a real device.

\begin{figure}[th!]
    \centering
    \includegraphics[width=6.0cm]{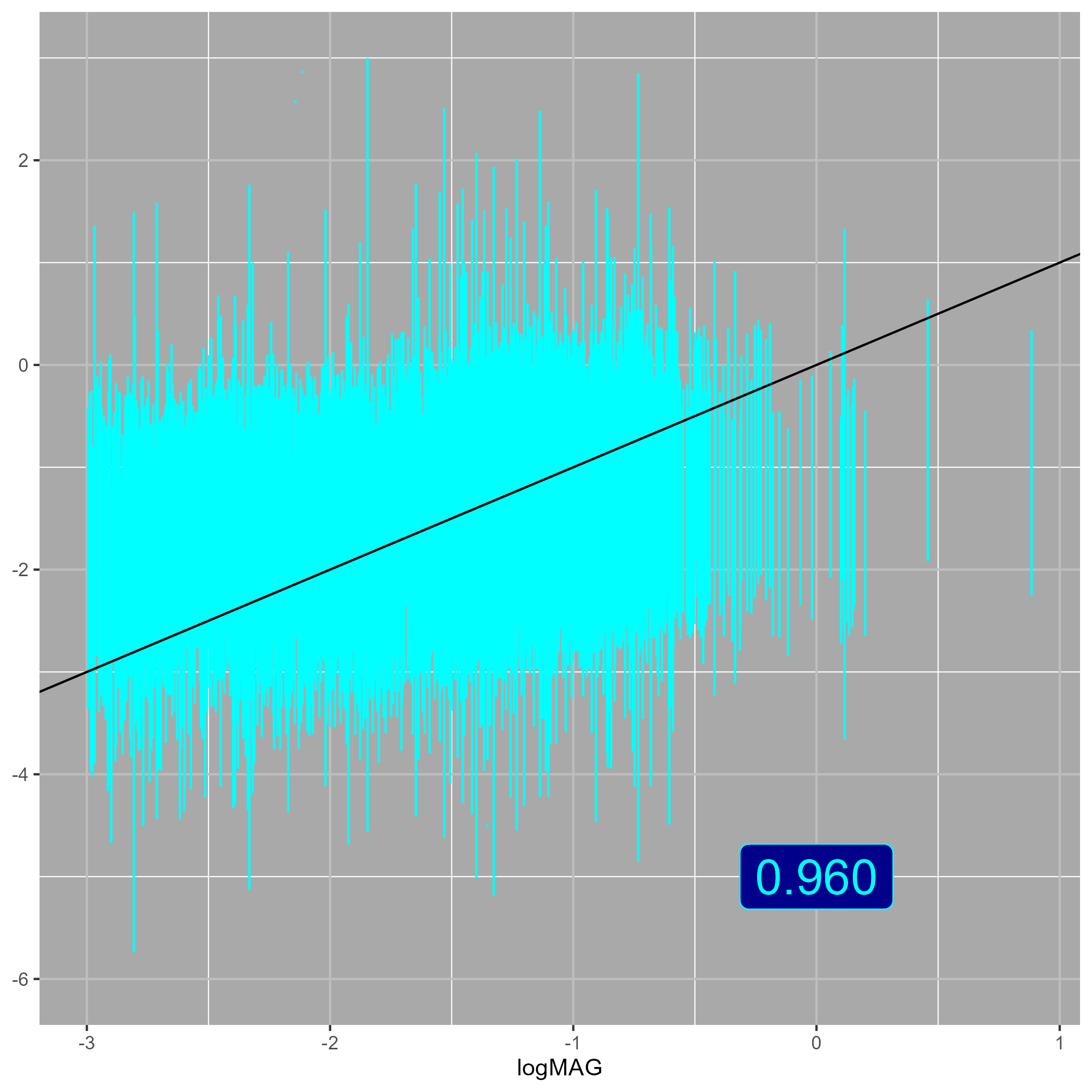}
    \includegraphics[width=6.0cm]{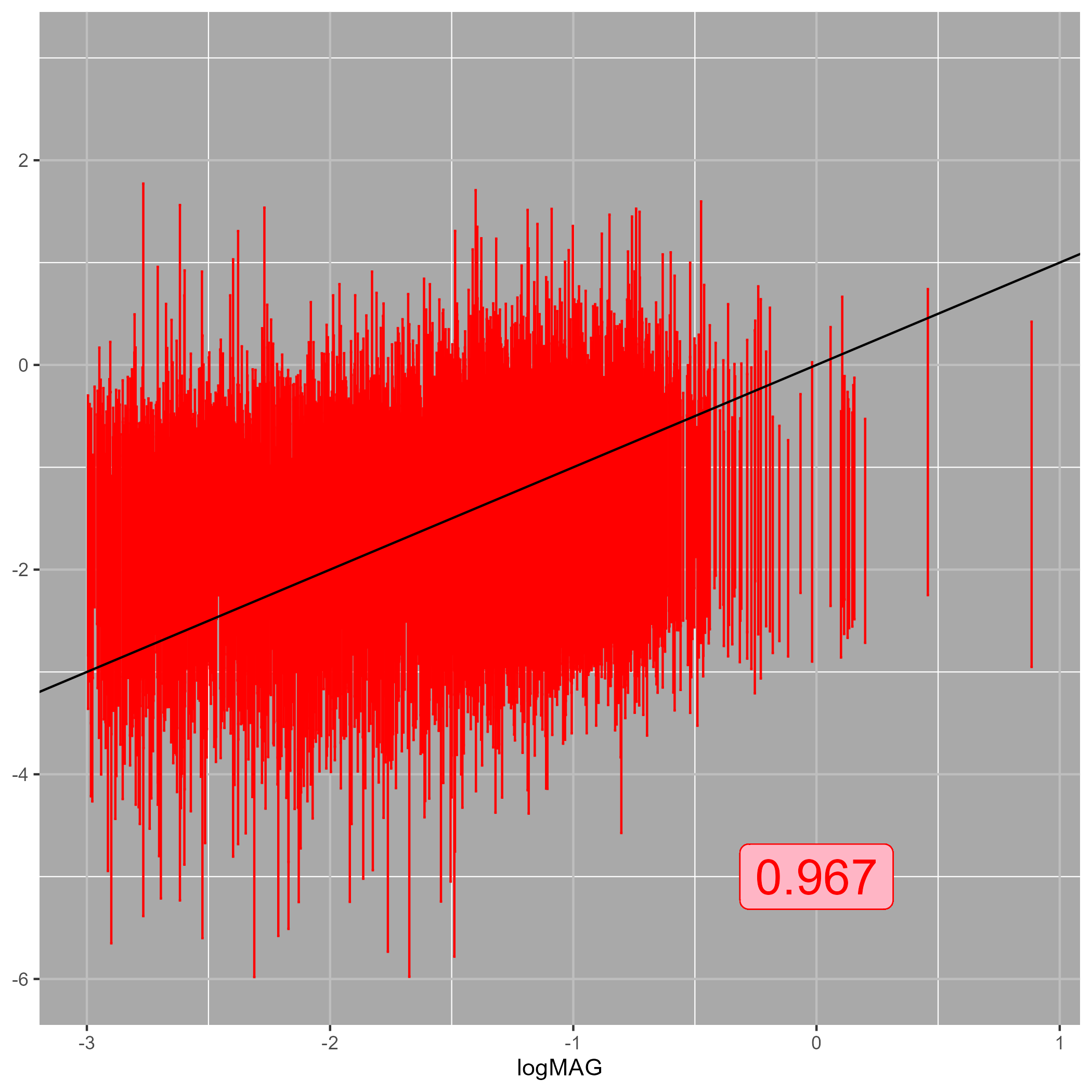}
    \caption{The predictive performance of the FFBS (cyan) and the trained ABI (red) on the $\log(\text{MAG})$. Both the coverage rates and overall rates of credible intervals are similar, showing how ABI is able to adapt to the behavior of the FFBS even after being fed new data to form a posterior distribution.}
    \label{fig:logMAG_vals}
\end{figure}

\begin{figure}[ht!]
    \centering
    \subfloat[Age]{
        \label{fig:logMAG_Age}
        \includegraphics[width=4cm]{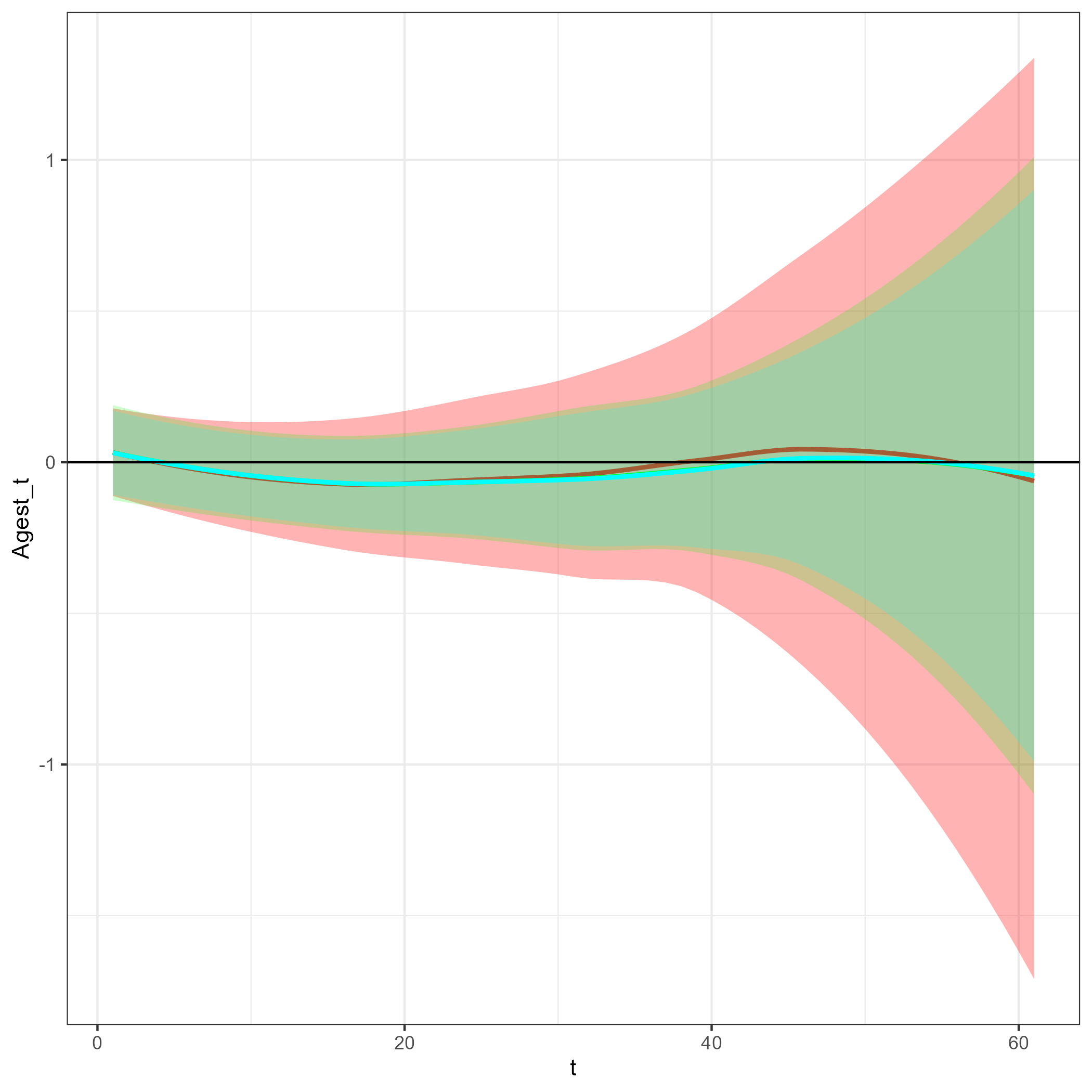}
    }
    \subfloat[Alt.]{
        \label{fig:logMAG_Altitude}
        \includegraphics[width=4cm]{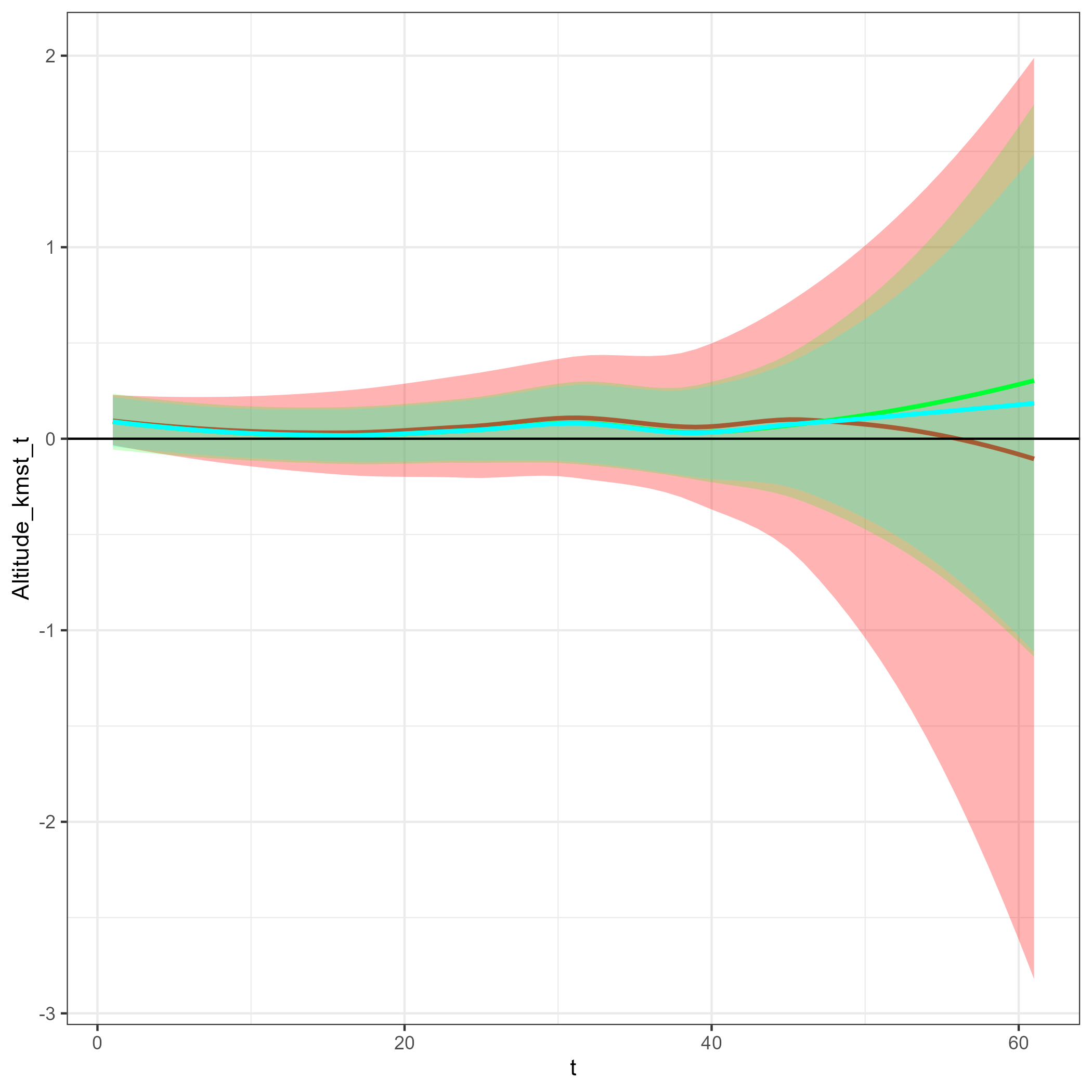}
        
    }
    \subfloat[BMI]{
        \label{fig:logMAG_BMI}
        \includegraphics[width=4cm]{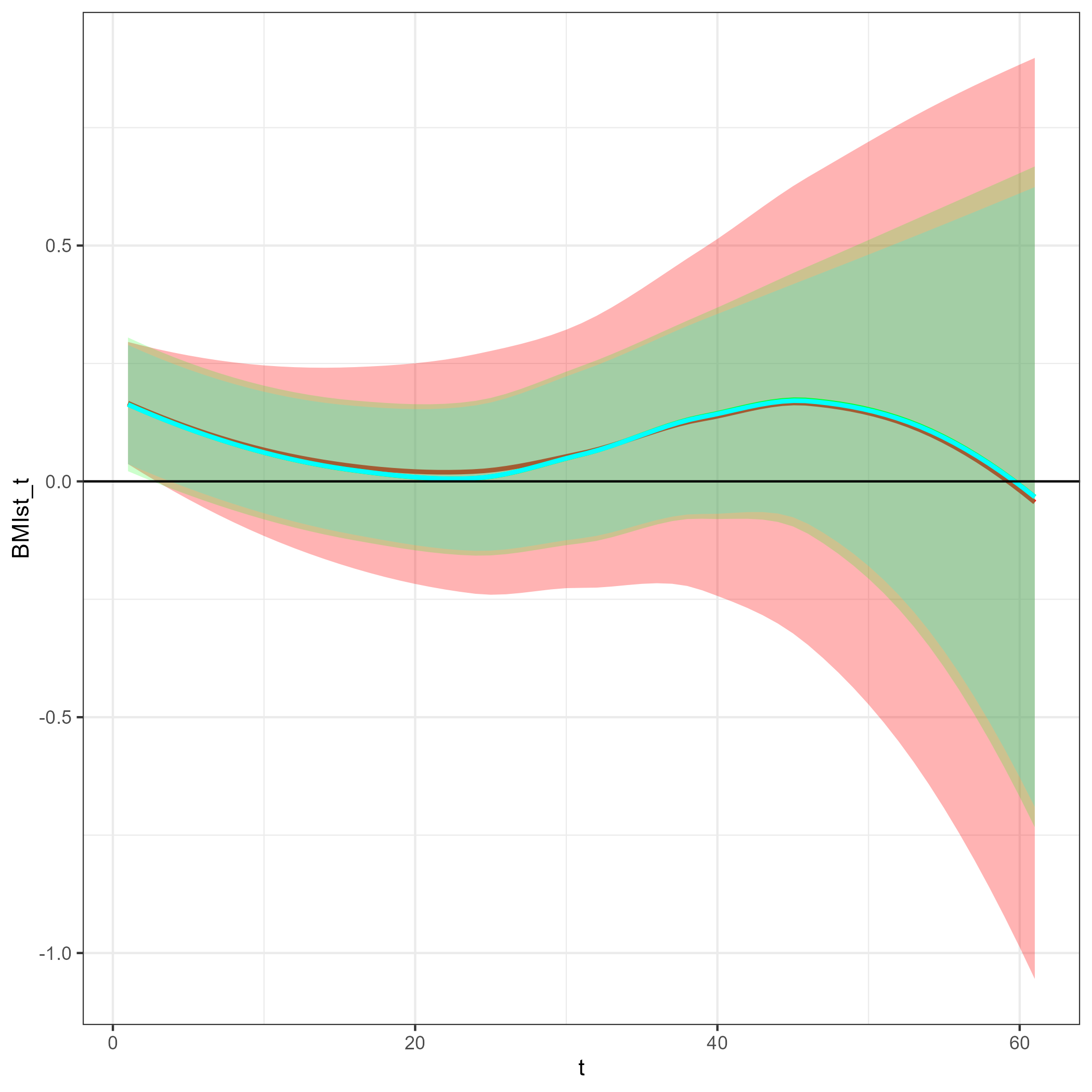}
    }\\
    \subfloat[Dist. from Home (km)]{
        \label{fig:logMAG_DistFromHome_km}
        \includegraphics[width=4cm]{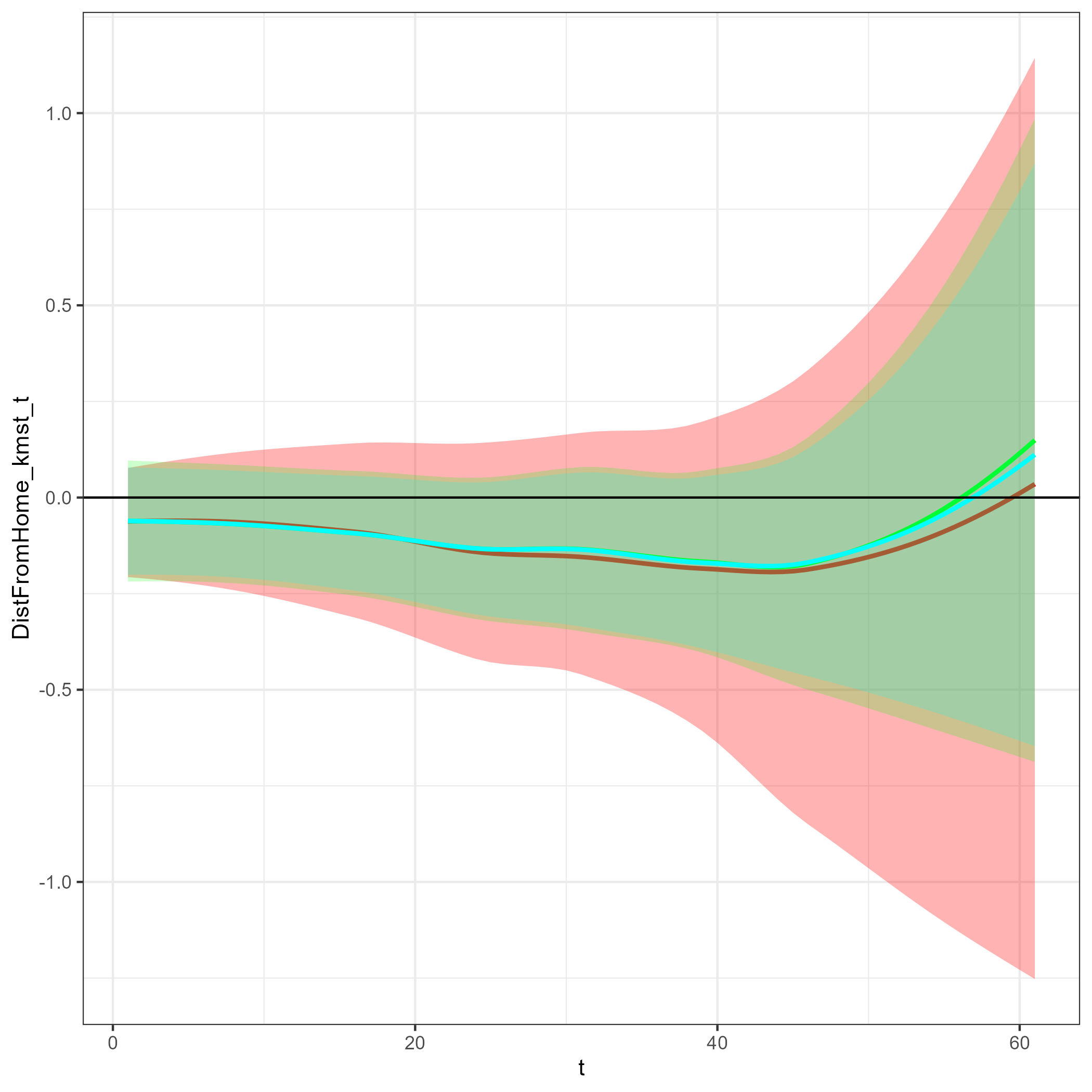}
    }
    \subfloat[Dist. to Parks (km)]{
        \label{fig:logMAG_kDTP}
        \includegraphics[width=4cm]{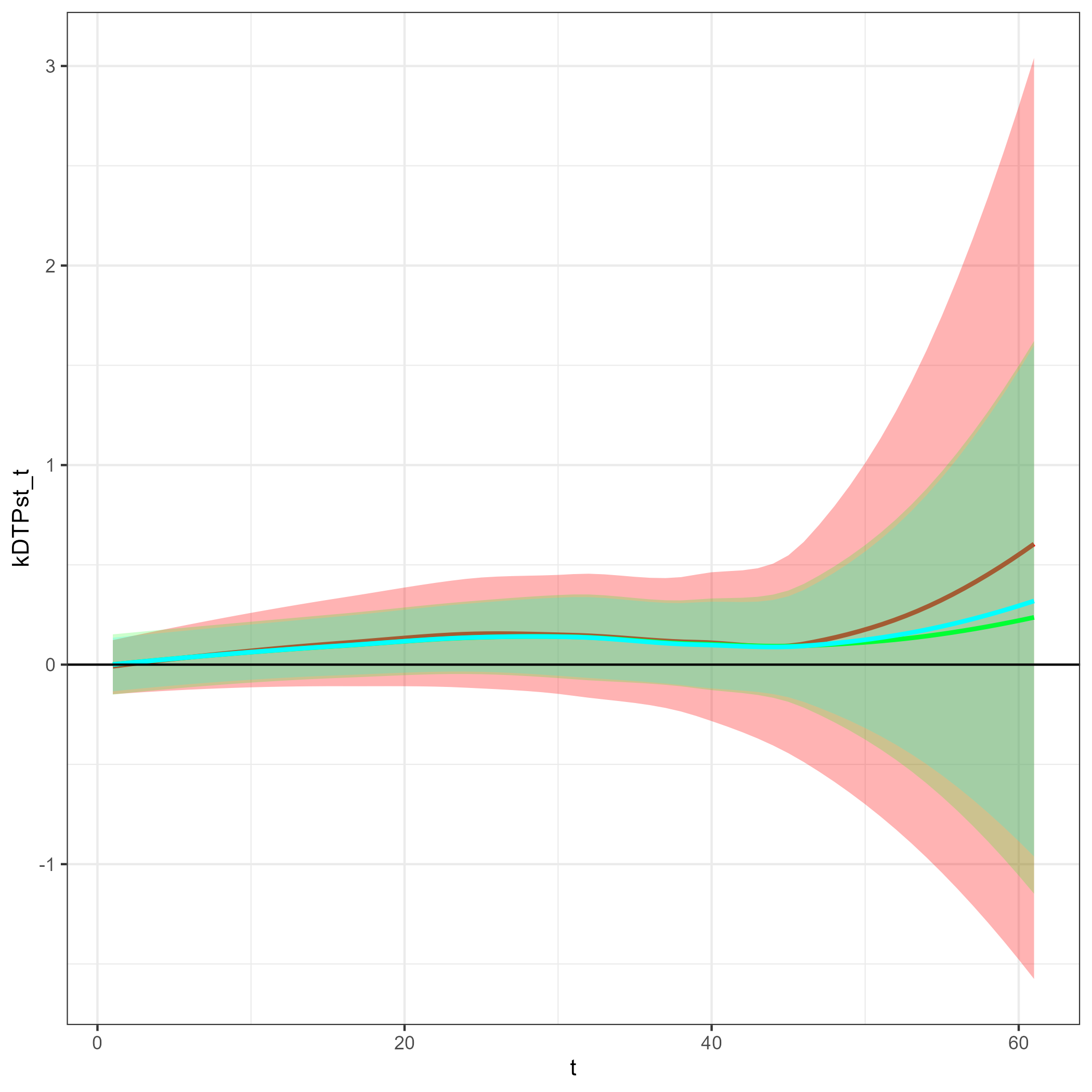}
    }
    \subfloat[Start Time in Day]{
        \label{fig:logMAG_frac_day_0}
        \includegraphics[width=4cm]{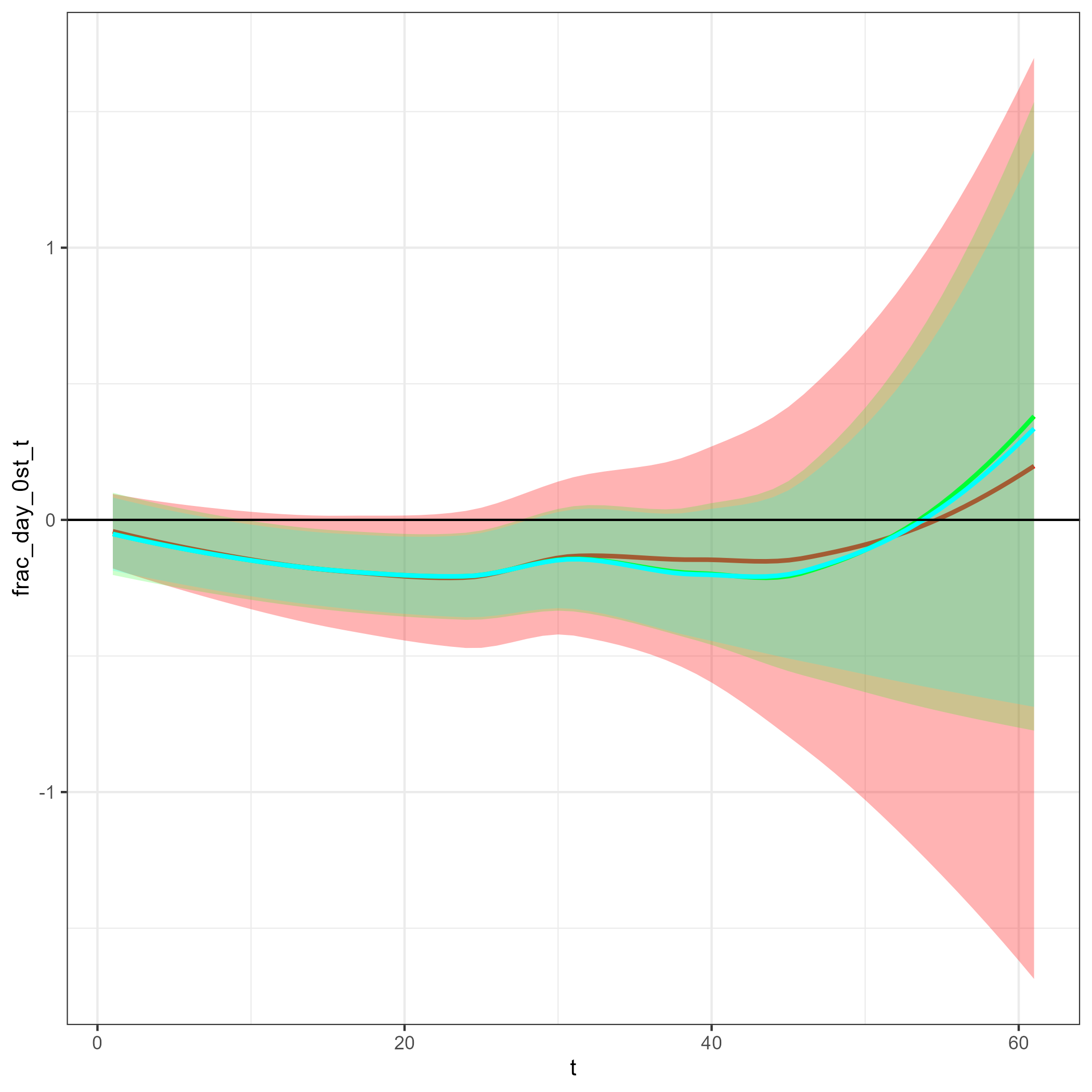}
    }\\
    \subfloat[NDVI]{
        \label{fig:logMAG_NDVI}
        \includegraphics[width=4cm]{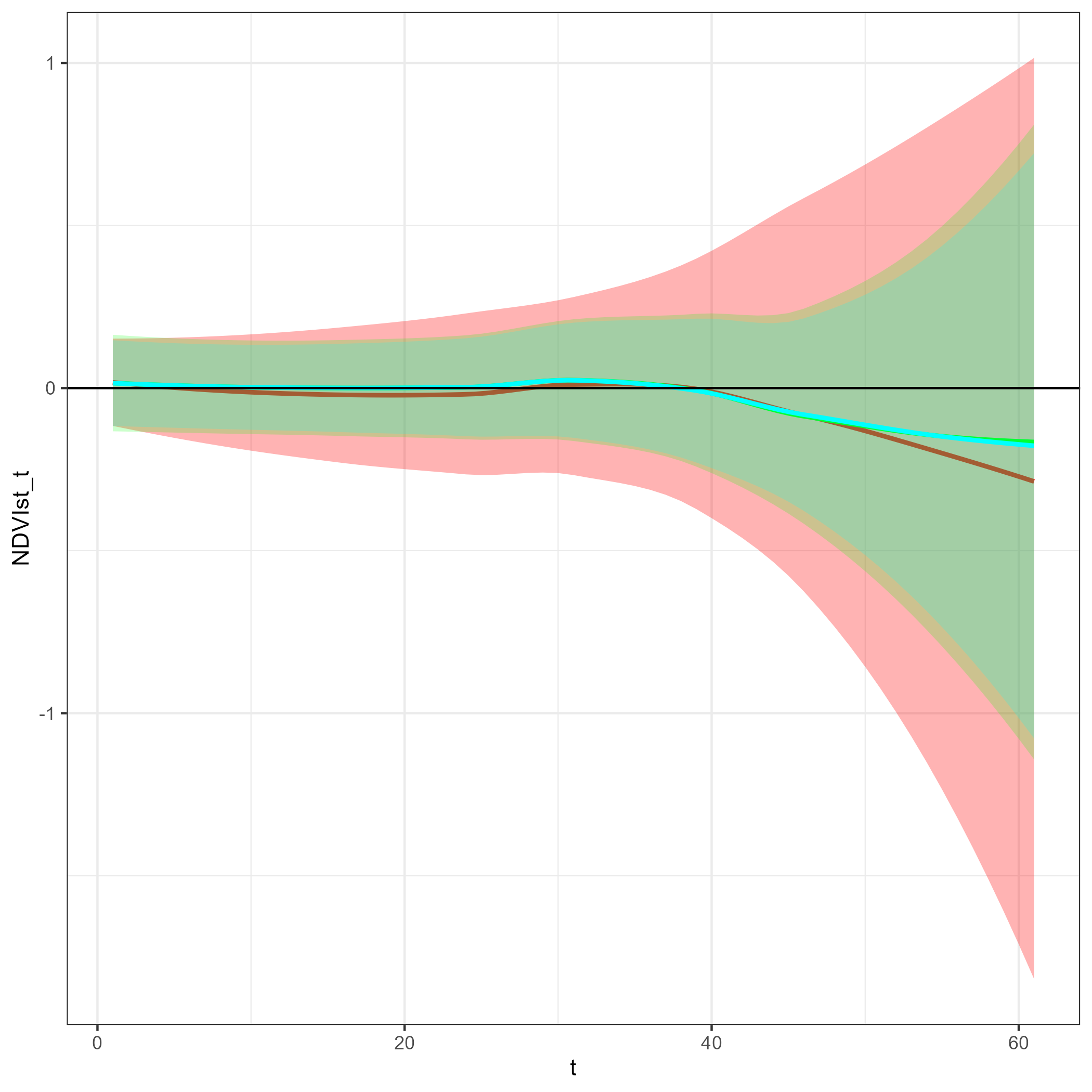}
    }
    \subfloat[Sex]{
        \label{fig:logMAG_Sex}
        \includegraphics[width=4cm]{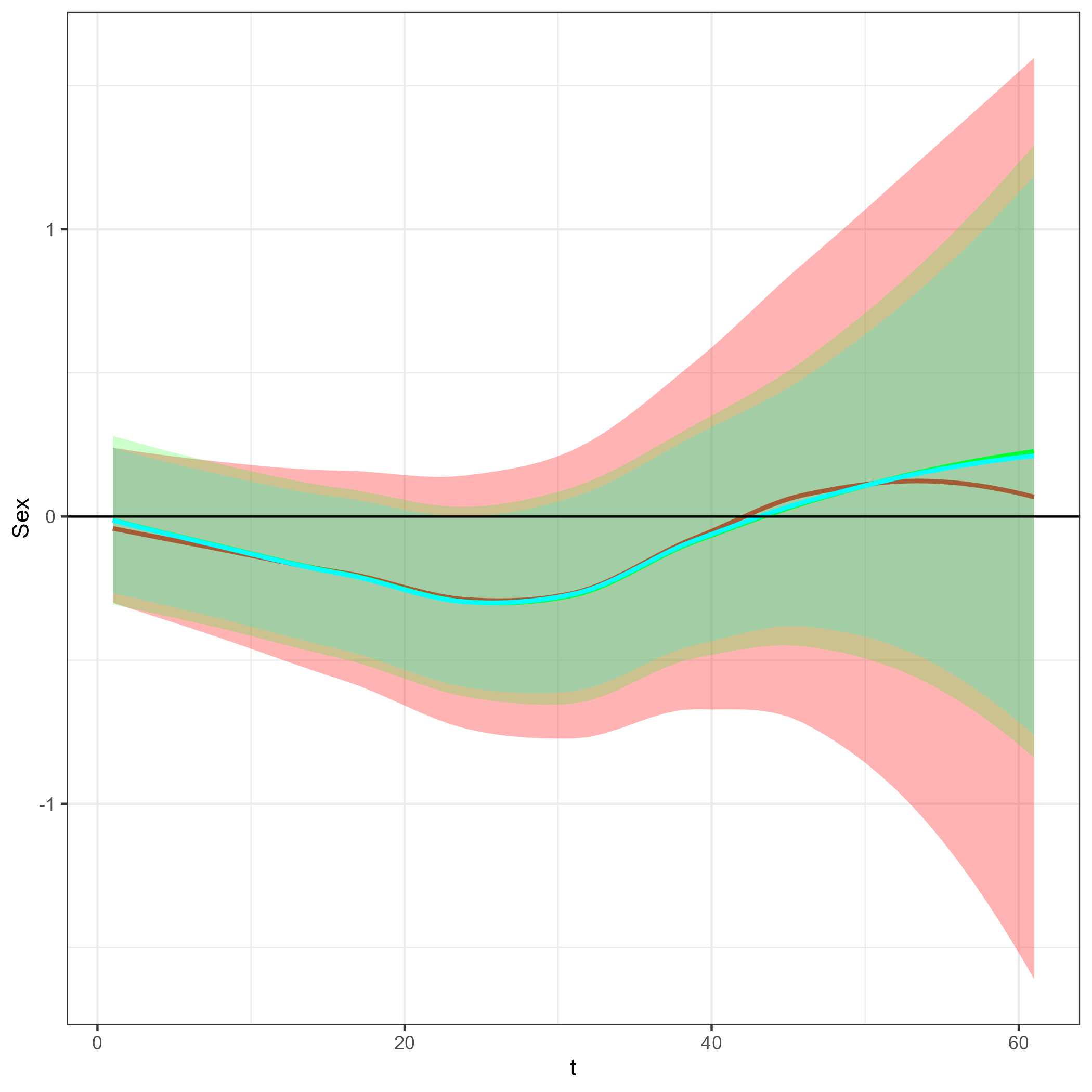}
    }
    \subfloat[Slope]{
        \label{fig:logMAG_Slope}
        \includegraphics[width=4cm]{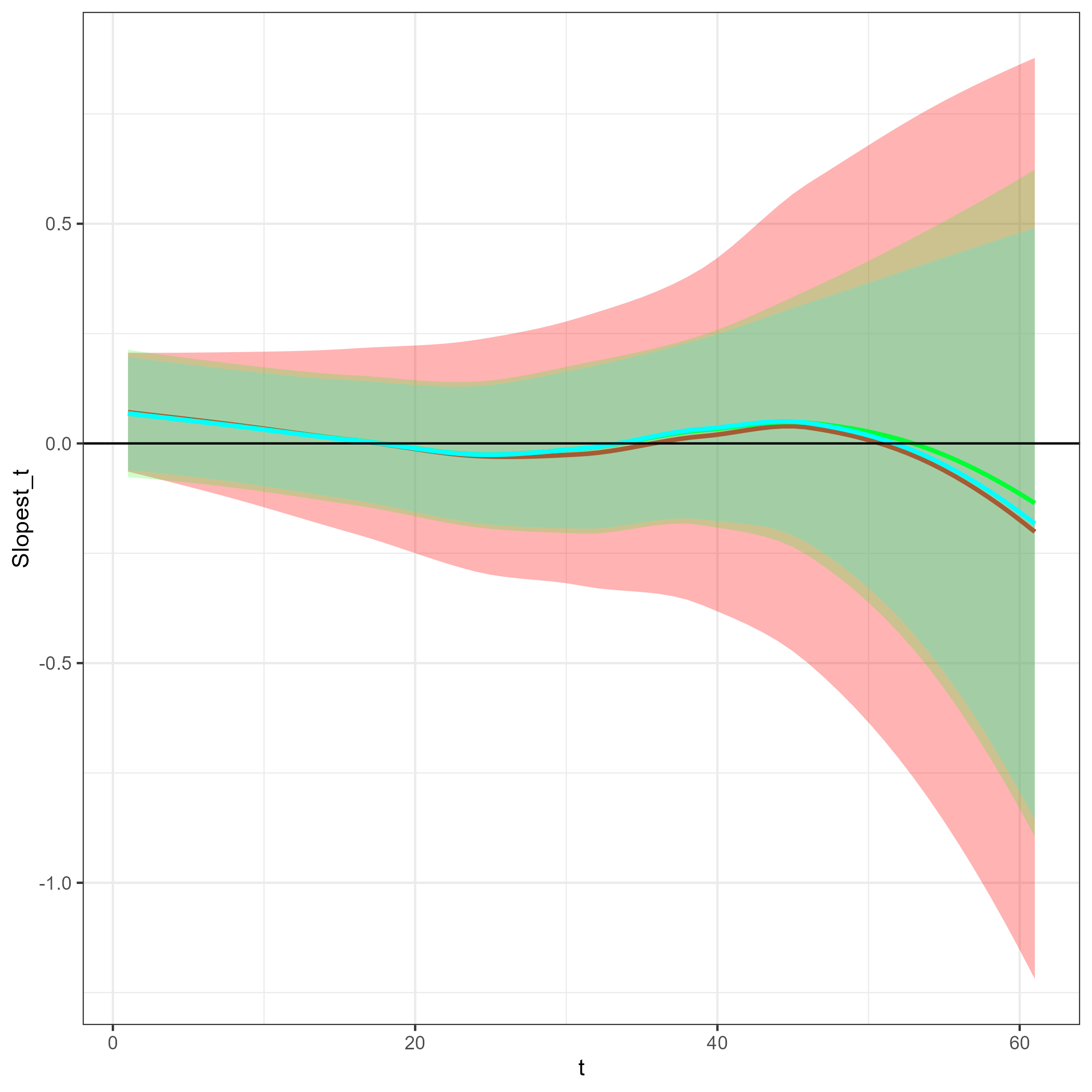}
    }
    
    \caption{The estimated parameter values regressing on $\log(\text{MAG})$ over time due to Algorithm \ref{alg:hier_ABI_DLM_timebatch}, and estimated trajectories due to the FF only (green) and the FFBS (cyan) of the coefficients at each time point. The credible intervals for the ABI output were generated using the same procedure for Figure \ref{fig:synth_plots}. A horizontal line is added at $\bm\beta_{t,j} = 0$ for non-intercept elements $j$ to visualize the statistical significance of each parameter over time.} 
    \label{fig:logMAG_param_plots}
\end{figure}

The results are shown in Figure \ref{fig:logMAG_vals}, with corresponding parameter estimates in Figure \ref{fig:logMAG_param_plots}: the credible intervals obtained through ABI appear to be within a similar range as the FFBS's, as they did for Figure \ref{fig:synth_plots}. $\bm\beta_{t,\text{Start Time in Day}}$ shows some significance for the FFBS around $t=10$ up to around $t=27$, though the significance is lost for ABI. Interestingly, $\bm\beta_{t,\text{BMI}}$ demonstrates some significance in early time steps before losing it later, while $\bm\beta_{t,\text{Sex}}$ is borderline insignificant at around the $t=22$ or $23$. 

\subsection{Computing Environments and Code}
\label{sec:compute_env}

Computations were executed on a laptop running 64-bit Windows 11 with a 12th Gen Intel(R) Core(TM) i7-12700H, 2.30 GHz processor, 32.0 GB RAM at 4800 MHz, 6 GB Graphics Card equipped with multiple GPUs. 

The plots were produced using R. All computer programs required to produce the numerical results in this manuscript are available from the Github repository \\https://github.com/Daniel-Zhou-93/Amortized-Bayesian-Inference-on-Actigraph-Data.  

\section{Discussion}
\label{sec:discussion}

We have extended the original domain of ABI problems to learn time-varying coefficients, which increase in dimension as the number of time steps increases. This particular scenario is additionally adaptable if data from even later time steps is desired, as it is possible to sample from the previous trained time step to train the next time step or collection of time steps as the domain increases. Due to the relatively small sizes of the time intervals, doing so to generalize to larger $T$ may be done without having to rerun the entire algorithm, as long as the underlying trajectories remain the same in existing time intervals which were a part of training. The adaptability of the trained ABI model has been demonstrated when multiple different outcome variables were passed in and used to generate posterior samples of the parameters for each outcome, including with the physically-measured log(MAG), with results closely approximating those of the analytic approach. We have also designed the actigraph timesheet as a model-based imputation procedure and validated its performance on data conforming to the underlying DLM. Its flexibility across models is leveraged by our use of ABI, with the potential to extend it to more complex models or other machine learning implementations.

A central limitation of the present framework is its reliance on the Markovian graphical structure. Many complex systems exhibit path-dependent or long-range interactions that cannot be captured by local conditional independence assumptions. Extending the FFBS to non-Markovian graphical models, where dependence may propagate beyond immediate neighborhoods, poses both conceptual and technical challenges, particularly in developing tractable factorization procedures \citep{peruzzi2022highly}. Specific block-dependent Markovian structures such as the season-episode model \citep{2025BayesianModelingMechSystems}, have been shown to scale the FFBS algorithm to massive datasets \citep[also see][]{presicce2026adaptivemarkovianspatiotemporaltransfer}. ABI has greater capability to generalize to graphs, and promising approaches have been demonstrated through graph neural networks and meta-learning methods \citep{SainsburyDale2025GNNIrregularSpatial, ortega2019metalearningsequentialstrategies}.

Future work will also investigate the specialized stochastic processes described in \\\citep{wakayama2024processbasedinferencespatialenergetics} to amortize Bayesian inference for spatial-temporal data from wearable devices. In addition to \texttt{BayesFlow}, we intend to pursue supervised training of deep learning networks. Supervision typically requires feeding posterior distributions (or summaries such as means, standard deviations, quantiles, etc.) to the network. However, the supervision of deep networks for such complex models will require large numbers of simulated Bayesian data analysis, rendering iterative algorithms such as MCMC or INLA unsuitable. Here, recent promising developments in Bayesian predictive stacking \citep{zhang2025geostacking, panEtAl2025} for geostatistical inference can be employed to supervise deep networks \citep{Presicce2025amortizedGeospatial}. These directions collectively point toward a flexible yet rigorously grounded framework for modeling complex dependence and improving predictive performance.

\section{Acknowledgements}
\label{sec:acknowledgements}

The authors acknowledge funding from the National Science Foundation (NSF) through grant DMS-2113778 and from the National Institute of Health (NIH/NIGMS) through NIGMS R01GM148761. The authors thank Xiang Chen for his help in the initial processing and visualization of the actigraph dataset. 

\bibliographystyle{plainnat}
\bibliography{dlm_all}


\appendix\section{Using Priors to Bridge Time Segments for ABI}
\label{sec:bridge_priors}

Since ABI is trained with synthetic data, and our synthetic data is generated from a model with an analytic solution, we consider the analytic properties of the parameters that generate the data. Specifically, we consider the values of $\mathbb{E}_{\bm{y}_{1:t},\bm{\beta}_t,\sigma^2}[\bm\beta_t] = \mathbb{E}_{\bm{y}_{1:t}}[\mathbb{E}_{\bm{\beta_t},\sigma^2\mid y_{1:t}}[\bm\beta_t]]$ and $\mathrm{Var}_{\bm{y}_{1:t},\bm{\beta}_t,\sigma^2}(\bm\beta_t)$, where we treat $\bm y_{1:t}$ (as an instance of $\bm y_{1:T}^{(m)}$ in Section \ref{sec:ABI}) as random rather than fixed:
\begin{equation}
    \label{eq:FFEy}
    \begin{split}
        \mathbb{E}_{\bm{\beta_t}\mid \sigma^2,y_{1:t}}[\bm\beta_t]] &= \bm{m}_t\\
        &= \bm{c}_t+\bm{C}_t\bm{X}_t^\T\bm{Q}_t^{-1}(\bm{y}_{t}-\bm{q}_t)\\
        \mathbb{E}_{\bm{y}_{1:t},\bm{\beta}_t,\sigma^2}[\bm\beta_t] &= \mathbb{E}_{\bm{y}_{1:t}}[\mathbb{E}_{\sigma^2\mid \bm{y}_{1:t}}[\mathbb{E}_{\bm{\beta_t}\mid \sigma^2,y_{1:t}}[\bm\beta_t]]]\\
        &= \mathbb{E}_{\bm y_{1:t}}[\bm m_t]\\
        &= \mathbb{E}_{\bm{y}_{1:(t-1)}}[\mathbb{E}_{\bm{y}_t\mid \bm{y}_{1:(t-1)}}[\bm{c}_t+\bm{C}_t\bm{X}_t^\T\bm{Q}_t^{-1}(\bm{y}_{t}-\bm{q}_t)]]\\
        &= \mathbb{E}_{\bm{y}_{1:(t-1)}}[\bm{G}_t\bm{m}_{t-1}]
    \end{split}
\end{equation}

We note that $\mathbb{E}_{\bm{y}_t\mid \bm{y}_{1:(t-1)}}[\bm{C}_t\bm{X}_t\bm{Q}_t^{-1}(\bm{y}_{t}-\bm{q}_t)] = \bm 0$ because from Algorithm \ref{alg:KF}, \\$\bm y_t \mid \bm y_{1:(t-1)} \sim \mathcal{T}(\bm q_t,\frac{b_{t-1}}{a_{t-1}}\bm Q_t)$. We expand $\bm c_t = \bm G_t \bm m_{t-1}$ for clarity. Recursion of the last two lines of Equation \eqref{eq:FFEy} from $\bm y_{t-1}$ down to $\bm y_1$ gives us:
\begin{equation}
    \label{eq:FFEy_recursion}
    \mathbb{E}_{\bm{y}_{1:t},\bm{\beta}_t,\sigma^2}[\bm\beta_t] = \left(\prod_{s=1}^{t}\bm G_s\right)\bm m_0
\end{equation}
where $\prod_{s=1}^{t}\bm G_s = \bm G_t\bm G_{t-1}\cdots \bm G_1$ is taken with respect to left-multiplication (i.e. the product series expands leftwards).

We proceed to $\mathrm{Var}_{\bm{y}_{1:t},\bm{\beta}_t,\sigma^2}(\bm\beta_t)$:
\begin{equation}
    \label{eq:FFVary_start}
    \begin{split}
        \mathrm{Var}_{\bm{y}_{1:t},\bm{\beta}_t,\sigma^2}(\bm\beta_t) &= \mathbb{E}_{\bm{y}_{1:t},\bm{\beta}_t,\sigma^2}[\bm\beta_t\bm\beta_t^{\T}] - \mathbb{E}_{\bm{y}_{1:t},\bm{\beta}_t,\sigma^2}[\bm\beta_t]\mathbb{E}_{\bm{y}_{1:t},\bm{\beta}_t,\sigma^2}[\bm\beta_t]^{\T}\\
        \mathbb{E}_{\bm{y}_{1:t},\bm{\beta}_t,\sigma^2}[\bm\beta_t\bm\beta_t^{\T}] &= \mathbb{E}_{\sigma^2}[\mathbb{E}_{\bm y_{1:t}\mid\sigma^2}[\mathbb{E}_{\bm{\beta}_t\mid \sigma^2,\bm y_{1:t}}[\bm\beta_t\bm\beta_t^\T]]]\\
        &= \mathbb{E}_{\sigma^2}[\mathbb{E}_{\bm y_{1:t}\mid\sigma^2}[\sigma^2 \bm M_t + \bm m_t \bm m_t^\T]]\\
        &= \frac{b_0}{a_0 - 1}\bm M_t + \mathbb{E}_{\sigma^2}[\mathbb{E}_{\bm y_{1:t}\mid\sigma^2}[\bm m_t \bm m_t^\T]]
    \end{split}
\end{equation}

Next, we simplify $\mathbb{E}_{\sigma^2}[\mathbb{E}_{\bm y_{1:t}\mid\sigma^2}[\bm m_t \bm m_t^\T]]$. We briefly set aside the expectation with respect to $\sigma^2$ and proceed with the same recursive strategy as with Equation \eqref{eq:FFEy}:
\begin{equation}
    \label{eq:FFVarymmT_1t}
    \begin{split}
        \mathbb{E}_{\bm y_{1:t}\mid\sigma^2}[\bm m_t \bm m_t^\T] &= \mathbb{E}_{\bm y_{1:(t-1)}\mid\sigma^2}[\mathbb{E}_{\bm y_{t}\mid \bm y_{1:(t-1)},\sigma^2}[\bm m_t \bm m_t^\T]]\\
        \mathbb{E}_{\bm y_{t}\mid \bm y_{1:(t-1)},\sigma^2}[\bm m_t \bm m_t^\T] &= \mathbb{E}_{\bm y_{t}\mid \bm y_{1:(t-1)},\sigma^2}[(\bm{G}_t \bm m_{t-1} + \bm{C}_t\bm{X}_t^\T\bm{Q}_t^{-1}(\bm{y}_{t}-\bm{q}_t))\\
        &\qquad\qquad\qquad\quad(\bm{G}_t \bm m_{t-1} + \bm{C}_t\bm{X}_t^\T\bm{Q}_t^{-1}(\bm{y}_{t}-\bm{q}_t))^\T]\\
        &= \mathbb{E}_{\bm y_{t}\mid \bm y_{1:(t-1)},\sigma^2}[\bm{G}_t \bm m_{t-1} \bm m_{t-1}^\T \bm G_t^\T +\\
        &\qquad\qquad \bm{G}_t \bm m_{t-1}(\bm{y}_{t}-\bm{q}_t)^\T \bm Q_t^{-1}\bm X_t \bm C_t + \\
        &\qquad\qquad \bm{C}_t\bm{X}_t^\T\bm{Q}_t^{-1}(\bm{y}_{t}-\bm{q}_t)\bm m_{t-1}^\T \bm G_t^\T + \\
        &\qquad\qquad \bm{C}_t\bm{X}_t^\T\bm{Q}_t^{-1}(\bm{y}_{t}-\bm{q}_t)(\bm{y}_{t}-\bm{q}_t)^\T \bm Q_t^{-1}\bm X_t \bm C_t]\\
        &= \bm{G}_t \bm m_{t-1} \bm m_{t-1}^\T \bm G_t^\T + \\
        &\qquad\bm{C}_t\bm{X}_t^\T\bm{Q}_t^{-1}\mathbb{E}_{\bm y_{t}\mid \bm y_{1:(t-1)},\sigma^2}[(\bm{y}_{t}-\bm{q}_t)(\bm{y}_{t}-\bm{q}_t)^\T]\bm Q_t^{-1}\bm X_t \bm C_t\\
        &= \bm{G}_t \bm m_{t-1} \bm m_{t-1}^\T \bm G_t^\T + \sigma^2\bm{C}_t\bm{X}_t^\T\bm{Q}_t^{-1}\bm{X}_t\bm{C}_t
    \end{split}
\end{equation}

Recursively taking the expectation of $\bm{G}_t \bm m_{t-1} \bm m_{t-1}^\T \bm G_t^\T$ and its analogous terms gives us the following expression:
\begin{equation}
    \label{eq:FFVarymmT_final}
    \begin{split}
    \mathbb{E}_{\bm y_{1:t}\mid\sigma^2}[\bm m_t \bm m_t^\T] &= \left(\prod_{s=1}^t \bm G_s\right)\bm m_0 \bm m_0^\T \left(\prod_{s=1}^t \bm G_s\right)^\T + \\
    &\qquad\sigma^2\sum_{r=1}^{t} \left(\prod_{s=r+1}^{t} \bm G_s\right)\bm{C}_r\bm{X}_r^\T\bm{Q}_r^{-1}\bm{X}_r\bm{C}_r \left(\prod_{s=r+1}^{t} \bm G_s\right)^\T
    \end{split}
\end{equation}

The expectation over $\sigma^2$ is trivial.

We have all the terms we need to compute $\mathrm{Var}_{\bm{y}_{1:t},\bm{\beta}_t,\sigma^2}(\bm\beta_t)$:
\begin{equation}
    \label{eq:FFVary}
    \begin{split}
        \mathrm{Var}_{\bm{y}_{1:t},\bm{\beta}_t,\sigma^2}(\bm\beta_t) &= \frac{b_0}{a_0 - 1}\left(\bm M_t + \sum_{r=1}^{t} \left(\prod_{s=r+1}^{t} \bm G_s\right)\bm{C}_r\bm{X}_r^\T\bm{Q}_r^{-1}\bm{X}_r\bm{C}_r \left(\prod_{s=r+1}^{t} \bm G_s\right)^\T\right) +\\
        &\qquad \left(\prod_{s=1}^t \bm G_s\right)\bm m_0 \bm m_0^\T \left(\prod_{s=1}^t \bm G_s\right)^\T - \mathbb{E}_{\bm{y}_{1:t},\bm{\beta}_t,\sigma^2}[\bm\beta_t]\mathbb{E}_{\bm{y}_{1:t},\bm{\beta}_t,\sigma^2}[\bm\beta_t]^{\T}\\
        &= \frac{b_0}{a_0 - 1}\left(\bm M_t + \sum_{r=1}^{t} \left(\prod_{s=r+1}^{t} \bm G_s\right)\bm{C}_r\bm{X}_r^\T\bm{Q}_r^{-1}\bm{X}_r\bm{C}_r \left(\prod_{s=r+1}^{t} \bm G_s\right)^\T\right)
    \end{split}
\end{equation}

\subsection{Simplification Under Chosen Parameters}
\label{sec:bridge_prior_simplifications}

In the context of our paper, we choose $\bm G_t = \bm I_p$ for all $t$. This dramatically simplifies Equations \eqref{eq:FFEy_recursion} and \eqref{eq:FFVary} into the respective expressions:
\begin{equation}
    \label{eq:FFEy_simple}
    \mathbb{E}_{\bm{y}_{1:t},\bm{\beta}_t,\sigma^2}[\bm\beta_t] = \bm m_0
\end{equation}

\begin{equation}
    \label{eq:FFVary_simple_step1}
    \begin{split}
        \mathrm{Var}_{\bm{y}_{1:t},\bm{\beta}_t,\sigma^2}(\bm\beta_t) &= \frac{b_0}{a_0 - 1}\left(\bm M_t + \sum_{r=1}^{t} \bm{C}_r\bm{X}_r^\T\bm{Q}_r^{-1}\bm{X}_r\bm{C}_r \right)\\
        &= \frac{b_0}{a_0 - 1}\left(\bm C_t - \bm{C}_t\bm{X}_t^\T\bm{Q}_t^{-1}\bm{X}_t\bm{C}_t + \sum_{r=1}^{t} \bm{C}_r\bm{X}_r^\T\bm{Q}_r^{-1}\bm{X}_r\bm{C}_r \right)\\
        &= \frac{b_0}{a_0 - 1}\left(\bm G_t \bm M_{t-1} \bm G_t^\T + \bm W_t + \sum_{r=1}^{t-1} \bm{C}_r\bm{X}_r^\T\bm{Q}_r^{-1}\bm{X}_r\bm{C}_r \right)\\
        &= \frac{b_0}{a_0 - 1}\left(\bm M_{t-1} + \bm W_t + \sum_{r=1}^{t-1} \bm{C}_r\bm{X}_r^\T\bm{Q}_r^{-1}\bm{X}_r\bm{C}_r \right)
    \end{split}
\end{equation}

Recursively simplifying equation \eqref{eq:FFVary_simple_step1} results in the following simplification:
\begin{equation}
    \label{eq:FFVary_simple_step2}
    \mathrm{Var}_{\bm{y}_{1:t},\bm{\beta}_t,\sigma^2}(\bm\beta_t) = \frac{b_0}{a_0 - 1}\left(\bm M_0 + \sum_{r=1}^{t}\bm W_r\right)
\end{equation}

Finally, we simplify to $\mathrm{Var}_{\bm{y}_{1:t},\bm{\beta}_t,\sigma^2}(\bm\beta_t) = \frac{b_0(t+1)}{a_0 - 1}\bm I_p$, since we also set $\bm M_0 = \bm I_p$ and $\bm W_t = \bm I_p$ for all $t$. More complicated expressions can be acquired for different choices of $\bm M_0$ and $\bm W_t$ (as well as for $\bm G_t$).

This prior may then be used to bridge time segments to allow for theoretically independent training between segments. Furthermore, in settings where expressions for the prior for the next time segment may be intractable, it may be possible to use the iterated expectations to approximate the desired parameters for the priors for the next time segment, such as with Monte Carlo methods.


\section{Analytic Imputation Specifications}
\label{sec:imputation_details}

We detail the imputation procedure here used to generate $\bm X_{1:T,u}^{*}$. Each entry in $\bm X_{1:T,o}$ is indexed by subject ID, time step $t$, and Latitude and Longitude coordinates.

The first and most basic step is to account for subject- and trajectory-specific covariates, because these can simply be substituted into the corresponding columns in $\bm X_{1:T,u}^{*}$ without need for any creative imputation strategy. These columns correspond to the subject's Age, BMI, and Sex, and the relative time of day from 7 am - 11 pm that the subject's trajectory started. While the PASTA-LA dataset records subject trajectories over the course of two separate two-week periods per subject differing by about 6 months, it only accounts for the subject's Age and BMI at the start of the subject's entry and treats them as constant throughout the course of the study. The last variable, the relative time of day the subject began to move, is constructed based on the first time the trajectory is recorded, and by its definition is constant per trajectory per subject.

To prepare ourselves to detail the imputation strategy, we borrow terminology from \cite{wakayama2024processbasedinferencespatialenergetics} and define $\bm\gamma_{i,t_0}(t)$ for the geographical trajectory indexed by the subject $i$ and starting date and time $t_0$ at time $t$. $\bm\gamma_{i,t_0}(t)$ encodes the Latitude and Longitude coordinates at $t$ time steps from $t_0$. We also define the set of relative time points of interest as $\tau = \{1,\ldots,61\}$, and the set of known and unknown time points for subject $i$ and starting date and time $t_0$ as $\tau_{i,t_0,o}$ and $\tau_{i,t_0,u}$ respectively.

$\bm\gamma_{i,t_0}$ then has its geocoordinates linearly imputed for times $t\in \tau_{i,t_0,u}$, resulting in the imputed trajectory $\bm\gamma_{i,t_0}^{*}(t)$ that returns geocoordinates at observed times in $\tau_{i,t_0,o}$ and imputed geocoordinates at unobserved times in $\tau_{i,t_0,u}$:
\begin{equation}
    \label{eq:gamma_impute}
    \bm\gamma_{i,t_0}^{*}(t) = \frac{t - t_1}{t_2 - t_1}(\bm\gamma_{i,t_0}(t_2) - \bm\gamma_{i,t_0}(t_1)) + \bm\gamma_{i,t_0}(t_1)
\end{equation}
where
\begin{equation*}
    t_1 := \max \{t^*\mid t^*<t\text{ and }t^*\in \tau_{i,t_0,o}\},\quad t_2 := \min \{t^*\mid t^*>t\text{ and }t^*\in \tau_{i,t_0,o}\}
\end{equation*}

Call the set of columns to be imputed with Gaussian radial averaging $\mathcal{I}$, and $\mathscr{G}(\bm\gamma, r_S) := \{\bm\gamma_{j,t_0'}(t')\mid \mathrm{hav}(\bm\gamma_{j,t_0'}(t'), \bm\gamma) < r_S,\ \forall j,t_0',t'\}$, the set of all geocoordinates across all subjects, starting trajectory times, and times spanned by the trajectories in the observed data within an $r_S$ radius from the coordinates $\bm\gamma$, and $\bm{X}_{t,o,\mathcal{I},(i,t_0)}$ the entries of $\bm X_{t,o}$ for columns $\mathcal{I}$ and rows corresponding to subject $i$ at their start time $t_0$. $\bm X_{1:T,\mathcal{I},u}^{*}$, the values of columns $\mathcal{I}$ for $\bm X_{1:T,u}^{*}$ are then imputed for all observed locations within some radial distance $r_S$ from $\bm\gamma_{i,t_0}(t)$ for each $i$ and $t_0$ with the following Gaussian radial averaging procedure:
\begin{equation}
    \label{eq:radial_averaging}
    \bm{X}_{t,u,\mathcal{I},(i,t_0)}^{*} = \frac{\sum_{t'=1}^{T}\left(\sum_{\bm\gamma_{j,t_0'}(t')\in\mathscr{G}(\bm\gamma_{i,t_0}^{*}(t), r_S)} w(\bm\gamma_{j,t_0'}(t'), \bm\gamma_{i,t_0}^{*}(t))\bm{X}_{t',o,\mathcal{I},(j,t_0')} \right)}{\sum_{t'=1}^{T}\left(\sum_{\bm\gamma_{j,t_0'}(t')\in\mathscr{G}(\bm\gamma_{i,t_0}^{*}(t), r_S)}w(\bm\gamma_{j,t_0'}(t'), \bm\gamma_{i,t_0}^{*}(t))\right)}
\end{equation}
where
\begin{equation*}
    \begin{split}
        w(\bm\gamma_{j,t_0'}(t'), \bm\gamma_{i,t_0}^{*}(t)) = \exp\left(-\frac{(\mathrm{hav}(\bm\gamma_{j,t_0'}(t'), \bm\gamma_{i,t_0}^{*}(t))^2}{2r_S^2}\right)   
    \end{split}
\end{equation*}
and $\mathrm{hav}(\cdot,\cdot)$ denotes the haversine function, which we use to translate distances between geocoordinates into meters.\footnote{Since the factor $(\sqrt{2\pi}r_S)^{-1}$ appears outside of the exponent for all weights and is constant for a set $r_S$, we omit it from $w(\cdot,\cdot)$ for redundancy and to minimize the risk of floating point errors from making $w(\cdot,\cdot)$ too small.}

The covariates specified for imputation by radial averaging are Altitude, Slope, distance to parks, NDVI, and distance from home. In our application, we also set $r_S = 200$ to average measurements within a 200 m. radius from the imputed geolocations.

\section{Neural Network Backpropagation with the Coupling Flow}
\label{sec:backprop_w_CF}

We have parametrized the neural network $f_{\bm\phi}$ with the parameter vector $\bm\phi$, but opt to discuss here specifically what $\bm\phi$ entails. In our implementation and setting, we specify $\bm\phi$ as the set of all the appended weights of every layer that comprises the coupling flow $f_{\bm\phi}$.

\subsection{The Coupling Flow}
\label{sec:couplingflow}

We select the coupling flow for its structure and relative simplicity: Each layer consists of an affine coupling block: a function that incorporates affine transformations, i.e. a scaling and a translation \citep{dinh2017densityestimationusingreal, radev2022}. The specific ways to structure an ACB differs between contexts and depends on the use case. To start, we define the following quantities related to the input vector $\bm u$: 
\begin{equation*}
    \begin{split}
        d := \lvert\bm u\rvert, \quad \bm u_1 := \bm u_{1:\floor{d/2}}, \quad\mbox{and}\quad \bm u_2 := \bm u_{(\floor{d/2}+1):d}
    \end{split}
\end{equation*}

 The \textit{dual coupling block} (DCB) utilized by \citep{radev2022} is then defined in terms of the following operations:
\begin{equation}
    \label{eq:ACB_Radev}
    \begin{split}
        \bm v_1 &= \bm u_2 \odot \exp(g_1(\bm u_1)) + r_1(\bm u_1)\\
        \bm v_2 &= \bm u_1 \odot \exp(g_2(\bm v_1)) + r_2(\bm v_1)
    \end{split}
\end{equation}
where $\odot$ denotes element-wise multiplication and $g_1$, $g_2$, $r_{1}$, and $r_{2}$ are four separate fully-connected neural networks whose weights we will train.\footnote{\citep{radev2022} had $\bm u_1$ and $\bm u_2$ in Equation \eqref{eq:ACB_Radev} swapped in their paper. Equation \eqref{eq:ACB_Radev} specifies the order as is utilized by their software package.} Note that the inverse of the DCB is also readily obtained in the reverse direction:
\begin{equation}
    \label{eq:ACB_Radev_inv}
    \begin{split}
        \bm u_1 &= (\bm v_2 - r_2(\bm v_1)) \odot \exp(-g_2(\bm v_1))\\
        \bm u_2 &= (\bm v_1 - r_1(\bm u_1)) \odot \exp(-g_1(\bm u_1))
    \end{split}
\end{equation}
Due to the design of the DCB, none of $g_1$, $g_2$, $r_1$, or $r_2$ have to be invertible themselves. We call Equation \eqref{eq:ACB_Radev} a "dual coupling block" for reasons that will become apparent.

It is generally easier to work with the \textit{single coupling block} (SCB) of \citep{dinh2017densityestimationusingreal}:
\begin{equation}
    \label{eq:ACB_dinh}
    \begin{split}
        \bm v_1 &= \bm u_1\\
        \bm v_2 &= \bm u_2 \odot \exp(g_1(\bm u_1)) + r_1(\bm u_1)
    \end{split}
\end{equation}
Note that we can rewrite Equation \eqref{eq:ACB_Radev} in terms of compositions of Equation \eqref{eq:ACB_dinh}:
\begin{equation}
    \label{eq:Single2DualCoupling}
    f_{\text{DCB}} = f_{2,\text{SCB}}\circ \bm P\circ f_{1,\text{SCB}},    
\end{equation}
where $\bm P$ denotes the permutation matrix that swaps the two halves of the input vector.

Taking the Jacobian of $f_{1,\text{SCB}}$, which we choose without loss of generality, is simple:
\begin{equation}
    \label{eq:Jacobian_ACB_Dinh}
    \begin{split}
        \bm J_{f_{1,\text{SCB}}} = \begin{bmatrix}
            \bm I_{\floor{d/2}} & \bm O\\
            \frac{\partial\bm v_2}{\partial u_1} & \mathrm{diag}(\exp(g_1(\bm u_1)))
        \end{bmatrix}
    \end{split}
\end{equation}
Its determinant can be computed effortlessly:
\begin{equation}
    \label{eq:detJ_ACB_Dinh}
    \det \bm J_{f_{1,\text{SCB}}} = \prod_{i=1}^{d-\floor{d/2}}\exp(g_1(\bm u_1)_i),
\end{equation}
where $g_1(\bm u_1)_i$ denotes the $i$th entry of $g_1(\bm u_1)$.

It is clear that we can rewrite $f_{\bm\phi}$ in terms of compositions of DCB's, and thus, SCB's. Denote $f_{1,\bm\phi_1},\ldots,f_{\Lambda,\bm\phi_{\Lambda}}$ as our $\Lambda$ SCB's, each parametrized by its own set of weights $\bm\phi_{k}$, $k=1,\ldots,\Lambda$. Then:
\begin{equation}
    \label{eq:CF_comp}
    f_{\bm\phi} = f_{\Lambda,\bm\phi_{\Lambda}}\circ \bm P_{\Lambda -1} \circ f_{\Lambda-1,\bm\phi_{\Lambda-1}}\circ \cdots \circ \bm P_{2}\circ f_{2,\bm\phi_2}\circ \bm P_{1}\circ f_{1,\bm\phi_1}
\end{equation}
with the natural constraint that $\Lambda$ must be even when we layer DCB's, since each DCB consists of two SCB's per Equation \eqref{eq:Single2DualCoupling}. The permutation operations are included to enable different combinations of the entries of the input to be computed with one another; we add that $\bm P_{k}$ for $k$ even is chosen to be a permutation that exchanges two parts of its input in an orthogonal manner compared with the previous $\bm P_{k-2}, \bm P_{k-4},\ldots \bm P_{2}$ permutation matrices, and that for $k$ odd, $\bm P_{k} = \bm P$ from Equation \eqref{eq:Single2DualCoupling}. 

In practice, we specify all $g_k$ and $r_k$ for $f_{k,\bm\phi_k}$, $k=1,\ldots,\Lambda$, to be single-layer 128-length perceptrons with ReLU activation functions, with an output layer added at the end to enable the dimensions of the output to conform to the other half of the input vector (particularly relevant when $\bm u$ is odd). Taking $g_1$ as an example,
\begin{equation}
    \label{eq:CF_compspec}
    g_1(\bm u_1) = \bm W_{1o}\mathrm{ReLU}(\bm W_1\bm u_1),
\end{equation}
so that $\bm W_1$ and $\bm W_{1o}$ are flattened to form part of the $\bm\phi_1$ vector, which is then incorporated into the entire $\bm\phi$. $\bm W_{1o}$ is the final output matrix with $d - \floor{d/2}$ rows to ensure that $g_{1}(\bm u_1)$ has the same length as $\bm u_2$; the entries of $\bm W_{1o}$ are also trainable.\footnote{$W_{1o}$ has no activation function by implementation and conventional usage of output layers.} ReLU is defined as $\mathrm{ReLU}(x) = \max(x,0)$ and is understood to apply entrywise where its argument is a vector or matrix. Its derivative is the Heaviside step function $\mathrm{H}(x) = 1_{x>0}$, i.e. $1$ if its argument is positive and $0$ otherwise.\footnote{The value of $\mathrm{H}(0)$ is sometimes explicitly set depending on the user or application domain, despite it not having a proper value.} Both for simplicity and because this is how we have utilized the component neural networks of the single and dual coupling flows in practice, we will assume all of $g_k$ and $r_k$ follow the structure of $g_1$ in Equation \eqref{eq:CF_compspec} for all coupling flow layers of $f_{\bm\phi}$.

\subsection{Backpropagation}
\label{sec:backprop}

We begin by discussing the objective function in Equation \eqref{eq:bf_objfunc}, where the $\argmin$ at the end of each training cycle is taken over the $\bm\phi$ for the objective function to minimize it. For clarity, we bring in the loss term to be minimized at the end of one training step of \texttt{BayesFlow} and simplify it in terms of the quantities in Equation \eqref{eq:detJ_ACB_Dinh}:
\begin{equation}
    \label{eq:kl_loss}
    \begin{split}
    \mathcal{L}(f_{\bm\phi}; \bm\theta^{(1:M)},\bm y^{(1:M)}) =& \frac{1}{M}\sum_{m=1}^{M}\left(\frac{1}{2}\lvert\lvert f_{\bm\phi}(\bm\theta^{(m)}; \bm y^{(m)})\rvert\rvert^{2} - \log\lvert\det \bm J_{f_{\bm\phi}}^{(m)}\rvert\right)\\
    =& \frac{1}{M}\sum_{m=1}^{M}\left(\frac{1}{2}f_{\bm\phi}(\bm\theta^{(m)}; \bm y^{(m)})^\T f_{\bm\phi}(\bm\theta^{(m)}; \bm y^{(m)}) \right.\\
    &\left.- \sum_{k=1}^{\Lambda}\sum_{i}g_{k}(\bm u_{k-1})_i\right)
    \end{split}
\end{equation}
where $\bm u_{k-1}$ denotes the output of $\bm P_{k-1}\circ f_{k-1,\bm\phi_{k-1}}\circ\cdots\circ f_{1,\bm\phi_1}$ and $g_{k}$ the scaling neural network analogous to $g_1$ in Equation \eqref{eq:ACB_dinh}, but specific to the $k$th SCB layer. We avoid specifying the number of indices that $i$ ranges, as it will differ between layers if $\bm\theta^{(m)}$'s length is odd.

In its most fundamental implementation, at the end of one training cycle, $\bm\phi \gets \bm\phi - \alpha \nabla_{\bm\phi}\mathcal{L}(f_{\bm\phi}; \bm\theta^{(1:M)},\bm y^{(1:M)})$, where $\alpha$ controls how far to allow the gradient descent update. In practice, the gradient term $\nabla_{\bm\phi}\mathcal{L}(f_{\bm\phi}; \bm\theta^{(1:M)},\bm y^{(1:M)})$ may be replaced by a more convenient expression or algorithm. $\alpha$ is externally specified and controlled various parameters, many from the user. As we will see, specific strategies to control the gradient descent such as ADAM can be used that will substitute the gradient term with other related quantities before updating $\bm\phi$. We defer to subsection \ref{sec:alpha} for a more detailed discussion.

Backpropagation is utilized to compute the gradient term $\nabla_{\bm\phi}\mathcal{L}(f_{\bm\phi}; \bm\theta^{(1:M)},\bm y^{(1:M)})$ efficiently without repeating terms between each layer \citep{werbos1982,rumelhart1986}. It generalizes across different machine learning architectures and loss functions. Spelling it out in terms of dense neural networks, a basic use case that motivates iterated updates of weights per layer, is well-known in the machine learning community \citep{bishop2006pattern, murphy2012machine}; we repeat the calculations specifically for the coupling flow for illustrative purposes.

We begin by taking a single term in the sum of Equation \eqref{eq:kl_loss} and differentiating it with respect to a single weight, say $w_{kij}$, the weight corresponding to the $i$th row and $j$th column of the weights $\bm W_{k}$ corresponding to the hidden layer in $g_{k}$:
\begin{equation}
    \label{eq:dlossw}
    \begin{split}
    \frac{\partial}{\partial w_{kij}}&\left(\frac{1}{2}f_{\bm\phi}(\bm\theta^{(m)}; \bm y^{(m)})^\T f_{\bm\phi}(\bm\theta^{(m)}; \bm y^{(m)}) - \sum_{k=1}^{\Lambda}\sum_{i'}g_{k}(\bm u_{k-1})_{i'} \right)\\
    =& f_{\bm\phi}(\bm\theta^{(m)}; \bm y^{(m)})^\T\frac{\partial f_{\bm\phi}(\bm\theta^{(m)}; \bm y^{(m)})}{\partial w_{kij}} - \sum_{i'}\frac{\partial g_{k}(\bm u_{k-1})_{i'}}{\partial w_{kij}}
    \end{split}
\end{equation}
We note that $w_{kij}$ affects $f_{\bm\phi}$ only via $g_{k}$. It is sensible, then, to define $\bm a_{k} :=\bm W_{k}\bm u_{k-1}$ and to compute each derivative with respect to $a_{ki}$, as only the $i$th row of $\bm W_{k}$ is relevant to the derivatives:
\begin{equation}
    \label{eq:derivs_a}
    \begin{split}
        \frac{\partial f_{\bm\phi}(\bm\theta^{(m)}; \bm y^{(m)})}{\partial w_{kij}} &= \frac{\partial f_{\bm\phi}(\bm\theta^{(m)}; \bm y^{(m)})}{\partial a_{ki}}\frac{d a_{ki}}{dw_{kij}} =  \frac{\partial f_{\bm\phi}(\bm\theta^{(m)}; \bm y^{(m)})}{\partial a_{ki}}u_{(k-1)j}\\
        \frac{\partial g_{k}(\bm u_{k-1})_i}{\partial w_{kij}} &= w_{k(o)ii}h_{k}'(a_{ki})\frac{da_{ki}}{dw_{kij}} = w_{k(o)ii}h_{k}'(a_{ki})u_{(k-1)j}
    \end{split}
\end{equation}
where $h_{k}$ is the activation function corresponding to $g_{k}$, so that $h_{k}(\bm a_{k})$ forms the output of the first layer of the neural network output of $g_{k}$ and $\bm w_{k(o)ii}$ is the entry from the $i$th row and column of $\bm W_{ko}$, the final output matrix of $g_k$. (Note that carrying out the differentiation in the second line of Equation \eqref{eq:derivs_a} with respect to $g_{k}(\bm u_{k-1})_{i'}$ for $i'\neq i$ would yield zero.) We may also decompose the partial derivative $\partial f_{\bm\phi}/\partial a_{ki}$ in terms of the functions at the output layer $\bm W_{ko}$ for $k < \Lambda$:
\begin{equation}
    \label{eq:df_decomp}
    \begin{split}
        \frac{\partial f_{\bm\phi}(\bm\theta^{(m)}; \bm y^{(m)})}{\partial a_{ki}} &= \sum_{i'}\frac{\partial f_{\bm\phi}(\bm\theta^{(m)};\bm y^{(m)})}{\partial a_{k(o)i'}}\frac{\partial a_{k(o)i'}}{\partial a_{ki}}\\
        &= \sum_{i'} \frac{\partial f_{\bm\phi}(\bm\theta^{(m)};\bm y^{(m)})}{\partial a_{k(o)i'}}w_{k(o)i'i}h_{k}'(a_{ki})
    \end{split}
\end{equation}
where $\bm a_{k(o)} = \bm W_{k(o)}h_{k}(\bm W_{k}\bm u_{k}) = g_{k}(\bm u_{k})$ and $w_{k(o)i'i}$ denotes the $i'$th row and $i$th column entry of $\bm W_{ko}$. 

We also require derivative terms with respect to the entries of $\bm W_{ko}$, because it is a neural network weight layer that needs to be trained as well. Some of its entries would be processed by $g_{k+1}$ and $r_{k+1}$ and others by $g_{k+2}$ and $r_{k+2}$ by virtue of the permutation matrices between successive SCB's. We therefore adopt the notation of children for the index of the weight matrices to formalize this dependence structure for $k < \Lambda - 1$:
\begin{equation}
    \label{eq:children}
    \begin{split}
        \mathrm{ch}[k] &= \{k(o)\}, \quad\mbox{and}\\
        \mathrm{ch}[k(o)] &= \{k+1, (k+1)(r),k+2, (k+2)(r)\}.
    \end{split}
\end{equation}

While the weights of $r_{k}$ are separate (and we label the post-weight multiplication accordingly with $\bm a_{k(r)}$ and $\bm a_{k(ro)}$), Equation \eqref{eq:children} applies analogously to the weights of $r_{k}$. Substituting $k(r)$ and $k(ro)$ into the respective arguments of the two equations in Equation \eqref{eq:children} keeps their right-hand sides the same. Extending Equation \eqref{eq:df_decomp} to the output layers of layer $k$ gives us:
\begin{equation}
    \label{eq:df_decomp_outlayer}
    \begin{split}
        \frac{\partial f_{\bm\phi}(\bm\theta^{(m)}; \bm y^{(m)})}{\partial a_{k(o)i}} &= \sum_{c\in \mathrm{ch}[k(o)]}\sum_{i'}\frac{\partial f_{\bm\phi}(\bm\theta^{(m)};\bm y^{(m)})}{\partial a_{ci'}}w_{ci'i}
    \end{split}
\end{equation}

What we have done with Equation \eqref{eq:df_decomp} and \eqref{eq:df_decomp_outlayer} was to rewrite the derivative of $f_{\bm\phi}$ with respect to $\bm a_{k}$ in terms of the derivative of $f_{\bm\phi}$ with respect to the next layer after $\bm a_{k}$ has been passed through the activation function $h_{k}$. This means that we gather the value of the output and its derivatives in one forward pass, and then compute the gradient with respect to each of the weights by passing the values of the derivatives backwards, from the last layer in the network down to the first. Algorithm \ref{alg:backprop_CF} demonstrates this procedure in full.
 
Overall, the backpropagation procedure is not significantly different for the coupling flow than it is for a regular feedforward network. Besides the more complicated objective function by virtue of its domain, the procedure which backpropagation is utilized to update the weights $\bm\phi$ remains largely the same because the underlying neural network still utilizes a forward pass before relevant quantities and derivatives are propagated backwards.

\begin{algorithm}
    \caption{One-step Backpropagation with the Coupling Flow}\label{alg:backprop_CF}
    \begin{algorithmic}[1]
        \vspace{2mm}
        \State \textbf{Input:} Coupling Flow neural network $f_{\bm\phi}$
        \vspace{2mm}
        \State \qquad\quad Loss $\mathcal{L}(f_{\bm\phi}; \bm\theta^{(1:M)}, \bm y^{(1:M)})$
        \vspace{2mm}
        \Function{\texttt{BackProp\_CF}}{$f_{\bm\phi}, \mathcal{L}(f_{\bm\phi}; \bm\theta^{(1:M)}, \bm y^{(1:M)})$}
        \vspace{2mm}
        \For{$m=1,\ldots,M$}
            \vspace{2mm}
            \State Compute $f_{\bm\phi}(\bm\theta^{(m)};\bm y^{(m)})$, and $\bm a_{k}$, $\bm a_{k(o)}$, $\bm a_{k(r)}$, and $\bm a_{k(ro)}$ for $k=1,\ldots,\Lambda$ through the forward pass.
            \vspace{2mm}
            \State Compute the derivatives of $f_{\bm\phi}$ with respect to the entries of $\bm W_{\Lambda(o)}$, $\bm W_{\Lambda(ro)}$, $\bm a_{\Lambda(o)}$ and $\bm a_{\Lambda(ro)}$ through Equation \eqref{eq:df_decomp_outlayer}.
            \vspace{2mm}
            \State Compute the derivatives of $f_{\bm\phi}$ with respect to the entries of $\bm W_{\Lambda}$, $\bm W_{\Lambda(r)}$, $\bm a_{\Lambda}$ and $\bm a_{\Lambda(r)}$ through Equation \eqref{eq:df_decomp}.
            \vspace{2mm}
            \For{$k=\Lambda-1,\ldots,1$}
                \vspace{2mm}
                \State Compute the derivatives of $f_{\bm\phi}$ with respect to the entries of $\bm W_{k(o)}$, $\bm W_{k(ro)}$, $\bm a_{k(o)}$ and $\bm a_{k(ro)}$ through Equation \eqref{eq:df_decomp_outlayer}, using the derivatives from layer $k+1$.
                \vspace{2mm}
                \State Compute the derivatives of $f_{k,\bm\phi_{k}}$ with respect to the entries of $\bm W_{k}$, $\bm W_{k(r)}$, $\bm a_{k}$ and $\bm a_{k(r)}$ through Equation \eqref{eq:df_decomp}, using the derivatives from layer $k(o)$.
                \vspace{2mm}
                \State Repeat line 9 for $g_k$ and $r_k$, but only with $\bm W_{k(o)}$ and $\bm a_{k(o)}$ for $g_k$, and $\bm W_{k(ro)}$ and $\bm a_{k(ro)}$ for $r_k$.
                \vspace{2mm}
                \State Repeat line 10 for $g_k$ and $r_k$, but only with $\bm W_{k}$ and $\bm a_{k}$ for $g_k$, and $\bm W_{k(r)}$ and $\bm a_{k(r)}$ for $r_k$.
                \vspace{2mm}
            \EndFor
            \vspace{2mm}
        \EndFor 
        \vspace{2mm}
        \State Gather the component partial derivatives of $f_{\bm\phi}$, $g_{k}$, and $r_{k}$, with respect to the entries of the weight matrices for all $k$, to compute $\nabla_{\bm\phi}\mathcal{L}(f_{\bm\phi};\bm\theta^{(m)},\bm y^{(m)})$.
        \vspace{2mm}
        \State Compute $\nabla_{\bm\phi} \mathcal{L}(f_{\bm\phi};\bm\theta^{(1:M)},\bm y^{(1:M)}) = \sum_{m=1}^{M}\nabla_{\bm\phi}\mathcal{L}(f_{\bm\phi};\bm\theta^{(m)},\bm y^{(m)})$.
        \vspace{2mm}
        \State \Return $\nabla_{\bm\phi} \mathcal{L}(f_{\bm\phi};\bm\theta^{(1:M)},\bm y^{(1:M)})$ 
        \vspace{2mm}
        \EndFunction
        \vspace{2mm}
    \end{algorithmic}
\end{algorithm}

\subsection{Parameter Update Control}
\label{sec:alpha}

Controlling the update step is desired to avoid exploding or vanishing gradients, although too much control may result in convergence at local minima instead of tending towards a global minima. We utilize the ADAptive Moment estimation, or \texttt{ADAM}, optimizer, specifying the cosine decay strategy of \citep{loshchilov2017cosinedecay} to avoid overshooting the optimal solution, and cap the norm of the gradient at 1 to avoid numerical instability from exploding or vanishing gradients.

We detail \texttt{ADAM} \citep{KingmaBa2017ADAM} in Algorithm \ref{alg:ADAM}. \texttt{ADAM} presents an alternative to standard batch gradient descent, with advantages in that it works with sparse gradients and naturally performs a form of step size annealing, or reducing the learning rate over time to improve convergence. Adopting \texttt{ADAM} as our alternative to gradient descent would replace line 13 in Algorithm \ref{alg:ABI} with a specification that \texttt{ADAM} specifically is utilized; backpropagation is still used in both approaches.\footnote{Other popular optimizers include SGD, RMSprop, and Adagrad. The interested reader is encouraged to consult https://keras.io.api/optimizers/ for more information.} Here, we skip the new generation of $\bm\theta^{(1:M)}$ and $\bm y^{(1:M)}$ present in \texttt{TRAIN\_BAYESFLOW} Algorithm \ref{alg:ABI} for the sake of describing \texttt{ADAM}.

\begin{algorithm}
    \caption{ADAptive Moment estimation (ADAM)}
    \label{alg:ADAM}
    \begin{algorithmic}[1]
        \vspace{2mm}
        \State \textbf{Input:} INN $f_{\bm\phi}$, loss $\mathcal{L}(f_{\bm\phi}; \bm\theta^{(1:M)}, \bm y^{(1:M)})$, and initial parameter vector $\bm\phi$.
        \vspace{2mm}
        \State \qquad\quad Stepsize $\alpha()$. \Comment{\citep{KingmaBa2017ADAM} recommend $\alpha() = 0.001$ as a good default.}
        \vspace{2mm}
        \State \qquad\quad Exponential decay rates $\delta_1,\delta_2 \in [0,1)$ for moment estimates.
        \State \Comment{$\delta_1 = 0.9, \delta_2 = 0.999$ are considered good default settings.}
        \vspace{2mm}
        \State \qquad\quad Number of iterations $N_{ITER}$.
        \vspace{2mm}
        \Function{\texttt{ADAM}}{$\alpha(), \delta_1, \delta_2, \mathcal{L}(f_{\bm\phi}; \bm\theta^{(1:M)}, \bm y^{(1:M)}), \bm\phi$}
        \vspace{2mm}
            \State Initialize $\bm m_0 \gets \bm 0$, $\bm v_0 \gets \bm 0$, $\epsilon \gets 10^{-8}$.
            \vspace{2mm}
            \For{$j=1,\ldots,N_{ITER}$}
                \vspace{2mm}
                \State $\bm g_j \gets \texttt{BACKPROP\_CF}(\bm f_{\bm\phi}; \mathcal{L}(f_{\bm\phi}; \bm\theta^{(1:M)}, \bm y^{(1:M)}))$
                \vspace{2mm}
                \State $\bm m_j \gets \delta_1 \bm m_{j-1} + (1 - \delta_1)\bm g_j$
                \vspace{2mm}
                \State $\bm v_j \gets \delta_2 \bm v_{j-1} + (1 - \delta_2)\bm g_j^2$ \Comment{$\bm g_j$ is squared entrywise.}
                \vspace{2mm}
                \State $\widehat{\bm m}_j \gets \bm m_j/(1 - \delta_1^j)$ \Comment{Correcting for bias for first and second moment estimates.}
                \vspace{2mm}
                \State $\widehat{\bm v}_j \gets \bm v_j / (1 - \delta_2^j)$
                \vspace{2mm}
                \State $\bm \phi \gets \bm\phi - \alpha() \widehat{\bm m}_j/(\sqrt{\widehat{\bm v}_j}+ \epsilon)$ \Comment{Division is also entrywise.}
                \vspace{2mm}
            \EndFor
        \vspace{2mm}
        \State\Return $\bm\phi$
        \vspace{2mm}
        \EndFunction
        \vspace{2mm}
    \end{algorithmic}
\end{algorithm}

We write $\alpha$ as the function $\alpha()$ because, in practice, the step size accounts for the computed gradient of the loss function, the number of iterations, or other variables that control the parameter update. These additional parameters are passed to $\alpha()$, e.g. $\alpha(\nabla_{\bm\phi} f_{\bm\phi})$, to enable it to control the step size of each update.

\section{\texttt{BayesFlow} Implementation with the Summary Network}
\label{sec:summarynet}

\citep{radev2022} implements the \texttt{BayesFlow} software package with the aid of another neural network $h_{\bm\psi}$ parametrized by $\bm\psi$ to condense their output into a summary statistic of fixed length, which is called the \textit{summary network}. This is further enhanced by allowing their underlying model $p(\bm y\mid \bm\theta)$ to sample multiple instances $N$ of the synthetic outcome at once and allowing it to differ between distinct Monte Carlo batches, so that $h_{\bm\psi}(\bm y_{1:N})$ takes in multiple samples of the data and learns a function to summarize the data even for different numbers of synthetic data instances. Owing to the particular structure of the actigraph data with its missingness resulting in trajectories of different lengths and places in time, which would make constructing a summary network on subsamples difficult to design, only $N = 1$ is relevant to our particular case. Additionally, while our usage of ABI depends on \cite{radev2022}'s package, we bypass the use of $h_{\bm\psi}$ by setting $h_{\bm\psi} = \text{id}$, the identity transform.

Still, there is practical justification for the use of the summary network. Previous generative models such as variational autoencoders have utilized neural networks to encode data into more compact representations \citep{vae2013}. A less analytically tractable model such as the Lotka-Volterra may be better summarized by a fitted though generally noninterpretable means when the summary procedure itself is not of interest; Radev et al. demonstrate the superior parameter recovery capabilities of a neural network implementation in their \cite[Supplement]{radev2022}. Notably, in our applications of \texttt{BayesFlow} with a nontrivial $h_{\bm\psi}$, we have found that a relatively small speedup can be measured\footnote{Approximately 1.03 hours could be saved, corresponding to an approximate 8\% speedup for the first training epoch where the corresponding runtime without the summary network would have been close to 13 hours.} if a summary network is employed for coarse blocking arrangements of Algorithm \ref{alg:hier_ABI_DLM_timebatch}, specifically where the blocks of $\bm y_{T_{b-1}:T_{b}}^{(m)}$ totaled 2,130 entries per sample for $T_{b-1} = 1$ to $T_{b} = 33$ and the number of Monte Carlo samples used was $1,024$. In this setting, $h_{\bm\psi}$ was specified as a three-layer neural network with a 1024-length dense layer with a Rectified Linear Unit (ReLU) activation function, a dropout layer to avoid overfitting, and a $p(T_b - T_{b-1} + 1) + 2$-length dense layer for batch $b$ with no activation function to summarize the statistics into the number of parameters needed to fully express the model at each time step.

The inclusion of the summary network turns our problem into inference of $\bm\theta\mid h_{\bm\psi}(\bm y)$, and the training goal involves finding the optimal values of $\bm\psi$ in addition to $\bm\phi$, with the expression on the left hand side of Equation \ref{eq:kl_2_objective} now including the summary network term:
\begin{equation}
    \label{eq:kl_objective_sum}
    \begin{split}
        \argmin_{\bm\phi, \bm\psi} \mathbb{E}_{p(\bm y)}\mathbb{KL}(p(\bm\theta\mid \bm y)\mid\mid p(f_{\bm\phi}^{-1}(\bm z; h_{\bm\psi}(\bm y))))
    \end{split}
\end{equation}

The change carries analogously to the objective function (Equation \eqref{eq:bf_objfunc}), which now requires minimization $\bm\psi$ and incudes $h_{\bm\psi}(\bm y^{(m)})$ in place of $\bm y^{(m)}$:
\begin{equation}
    \label{eq:bf_objfunc_sum}
    \begin{split}
        \widehat{\bm\phi}, \widehat{\bm\psi} = \argmin_{\bm\phi, \bm\psi}\frac{1}{M}\sum_{m=1}^{M}\left(\frac{1}{2}\lvert\lvert f_{\bm\phi}(\bm\theta^{(m)}; h_{\bm\psi}(\bm y^{(m)}))\rvert\rvert^{2} - \log\lvert\det \bm J_{f_{\bm\phi}}^{(m)}\rvert\right)
    \end{split}
\end{equation}

Equation \eqref{eq:bf_objfunc_sum} is also incorporated analogously into the procedure for Algorithm \ref{alg:ABI}, with the addition of the summary network $h_{\bm\psi}$ and its related parameters and terms.

\end{document}